\documentclass[10pt,twocolumn,letterpaper]{article}

\usepackage{cvpr}              

\definecolor{cvprblue}{rgb}{0.21,0.49,0.74}
\usepackage[pagebackref,breaklinks,colorlinks,allcolors=cvprblue]{hyperref}
\usepackage{varwidth}
\usepackage{tabularx}
\usepackage{makecell}
\usepackage{amsmath}
\usepackage{dirtytalk}
\usepackage[figuresright]{rotating}
\usepackage{multicol}

\usepackage{collcell}
\usepackage{hhline}
\usepackage{pgf}

\def\ccainscifar#1{%
	\pgfmathsetmacro\calc{100*(#1-0.534)/(0.599-0.534)}%
	\edef\clrmacro{\noexpand\cellcolor{blue!\calc}}%
	\clrmacro%
	\ifdim \calc pt>50pt\color{white}\fi{#1}%
}

\def\ccadelcifar#1{%
	\pgfmathsetmacro\calc{100*(1.0 - (#1-0.299)/(0.607-0.299))}%
	\edef\clrmacro{\noexpand\cellcolor{olive!\calc}}%
	\clrmacro%
	\ifdim \calc pt>50pt\color{white}\fi{#1}%
}

\def\ccainsferplus#1{%
	\pgfmathsetmacro\calc{100*(#1-0.186)/(0.379-0.186)}%
	\edef\clrmacro{\noexpand\cellcolor{blue!\calc}}%
	\clrmacro%
	\ifdim \calc pt>50pt\color{white}\fi{#1}%
}

\def\ccadelferplus#1{%
	\pgfmathsetmacro\calc{100*(1.0 - (#1-0.239)/(0.365-0.239))}%
	\edef\clrmacro{\noexpand\cellcolor{olive!\calc}}%
	\clrmacro%
	\ifdim \calc pt>50pt\color{white}\fi{#1}%
}

\def\paperID{UNCV-14} 
\def\confName{CVPR}
\def\confYear{2025}

\title{Uncertainty Quantification for Gradient-based Explanations in Neural Networks}

\author{Mihir Mulye\\
Hochschule Bonn-Rhein-Sieg, Sankt Augustin, Germany \\
Heinrich-Heine University, Duesseldorf, Germany \\
{\tt\small mihir.mulye@hhu.de}
\and
Matias Valdenegro-Toro\\
Department of Artificial Intelligence\\
University of Groningen, Netherlands \\
{\tt\small m.a.valdenegro.toro@rug.nl}
}

\begin{document}
\maketitle
\begin{abstract}
Explanation methods help understand the reasons for a model's prediction. These methods are increasingly involved in model debugging, performance optimization, and gaining insights into the workings of a model. With such critical applications of these methods, it is imperative to measure the uncertainty associated with the explanations generated by these methods. In this paper, we propose a pipeline to ascertain the explanation uncertainty of neural networks by combining uncertainty estimation methods and explanation methods. We use this pipeline to produce explanation distributions for the CIFAR-10, FER+, and California Housing datasets. By computing the coefficient of variation of these distributions, we evaluate the confidence in the explanation and determine that the explanations generated using Guided Backpropagation have low uncertainty associated with them. Additionally, we compute modified pixel insertion/deletion metrics to evaluate the quality of the generated explanations. 
\end{abstract}
\section{Introduction}
\label{sec:intro}

\begin{figure}[t]
    \centering
    \begin{subfigure}{0.13\textwidth}
        \includegraphics[width=0.75\textwidth, height=1.75cm]{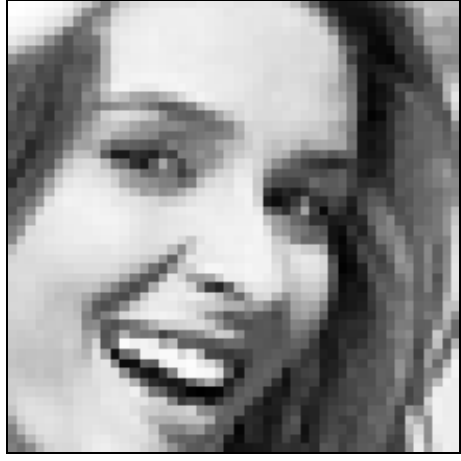}
        \caption{Input}
    \end{subfigure}
    \begin{subfigure}{0.15\textwidth}
        \includegraphics[width=0.75\textwidth, height=1.75cm]{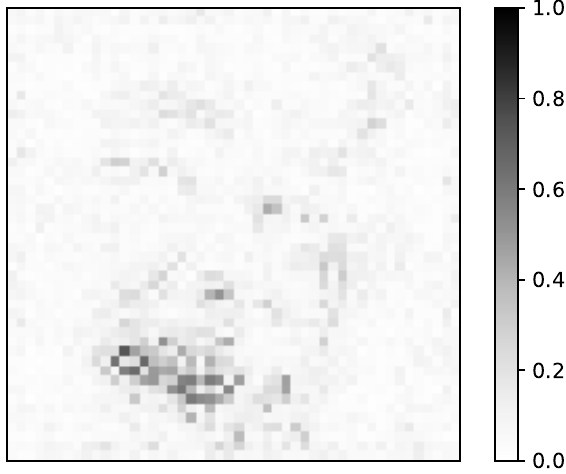}
        \caption{$E_\mu$}
    \end{subfigure}
    \begin{subfigure}{0.15\textwidth}
        \includegraphics[width=0.75\textwidth, height=1.75cm]{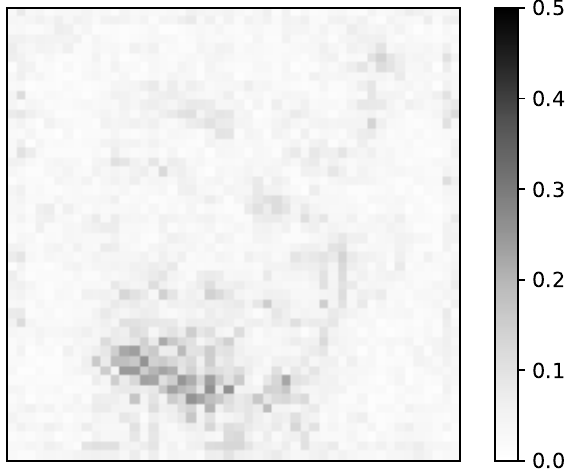}
        \caption{$E_\sigma$}
    \end{subfigure}
    \caption{Visualization of explanation with uncertainty on an image from FER+ dataset using Guided Backpropagation as explanation method and Deep Ensembles as uncertainty estimation method. (a) Input Image, (b) Explanation Mean, and (c) Explanation Uncertainty.}
    \label{example_of_explanation_uncertainty}
\end{figure}

With the increase in the application of neural networks in safety-critical cases such as autonomous driving, medical diagnosis, and quality control procedures, the number of methods developed to explain the network output has also increased. These explanation methods are now being coupled with the knowledge of human experts to
understand the workings, debug and optimize the performance, and sometimes even extend the scope of application of these neural networks. As explanation methods gradually become indispensable, it is also critical to analyze the
uncertainty in the output of these explanation methods themselves (see Figure \ref{example_of_explanation_uncertainty}) as it would help in justifying the trust being placed in these explanations. %
\par  
In this paper, we attempt to solve this problem and investigate the possibility of combining the concepts of uncertainty estimation with the explainability of neural networks to quantify the confidence in the importance attributed to the input features. To this end, we design and implement a pipeline that induces a distribution in the explanation method output of a neural network by utilizing the prediction distribution of the said neural network trained using uncertainty estimation methods. We verify the robustness of the aforementioned pipeline by testing it on two learning tasks namely, image classification and tabular regression. We select the CIFAR-10 \cite{krizhevsky2009learning} and FER+ \cite{barsoum2016training} datasets for testing the pipeline on the task of image classification and choose the California Housing Dataset \cite{pace1997sparse} for testing on the task of tabular regression.
\par
By implementing our proposed approach, we attempt to answer the following research questions: RQ1: Could uncertainty estimation methods and explanation methods be combined to ascertain the explanation uncertainty of neural networks? RQ2: How can the explanation uncertainty by combined into a single representation of explanation uncertainty? RQ3: What are the possible approaches to evaluate explanation uncertainty? RQ4: Which combination of uncertainty estimation method and explanation method is the best to analyze the explanation uncertainty output for a neural network?

\par
The contributions of this paper are: we propose and implement a pipeline that helps ascertain the uncertainty in the explanation method output of a neural network. To represent this, we compute the mean, the standard deviation, and the coefficient of variation of the generated explanation distribution. We propose the coefficient of variation as a way to merge the explanation with its uncertainty into a single saliency explanation. Additionally, we modify the pixel insertion/deletion metric (originally defined for a single explanation instance) such that it can be used to quantify the quality of explanation distributions. Finally, we demonstrate the robustness of our proposed pipeline by testing it on two tasks namely, image classification (CIFAR-10 and FER+) and tabular regression (California Housing dataset).

\section{Related Work}
\label{sec:related_work}

\subsection{Uncertainty Estimation Methods}
Classical neural networks produce outputs using fixed weights and parameters. The outputs of these networks are deterministic, in that, for multiple forward passes for a given input, the prediction of the network remains unchanged. However, several approaches have been proposed to train neural networks capable of generating outputs with uncertainty. These methods are chiefly divided into two categories: \textit{(i)} Bayesian methods and their approximations and \textit{(ii)} Non-Bayesian approaches. 
\par
Bayesian methods treat the weights of the neural network as probability distributions as opposed to singular values. However, as implementing the backpropagation algorithm over a parametric distribution is computationally intensive, approximations of Bayesian methods are used more extensively. Monte Carlo (MC) Dropout \cite{gal2016dropout}\cite{srivastava2014dropout} belongs to this category. This method enables Dropout at inference time as well thereby inducing stochasticity in the model output. Another approach in this category is MC DropConnect \cite{mobiny2021dropconnect}\cite{wan2013regularization} which is similar to MC-Dropout. However, in this case, weights are set to zero instead of activations during inference to produce stochasticity. Flipout \cite{wen2018flipout} are also categorized as Bayesian Neural Networks. The weights are approximated with a Gaussian distribution $q_{\theta}(W)$ and the loss that approximates Evidence Lower Bound (ELBO) is defined as:
\begin{equation}
    L(\theta) = \text{KL}(\mathbb{P}(W), q_\theta(W)) - \mathbb{E}_{w \sim q_\theta(W)}[\log \mathbb{P}(y \, | \, x, w)]
\end{equation}
where $\text{KL}(\mathbb{P}(W), q_\theta(W))$ is the Kullback-Leibler divergence between the prior $\mathbb{P}(W)$ and the approximate posterior weight distribution denoted by $q_\theta(W)$. $-\log \mathbb{P}(y \, | \, x, w)$ is the negative log-likelihood that is calculated as an expectation by sampling weights from the $q_{\theta}(W)$. The negative log-likelihood takes the definition of mean squared error for regression and cross-entropy for classification.

\par 
Non-Bayesian methods, on the other hand, treat the weights of the network as deterministic. The uncertainty is introduced by training multiple copies of the same model architecture and using these trained model instances to predict an output distribution for a given input. A key aspect of this training paradigm is that each model instance is initialized randomly which results in differently learned weights and subsequently a different output for the same given input. This output distribution can then be analyzed as output with uncertainty by computing its mean and standard deviation. 
Deep Ensembles \cite{lakshminarayanan2016simple} %
belong to this category. 

\par
As the models trained using uncertainty estimation generate multiple samples for the same given input, these samples need to be combined to estimate the model uncertainty. For the task of classification, the mean of the predicted probability vector is computed, and for the regression task, the mean ($\mu(x)$) and the variance (and by extension standard deviation) ($\sigma^{2}(x)$) for the samples is computed as:
\begin{align}
    \mu(x) = T^{-1} \sum f(x)  , &&  \sigma^2(x) = T^{-1} \sum (f(x) - \mu(x))^2
\end{align}
where $T$ is the number of samples (or models in the case of Ensembles). The uncertainty is then defined by the standard deviation $\sigma(x)$.

\subsection{Saliency Explanation Methods} 
\par
Explanation methods help in understanding why the neural network produces a particular output for a given input. In this section, we discuss several such approaches. 

Class Activation Maps \cite{zhou2016learning} uses the global average pooling layer in CNNs to produce heatmaps that highlight pixels relevant to the model for generating an output. A drawback of this approach is that it can only be applied to models that have a global average pooling layer in their architecture. Layer-wise Relevance Propagation \cite{bach2015pixel} computes the relevance of input features by propagating the network prediction back to the input according to predefined local relevance propagation rules \cite{montavon2019layer}.
\par 
Integrated Gradients (IG)\cite{sundararajan2017axiomatic} produces explanations by integrating the gradient of output with respect to the input features along a linear path from a baseline image (usually a blank image with black pixels) to the input image. This is mathematically represented as:
\begin{equation}
    \text {IG}_{i}(x, F) =\left(x_{i}-x_{i}^{\prime}\right) \times \int_{\alpha=0}^{1} \frac{\partial F\left(x^{\prime}+\alpha \times\left(x-x^{\prime}\right)\right)}{\partial x_{i}} d \alpha
    \label{ig_equation}
\end{equation}
where $x$ is the input, $i$ is the feature dimension, $x^{'}$ is the baseline input, and $\alpha$ is a factor of interpolation that determines the amount of feature perturbation, and $F()$ is the function mapping the input to the output. Computing the integral in the Equation \ref{ig_equation} is difficult therefore the solution is obtained using numerical approximation as:
\begin{equation}
    \text {IG}_{i}(x, F) =\left(x_{i}-x_{i}^{\prime}\right) \times m^{-1} \sum_{k=1}^{m} \frac{\partial F\left(x^{\prime}+\frac{k}{m} \times\left(x-x^{\prime}\right)\right)}{\partial x_{i}}
    \label{ig_equation_approx}
\end{equation}
where $m$ denotes the number of steps in the numerical approximation of the integral. 
\par 
Guided Backpropagation (GBP)\cite{springenberg2014striving} computes explanations by multiplying the gradients computed during backpropagation by the sign of corresponding activations from the forward pass. This amplifies the gradients corresponding to positive activation and suppresses gradients corresponding to negative activations. The mathematical representation is presented as:
\begin{equation}
    \label{eq:activation}
    f_{i}^{l+1}=\operatorname{\textit{relu}}\left(f_{i}^{l}\right)=\max \left(f_{i}^{l}, 0\right)
\end{equation}
\begin{equation}
    \label{eq:gbp}
    R_{i}^{l}=\left(f_{i}^{l}>0\right) \cdot\left(R_{i}^{l+1}>0\right) \cdot R_{i}^{l+1}
\end{equation}
where $R$ is the reconstructed image obtained at layer $l$, $f_{i}^{l}$ is the positive forward pass activations and $R_{i}^{l+1}$ stands for positive error signals.

\par 
Local Interpretable Model-agnostic Explanations (LIME) \cite{ribeiro2016should} proposes substituting a complex learning model with a simple surrogate model in a local neighborhood. This simplified model is easier to explain in that particular neighborhood. LIME involves sampling around the input to be explained, and then fitting a linear surrogate model that is explainable in that neighborhood. This approach can be applied to a variety of input modalities such as text, tabular data, and images.

\subsection{Explanation Uncertainty in Neural Networks}
With advances in both uncertainty estimation methods and explainability, researchers have attempted to combine these concepts and produce uncertainty in the explanation method output of neural networks. Slack et al. \cite{slack2021reliable} utilize a Bayesian framework to analyze explanation uncertainty in LIME and KernelSHAP. Zhang et al. \cite{zhang2019should} aim to find sources of uncertainty in LIME by investigating randomness in sampling, and variation of explanation in differing proximities to data points. 
\par 
Bykov et al. \cite{bykov2021explaining} combine Bayesian Neural Networks and Layer-wise Relevance Propagation to generate explanation uncertainty. Wickstrom et al \cite{wickstrom2020uncertainty} test the possibility of combining MC-Dropout with GBP to generate uncertainty in the explanation. 
\par 
This work tests additional combinations of uncertainty estimation approaches and explanation methods systematically to produce explanation uncertainty. In the next sections, we discuss this in detail.

\section{Explanation Uncertainty Pipeline}
\label{sec:explanation_uncertainty_pipeline}

In this section, we discuss the proposed approach to compute the explanation uncertainty. To this end, we introduce the notation and the relevant underlying concepts. We also discuss the implementation of the pipeline along with providing the rationale for the selection of uncertainty estimation and explanation methods.

\subsection{Explanation Uncertainty}
Assume that a neural network $f_{\theta}$ having parameters $\theta$ is trained using uncertainty estimation methods such as MC-Dropout, Deep Ensemble, and Variational Inference among others. The inference for such a model involves multiple stochastic forward passes (MC-Dropout, Variation Inference), or multiple models (Deep Ensembles). This is mathematically formulated as:

\begin{align}\label{mean_of_multiple_stochastic_forward_passes}
    \mu(x) = T^{-1} \sum_i f_\theta(x),  && 
    \sigma^2(x) = T^{-1} \sum_i (f_\theta(x) - \mu(x))^2
\end{align}

where $T$ denotes the number of stochastic forward passes or constituent ensemble models, $\mu$ and $\sigma^{2}$ denote the mean and the variance of the outcome of stochastic forward passes/ensemble models. The gradient-based saliency explanations can then generally be represented as:

\begin{equation}\label{representing_gradient_based_explanations}
    \text{E}(x) = F\left(\hat{y}, \frac{\partial f_\theta(x)}{\partial x}\right)
\end{equation}

where $E(x)$ denotes the gradient-based saliency explanation, $\hat{y}$ is the network prediction and $x$ is the input. 
\par
To generate an explanation with uncertainty, we combine the previous two mathematical formulations as follows:

\begin{equation}
    \text{E}_\mu(x) = T^{-1} \sum_i \text{E}(x) = T^{-1} \sum_i F\left(\hat{y}, \frac{\partial f_\theta(x)}{\partial x}\right)
\end{equation}
\begin{equation}
   \text{E}_\sigma(x) = T^{-1} \sum_{i} ( F\left(\hat{y}, \frac{\partial f_\theta(x)}{\partial x}\right) - \text{E}_\mu(x))^2
\end{equation}

where $\text{E}_{\mu}(x)$ denotes the explanation mean and $\text{E}_{\sigma}(x)$ is the standard deviation of the explanations generated by outputs of multiple stochastic forward passes/models. If the explanations are drastically different, the standard deviation will be high. However, if the explanations are similar, the standard deviation will be low.
Additionally, we compute the coefficient of variation, a metric that combines the standard deviation and the mean to concisely represent a distribution. It is defined as:
\begin{equation}\label{coefficient_of_variation_equation}
    \text{CV}(x) = \frac{\text{E}_\sigma(x)}{\text{E}_\mu(x)}
\end{equation}
where $\text{E}_\sigma(x)$ and $\text{E}_\mu(x)$ are the standard deviation and mean explanations respectively. The higher the coefficient of variation, the more the dissimilar the individual explanation heatmaps are. This metric can be helpful to quantify the confidence in the explanation of the network output for particular input features.
Note that the explanation uncertainty is explicitly concerned with modeling uncertainty in the explanation and is unrelated to uncertainty explanation where the objective is to explain the uncertainty in a given prediction.

\subsection{Pipeline Implementation}
\begin{figure}[t]
\includegraphics[width=0.475\textwidth, height=1.75cm]{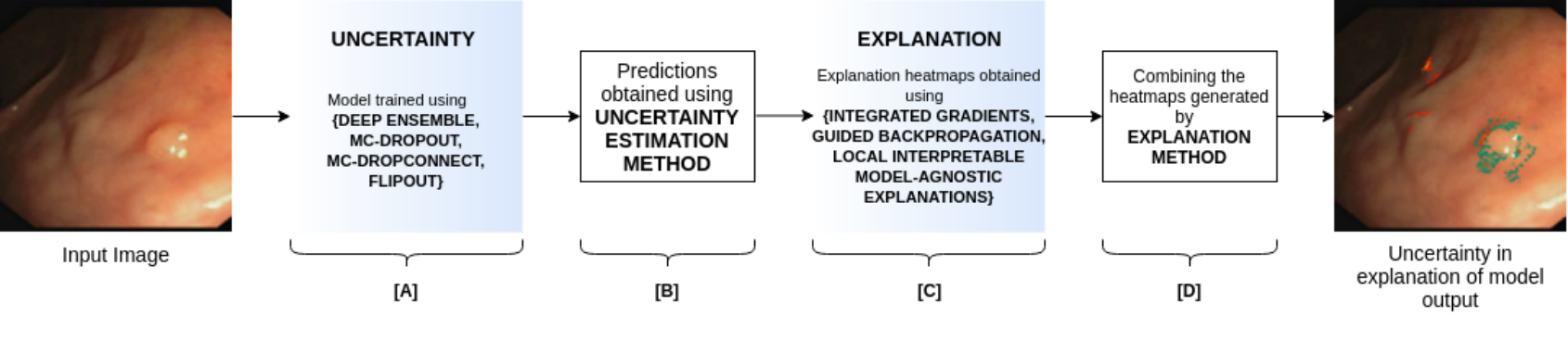}
\caption{Overview of our proposed approach to explore the underlying uncertainty associated with the explanation of neural network output. The demonstrated input and output images have been taken from \cite{wickstrom2020uncertainty} for the sake of representation. The labels [A], [B], [C], and [D] depict the different stages of this pipeline and these labels are used as a reference for the discussions in the text.} 
\label{proposed_approaches_uncertainty_in_explanation_methods}
\end{figure}

Figure \ref{proposed_approaches_uncertainty_in_explanation_methods} provides a visual representation of our proposed pipelines. The various stages in this pipeline are denoted by \textbf{[A], [B], [C], and [D]}. In stage \textbf{[A]}, we train a neural network using uncertainty estimation methods. This helps in generating an output distribution as visualized by \textbf{[B]}. Applying the explanation methods (denoted by \textbf{[C]}) to the samples of this output distribution then generates a distribution of explanation heatmaps as highlighted by \textbf{[D]}. 

\par
It is by analyzing samples of this explanation distribution that we will quantify the confidence in the generated explanation. We compute the mean of the heatmap samples of the explanation distribution. The mean heatmap would be highly activated for those features that predominantly influenced the model outcome. The higher the activation of a particular feature in the mean heatmap, the higher the confidence in that feature impacting the model decision-making process. Next, we compute the standard deviation heatmap of the samples from the explanation distribution. The standard deviation provides insights into the amount of uncertainty in the confidence of the feature's importance in the model decision-making. A higher standard deviation indicates higher uncertainty associated with the feature's importance. Finally as discussed previously, we also compute the coefficient of variation of this explanation distribution. The higher the coefficient of variation is, the higher the uncertainty associated with the importance of that feature in the explanation.

\subsection{Selection of Uncertainty and Explanation Methods}
From the literature survey in previous sections, we identify several uncertainty estimation methods and explanation methods to test in this paper. Here, we provide the rationale for selecting these approaches:

\textbf{Uncertainty Estimation methods}. We identify a total of 4 uncertainty estimation approaches to test in our pipeline: \textit{Deep Ensemble, MC-Dropout, MC-DropConnect, Flipout}.

\textbf{Explanation methods}. We select 3 explanation methods to test our proposed approach: \textit{Guided Backpropagation, Integrated Gradients, Local Interpretable Model-agnostic Explanations} (LIME).

These methods are selected by taking into consideration the complexity of their implementation, the amount of compute necessary to generate the output using these methods, and the availability of literature supporting their performance.

\section{Experimental Setup}
\label{sec:experimental_setup}

\subsection{Image Classification Experiments} 
We test the proposed approach for the image classification task using the CIFAR-10 and the FER+ datasets. For this, we use the \textit{\{Deep Ensemble, MC-Dropout, MC-DropConnect, Flipout\}} as uncertainty estimation methods and \textit{\{Guided Backpropagation, Integrated Gradients\}} as the explanation methods. This provides us with a total of eight combinations of uncertainty estimation and explanation methods that we test. It should be noted that the LIME has not been used along with the image classification task as it is computationally expensive.

The neural network used is a variant of miniVGG discussed in \cite{valdenegro2022deeper}. The modified network has three sequential blocks each consisting of a \textit{convolution} layer followed by \textit{batch normalization} and a \textit{max pooling layer}. The output of these blocks is supplied to a \textit{flatten} layer and subsequently to two \textit{dense} layers. The number of units in the final \textit{dense} layer of this architecture has 10 and 8 units for CIFAR-10 and FER+ respectively as these are the total number of different classes available in these datasets.

After training the neural network with uncertainty estimation methods, we can generate multiple output predictions for a single input. Each of these predictions is a sample from the output distribution. As the next step, we apply the selected explanation methods to these output samples, resulting in an explanation heatmap per output sample. The size of each explanation heatmap is the same as our original input image. We visualize the mean, the standard deviation, and the coefficient of variation of these explanation distribution samples. As these quantities are also computed on a per-pixel basis, the heatmaps for these quantities will also be of the same size as the original input image.

\textbf{Pixel Deletion/Insertion}. We focus on quantifying the quality of the generated explanations using our proposed approach. To this end, we identify the \textit{pixel deletion} and the \textit{pixel insertion} metric proposed by \cite{petsiuk2018rise} in their work. The objective is to compute the changes in the predicted class score by the subsequent addition/removal of pixels to/from an image. We plot this change in the class score against the number of pixels added/deleted and compute the area under this curve (AUC) which is the pixel insertion/deletion metric. The addition/removal of the pixels to/from the input image is governed by the explanation heatmap associated with the image under consideration.
\par
To compute the pixel deletion metric, we begin with the original input image and record the decrease in the class score as a function of number of pixels deleted from the original image. The order of pixel deletion is based on the importance of the pixel in the explanation heatmap of the input image. The most important pixels are deleted first and the class score is observed. This step is repeated till all the pixels in the input image are deleted. The pixel deletion metric is then defined as the AUC obtained by plotting the class score as a function of \% of pixels deleted.
\par
Conversely, to compute the pixel insertion metric, we begin with a noisy image and observe the increment in the class score as image pixels are gradually de-noised from this image based on their importance in the explanation heatmap. The process of adding pixels is similar to the one adopted while calculating the deletion metric. The most important pixels are used to de-noise the image first. This process is repeated till the entire image is de-noised and the corresponding class score is observed. The pixel insertion metric is then the AUC obtained by plotting the class score as a function of the \% of pixels inserted.
\par
It should be noted that a lower AUC for the pixel deletion corresponds to a good explanation as the removal of highly relevant pixels should diminish the class score significantly in the initial stages of the pixel removal process. Conversely, a higher AUC for the pixel insertion translates to a good explanation as the addition of highly important pixels early in the pixel insertion process should boost the class scores.
\par
In this work, we modify the originally proposed pixel deletion/insertion metrics to suit our requirements. As opposed to computing the pixel deletion and pixel insertion directly on individual explanation heatmaps, we propose to compute these metrics on the mean and the standard deviation heatmap representation of the explanation distribution. This helps us to quantify the quality of these distributions. The insertion and removal of the pixels are governed by the importance of the pixel in the mean and the standard deviation heatmaps. Highly weighted pixels in the mean and the standard deviation heatmap are added/removed first. Additionally, we compute these metrics on batches of images belonging to the same class rather than computing on a per-image level. This provides a measure of the quality of explanations on a per-class basis.

\begin{figure}[t]
\includegraphics[width=0.475\textwidth]{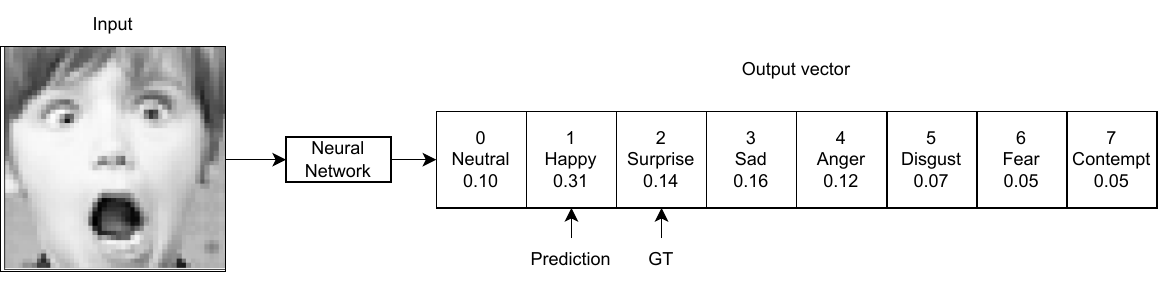}
\caption{Illustration describing an alternative approach to compute gradients and generate explanations. The original approach generates explanations by computing the gradient of the network prediction logit with respect to the input image whereas, in the modified variant, we compute the gradient of the ground truth logit with respect to the input image.}
\label{experiment_3_gradient_computation_step}
\end{figure}

\textbf{Prediction vs Ground Truth Neurons}. We propose modifying the computation of explanation uncertainty by varying the gradient computation step in our original approach. This modification is only tested with the FER+ dataset and is explained with the help of Figure \ref{experiment_3_gradient_computation_step}. 

In our approach, the explanation is generated by calculating the gradient of the network output with respect to the input image. Mathematically, this is formulated as:
\begin{align}
\label{experiment_1_equation} 
        E(x) = \frac{\partial (\text{Output}_\text{Pred})}{\partial (\text{Input}\ \text{Image})}, %
        \,      
        E'(x) = \frac{\partial (\text{Output}_{\text{GT}})}{\partial (\text{Input\ Image)}}
\end{align}
where $\text{Output}_\text{Pred}$ is the same as network prediction from Figure \ref{experiment_3_gradient_computation_step}. The explanation generated from this calculation helps us understand \textit{why} the network predicted what it predicted. 

\par
However, now we modify the gradient computation step as can be seen in the Equation \ref{experiment_1_equation}, where $\text{Output}_{\text{Gt}}$ is the activation of the unit corresponding to the ground truth (GT) as depicted in Figure \ref{experiment_3_gradient_computation_step}. The explanation ($E'(x)$) generated using this calculation helps to visualize the features/image regions that the network \textit{should have} focused on for its prediction to match the ground truth. It is important to note that formulations listed in Equation \ref{experiment_1_equation} converge when the network prediction matches the ground truth for a given input.

\par
In the case of the FER+ dataset, we observed that the network prediction often does not match the ground truth owing to the inherent nature of the dataset. Hence, we propose modifying the approach to compute the explanations. The process of generating the mean, the standard deviation heatmap, and the coefficient of variation heatmap is identical to our original approach.

\subsection{Regression Experiments}
We test our approach with tabular regression as it allows us to test its robustness. For this, we use the \textit{\{Deep Ensemble, MC-Dropout, MC-DropConnect, Flipout\}} as uncertainty estimation methods and \textit{\{Guided Backpropagation, Local Interpretable Model-agnostic Explanation\}} as the explanation methods.
We select the LIME method as it can be applied to tabular datasets. Additionally, to the best of our knowledge, the GBP method has not been used to generate explanations for tabular data. We use a Multi-Layer Perceptron with 4 \textit{dense} layers for this task.

\section{Experimental Results}
\label{sec:experimental_results}

\begin{figure}[t]
    \centering
    \begin{varwidth}{0.5\textwidth}
    
    \begin{tabular}
    {
    p{0.001\textwidth}
    p{0.1\textwidth} %
    p{0.1\textwidth}
    p{0.1\textwidth}
    p{0.1\textwidth}
    p{0.1\textwidth}
    p{0.1\textwidth} %
    p{0.1\textwidth} %
    }
    \multicolumn{1}{p{0.001\textwidth}}{\textbf{}}   
    &
    \multicolumn{1}{p{0.105\textwidth}}{} 
    &
    \multicolumn{1}{p{0.105\textwidth}}{}   
    &
    \multicolumn{1}{p{0.105\textwidth}}{}  &
    \multicolumn{1}{p{0.105\textwidth}}{}
    &
    \multicolumn{1}{p{0.105\textwidth}}{}  
    &
    \multicolumn{1}{p{0.105\textwidth}}{} &
    \multicolumn{1}{p{0.105\textwidth}}{} 
    \\

    \rotatebox{90}{\centering \fontsize{6}{4} \textbf{Input}}
    &
    \includegraphics[width=1cm, height=1cm]
    {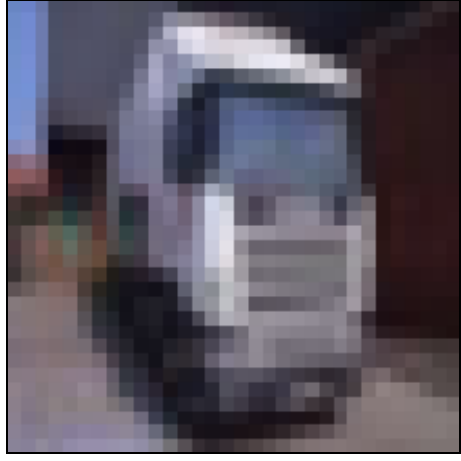}
    & 
    \rotatebox{90}{\centering \textbf{}}
    & 
    \rotatebox{90}{\centering \textbf{}}
    & 
    \rotatebox{90}{\centering \textbf{}}
    &
    \rotatebox{90}{\centering \textbf{}}
    &  
    \rotatebox{90}{\centering \textbf{}}
    &
    \rotatebox{90}{\centering \textbf{}}
    \\

    \multicolumn{1}{p{0.001\linewidth}}{}   
    &
    \multicolumn{1}{p{0.105\linewidth}}{\fontsize{6}{4} \selectfont \centering \textbf{$E_\mu$ IG}} 
    &
    \multicolumn{1}{p{0.105\linewidth}}{\fontsize{6}{4} \selectfont \centering \textbf{$E_\sigma$ IG}}   
    &
    \multicolumn{1}{p{0.105\linewidth}}{\fontsize{6}{4} \selectfont\centering \textbf{CV IG}}  &
    \multicolumn{1}{p{0.105\linewidth}}{\fontsize{6}{4} \selectfont \centering \textbf{$E_\mu$ GBP}}
    &
    \multicolumn{1}{p{0.105\linewidth}}{\fontsize{6}{4} \selectfont \centering \textbf{$E_\sigma$ GBP}}  
    &
    \multicolumn{1}{p{0.105\linewidth}}{\fontsize{6}{4} \selectfont \centering \textbf{CV GBP}}
    \\

    \rotatebox{90}{\centering \textbf{\fontsize{6}{4} \selectfont Ensemble}}
    & 
    \includegraphics[width=1cm, height=1cm]{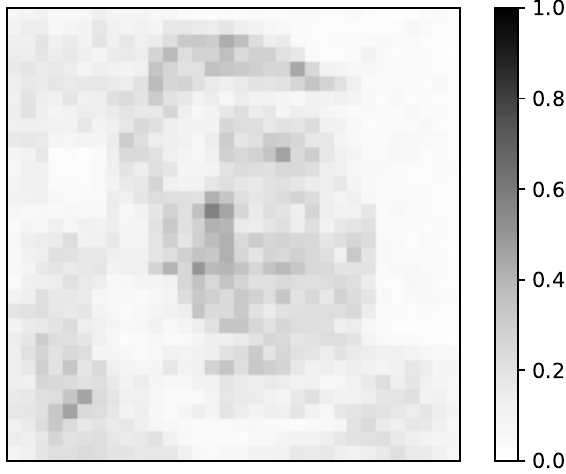}
    & 
    \includegraphics[width=1cm, height=1cm]{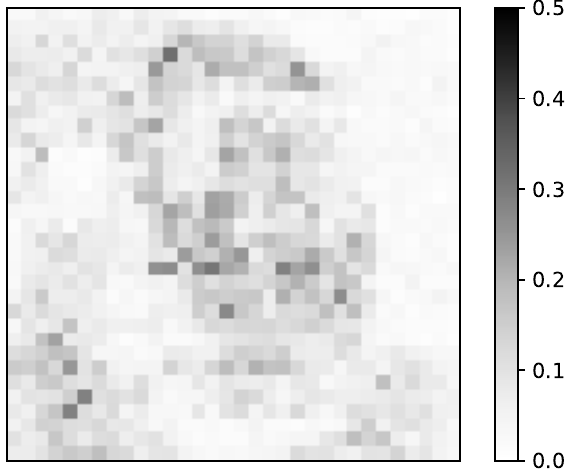}
    & 
    \includegraphics[width=1cm, height=1cm]{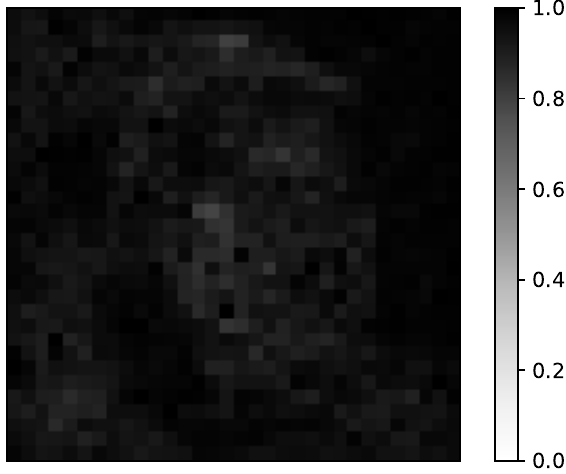}
    &
    \includegraphics[width=1cm, height=1cm]{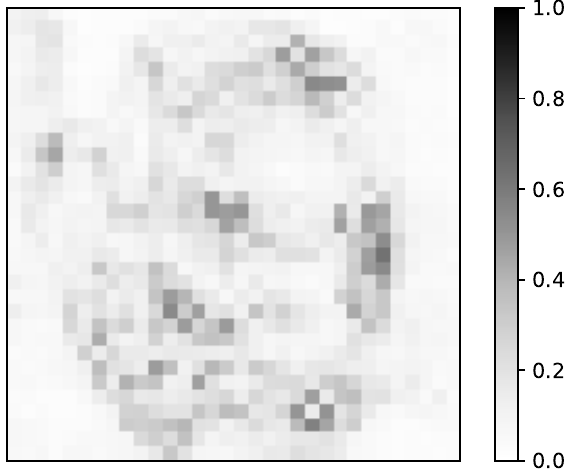}
    &  
    \includegraphics[width=1cm, height=1cm]{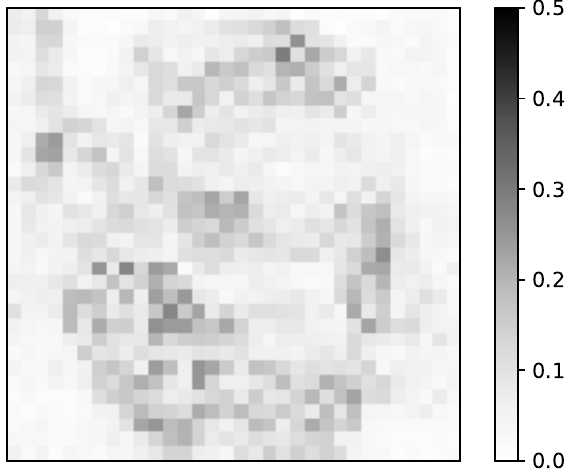}
    &
    \includegraphics[width=1cm, height=1cm]{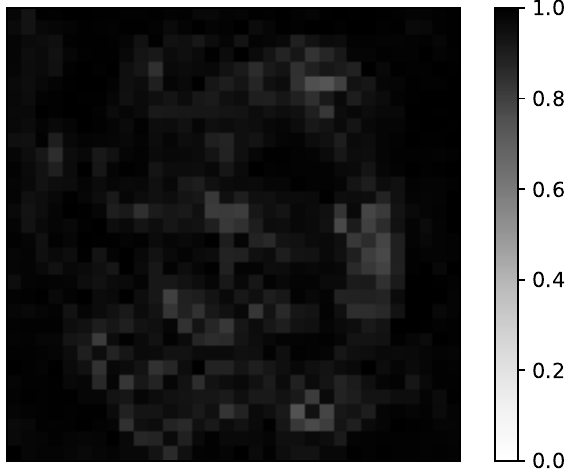}
    \\

    \rotatebox{90}{\centering \textbf{\fontsize{6}{4} \selectfont MC-D}}
    & 
    \includegraphics[width=1cm, height=1cm]{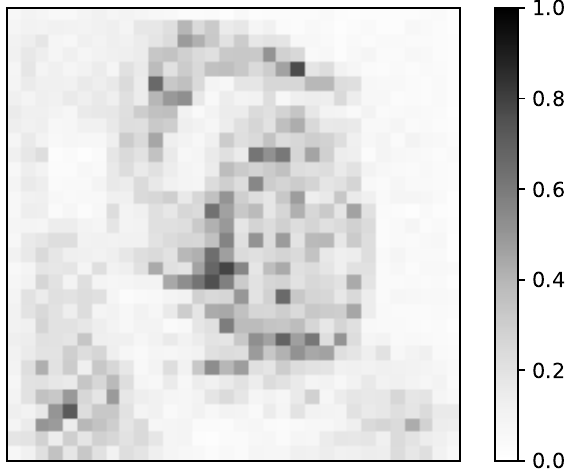}
    & 
    \includegraphics[width=1cm, height=1cm]{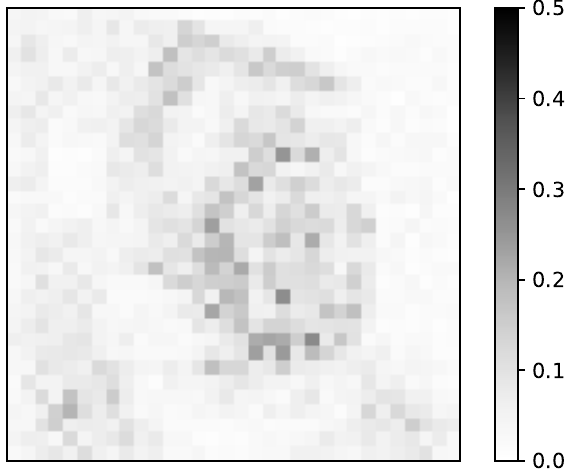}
    & 
    \includegraphics[width=1cm, height=1cm]{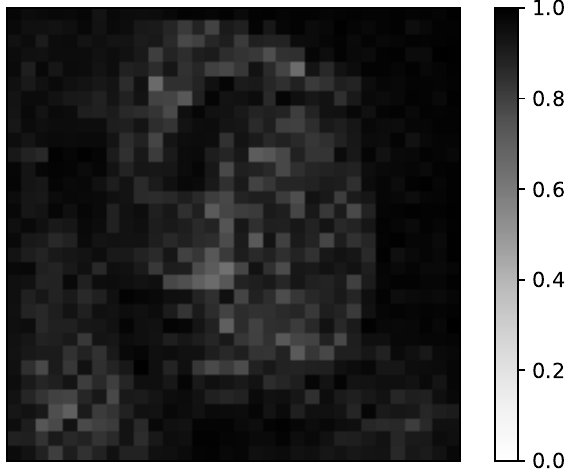}
    &
    \includegraphics[width=1cm, height=1cm]{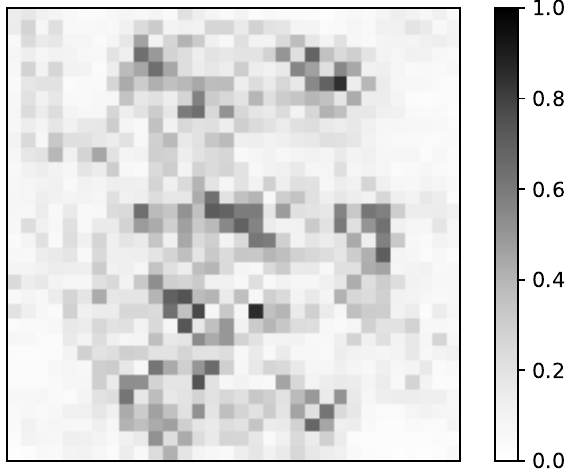}
    &  
    \includegraphics[width=1cm, height=1cm]{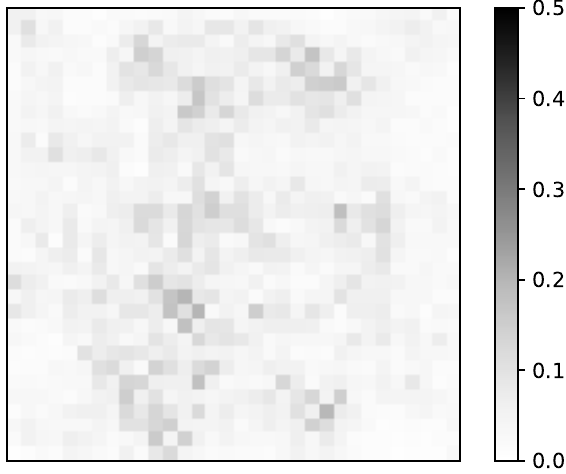}
    &
    \includegraphics[width=1cm, height=1cm]{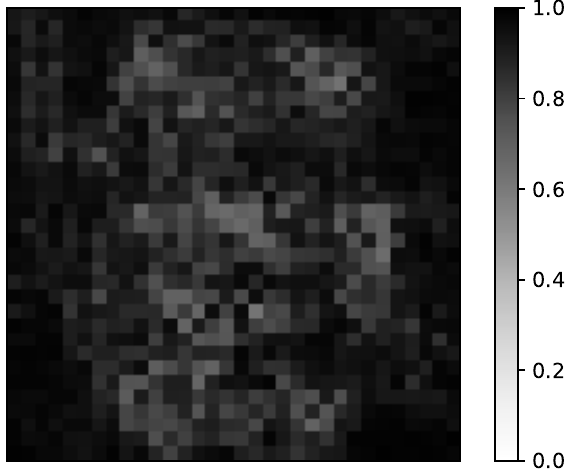}
    \\

    \rotatebox{90}{\centering \textbf{\fontsize{6}{4} \selectfont MC-DC}}
    & 
    \includegraphics[width=1cm, height=1cm]{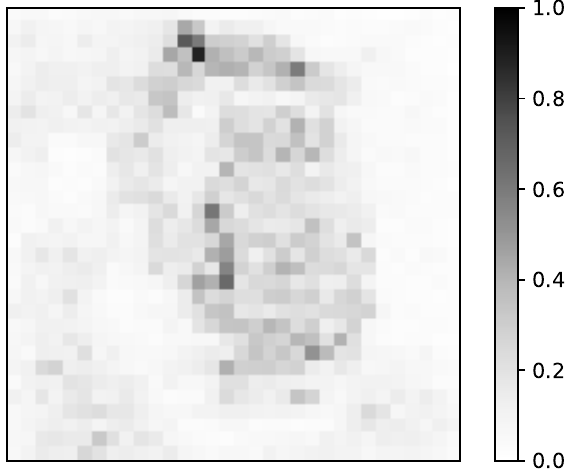}
    & 
    \includegraphics[width=1cm, height=1cm]{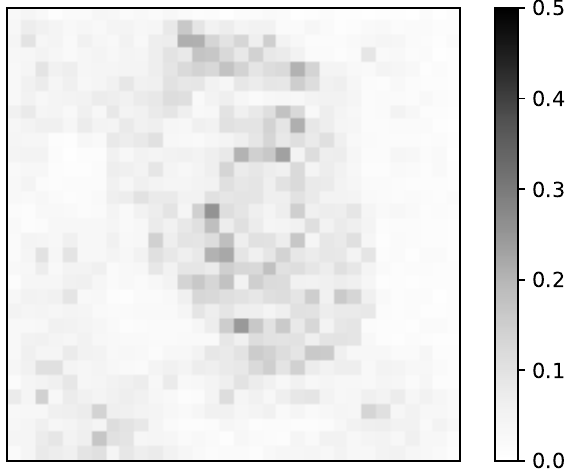}
    & 
    \includegraphics[width=1cm, height=1cm]{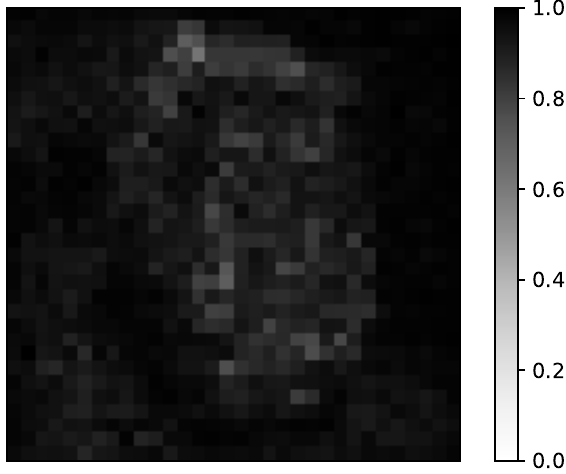}
    &
    \includegraphics[width=1cm, height=1cm]{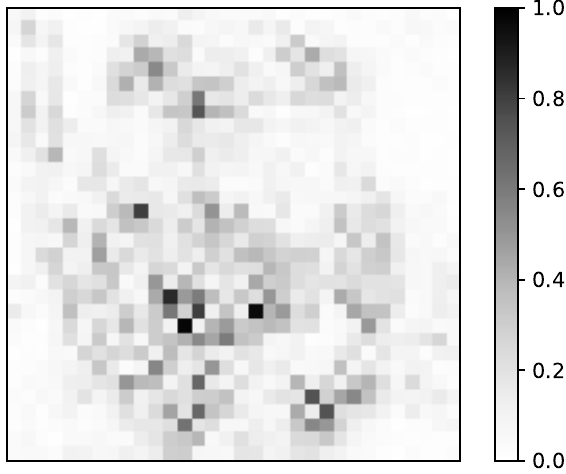}
    &  
    \includegraphics[width=1cm, height=1cm]{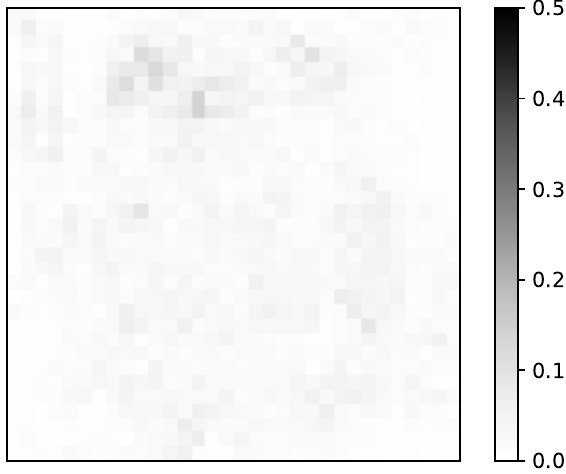}
    &
    \includegraphics[width=1cm, height=1cm]{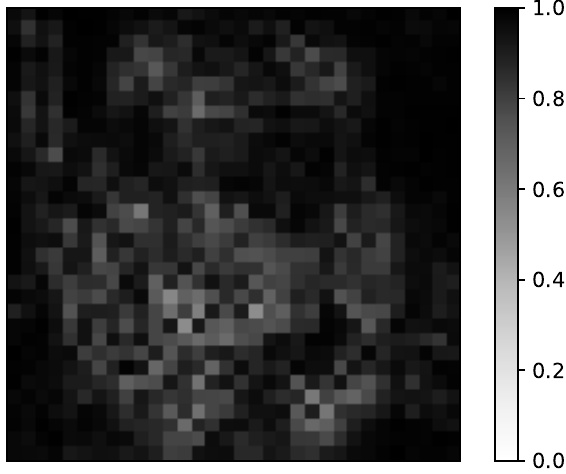}
    \\

    \rotatebox{90}{\centering \textbf{\fontsize{6}{4} \selectfont Flipout}}
    & 
    \includegraphics[width=1cm, height=1cm]{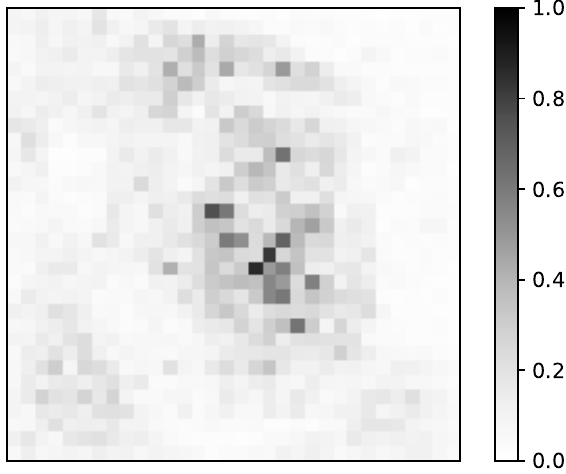}
    & 
    \includegraphics[width=1cm, height=1cm]{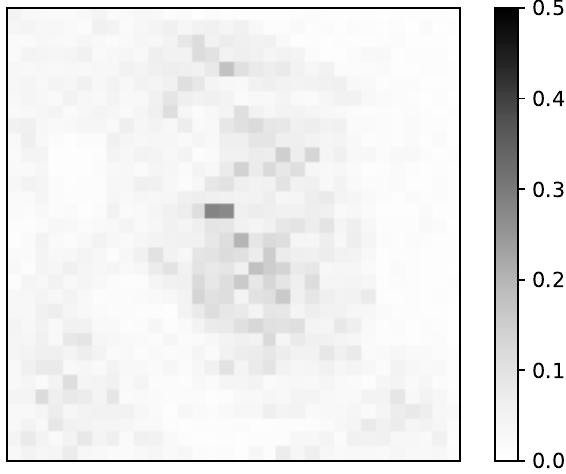}
    & 
    \includegraphics[width=1cm, height=1cm]{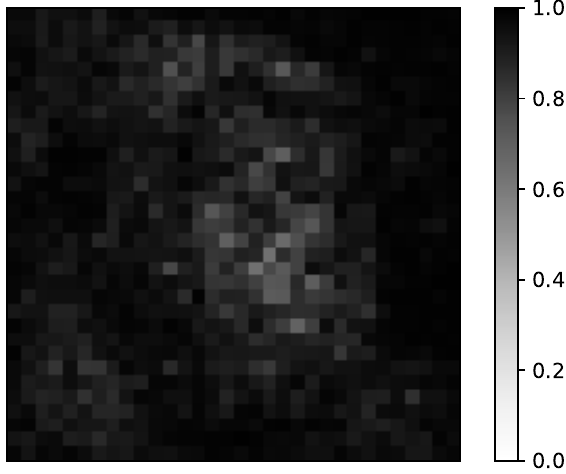}
    &
    \includegraphics[width=1cm, height=1cm]{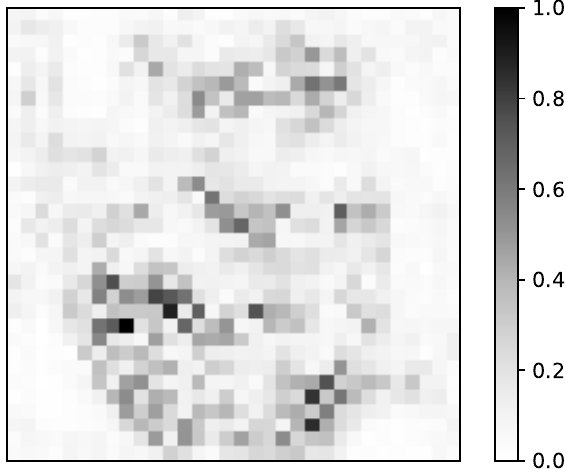}
    &  
    \includegraphics[width=1cm, height=1cm]{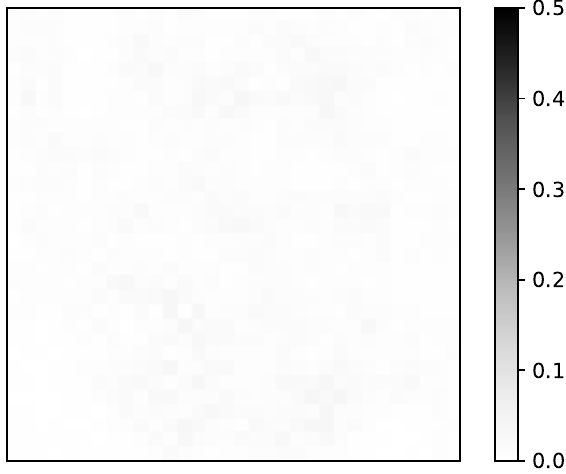}
    &
    \includegraphics[width=1cm, height=1cm]{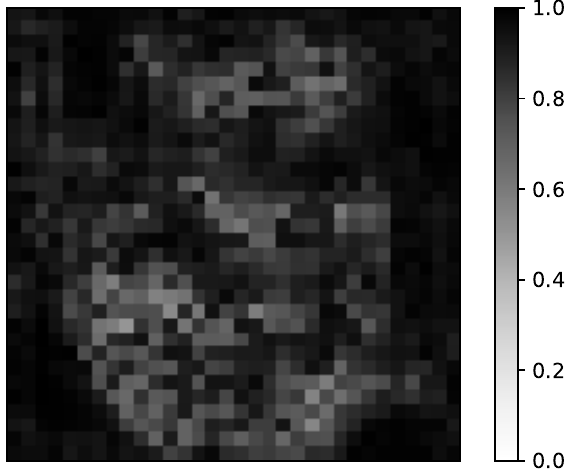}
    \\

    \end{tabular}

    \captionsetup{justification=centering}
    \caption{Visualization of the mean \textbf{($E_\mu$)}, the standard deviation \textbf{($E_\sigma$)}, and the coefficient of variation \textbf{($CV$)} heatmaps generated using the combinations of uncertainty estimation methods (Deep Ensemble, MC-Dropout, MC-DropConnect and Flipout) and explanation methods (Guided Backpropagation and Integrated Gradients). The uncertainty estimation methods are organized in individual rows and the explanation methods are aligned along the columns with the columns 1-3 representing the $\mu$, $\sigma$, and the $CV$ heatmaps for IG and the columns 4-6 representing the same for GBP.}
    \label{cifar_experiment_example_1}
    \end{varwidth}
\end{figure}  

\subsection{Image Classification Results} \label{explanation_uncertainty_for_image_classification_result}
The explanation heatmaps for an image from the CIFAR-10 dataset are visualized in Figure \ref{cifar_experiment_example_1}. As previously described, the size of these heatmaps is the same as the size of the input image. 
In the subsequent discussions, the heatmaps will be referred to by their identifier namely the mean explanation heatmap as $\mu$, the standard deviation explanation heatmap as $\sigma$, and the coefficient of variation explanation heatmap as $CV$.
\par
Figure \ref{cifar_experiment_example_1} depicts the image of a \textit{truck} taken from the CIFAR-10. We observe from the $\mu$ heatmaps that the explanations generated by GBP are highly activated for relevant pixel regions as compared to those by IG. We also observe
low activation in the $\sigma$ heatmaps generated by the GBP as compared to IG. This implies that the explanations generated by the GBP have lower uncertainty as compared to those generated by IG. 

\begin{table*}[t]
\centering
    \begin{tabular}{lllllllllll}		
    \toprule
        &           & \multicolumn{4}{c}{\textbf{CIFAR10}} & & \multicolumn{4}{c}{\textbf{FER+}}\\
        \midrule
		&			& \multicolumn{2}{c}{\textbf{Insert} $\uparrow$} & \multicolumn{2}{c}{\textbf{Delete} $\downarrow$} & & \multicolumn{2}{c}{\textbf{Insert}  $\uparrow$} & \multicolumn{2}{c}{\textbf{Delete} $\downarrow$}\\
        \midrule
		&			& $E_\mu$ & $E_\sigma$ & $E_\mu$ & $E_\sigma$ & & $E_\mu$ & $E_\sigma$ & $E_\mu$ & $E_\sigma$\\
		GBP & Ensemble 	& \ccainscifar{0.537} & \ccainscifar{0.534} & \ccadelcifar{0.554} & \ccadelcifar{0.540} & & \ccainsferplus{0.263} & \ccainsferplus{0.268} & \ccadelferplus{0.333} & \ccadelferplus{0.324}\\
        GBP & MC-Dropout 	& \ccainscifar{0.551} & \ccainscifar{0.541} & \ccadelcifar{0.320} & \ccadelcifar{0.323} & & \ccainsferplus{0.354} & \ccainsferplus{0.347} & \ccadelferplus{0.293} & \ccadelferplus{0.281}\\
        GBP & MC-DropConnect 	& \ccainscifar{0.582} & \ccainscifar{0.571} & \ccadelcifar{0.368} & \ccadelcifar{0.354} & & \ccainsferplus{0.186} & \ccainsferplus{0.187} & \ccadelferplus{0.256} & \ccadelferplus{0.253}\\
        GBP & Flipout 	& \ccainscifar{0.586} & \ccainscifar{0.557} & \ccadelcifar{0.301} & \ccadelcifar{0.299} & & \ccainsferplus{0.291} & \ccainsferplus{0.289} & \ccadelferplus{0.271} & \ccadelferplus{0.274}\\
        \midrule
		IG & Ensemble	& \ccainscifar{0.547} & \ccainscifar{0.542} & \ccadelcifar{0.606} & \ccadelcifar{0.571}& & \ccainsferplus{0.304} & \ccainsferplus{0.304} & \ccadelferplus{0.364} & \ccadelferplus{0.361}\\			
		IG & MC-Dropout	& \ccainscifar{0.582} & \ccainscifar{0.571} & \ccadelcifar{0.368} & \ccadelcifar{0.345} & & \ccainsferplus{0.374} & \ccainsferplus{0.378} & \ccadelferplus{0.311} & \ccadelferplus{0.301}\\			
		IG & MC-DropConnect	& \ccainscifar{0.585} & \ccainscifar{0.562} & \ccadelcifar{0.310} & \ccadelcifar{0.310} & & \ccainsferplus{0.271} & \ccainsferplus{0.266} & \ccadelferplus{0.239} & \ccadelferplus{0.244}\\			
		IG & Flipout	& \ccainscifar{0.598} & \ccainscifar{0.589} & \ccadelcifar{0.334} & \ccadelcifar{0.337} & & \ccainsferplus{0.328} & \ccainsferplus{0.341} & \ccadelferplus{0.308} & \ccadelferplus{0.323}\\
    \bottomrule
	\end{tabular}
    \caption{Area under the Insertion/Deletion curves, average across all classes, for CIFAR10 and FER+ datasets, computed separately for explanation mean $E_\mu$ and explanation uncertainty $E_\sigma$. The up/down arrows indicate direction of best metric value. }
    \label{classif_results_table}
\end{table*}

\par
Similar analysis can be conducted for images from the FER+ dataset. The explanation distribution obtained for an image taken from FER+ are presented in the first four rows of Figure \ref{fer_variant_experiment_example_1}. It can be observed in the $\mu$ heatmaps that the GBP highlights the relevant facial features to a greater extent as compared to IG. This observation is supported by the $\sigma$ heatmaps of both these methods as well. The $\sigma$ and the $CV$ heatmaps generated by the GBP have low activation around the relevant pixel regions for all the uncertainty estimation methods. This implies that the explanations generated by the GBP have a higher agreement and are therefore highly confident. However, this is not the case for IG where the pixel activations in the $\mu$ heatmaps are scattered. This indicates a low confidence in the generated explanation.

\par
From these figures, we also observe that IG generates more noisier explanations as compared to GBP. This could be attributed to the accumulation of activations from multiple steps while calculating explanations using IG that result in spurious highlighting of less relevant regions.

\textbf{Pixel Deletion/Insertion}. The results for pixel deletion/insertion are shown in Figure \ref{experiment_2_cde_cdo_cdc_cdf} (in appendix) and average AUC are shown in Figure \ref{classif_results_table}. The tables containing the AUCs for these plots are provided in the Supplementary Materials. It should be noted that a higher AUC for pixel insertion and a lower AUC for pixel deletion is desired to ensure good-quality explanations. The pixel insertion and deletion curves (see Supplementary Material) have been plotted for batches of images representing all the classes in both datasets (CIFAR-10 and FER+). The order of pixel insertion/deletion is determined using the mean and the standard deviation explanation heatmaps of the corresponding images from these datasets.

\par 
It is evident from these results that for CIFAR-10, the best explanations are generally generated for the \textit{frog} class and the worst explanations are generated for the \textit{dog} class. These observations are supported by high AUC for pixel insertion and low AUC for pixel deletion for these classes. Overall across uncertainty methods and datasets, Flipout seems to perform the best, while Ensembles seems to perform the worse, specially in deletion curve for CIFAR10, but the variation between area under insertion/deletion curve is small.

\par
In the case of the FER+, the trends are not generalizable. This can be attributed to the fact that the FER+ is a difficult dataset to train a network on resulting in a relatively poor quality of model predictions and by extension, poor quality of explanations.
This is also supported by the anomalous behavior of the pixel deletion plots for the case of Flipout. In these plots, the class score \textit{increases} despite relevant pixels being deleted from the image. 

\subsection{Prediction vs Ground Truth Neurons}
\label{explanation_uncertainty_prediction_vs_ground_truth_neurons}
\begin{figure*}[t]
    \centering

    \begin{tabular}{p{0.001\textwidth}p{0.055\textwidth}p{0.055\textwidth}p{0.055\textwidth}p{0.055\textwidth}p{0.055\textwidth}p{0.055\textwidth}p{0.055\textwidth}p{0.055\textwidth}p{0.055\textwidth}p{0.055\textwidth}p{0.055\textwidth}p{0.055\textwidth}
    }
    &
    \multicolumn{2}{l}{\includegraphics[width=1cm, height=1cm]{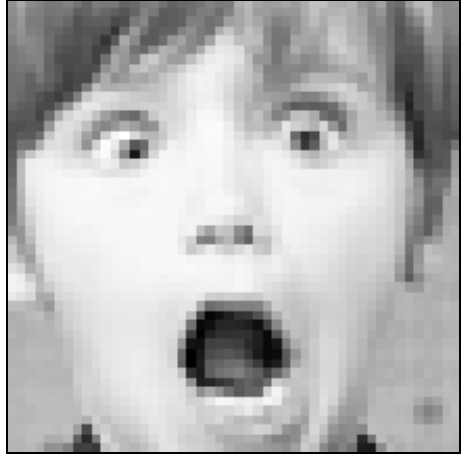}}
    &
    &
    &
    &
    &
    &
    &
    & 
    & 
    & 
    \\
    
     & \multicolumn{6}{l}{\footnotesize \textbf Explanation for Predicted Class} & \multicolumn{6}{l}{\footnotesize \textbf Explanation for Ground Truth Class}\\    
    \toprule
    &
    {\fontsize{4}{4} \selectfont \centering \textbf{$E_\mu$ IG}} 
    &
    {\fontsize{4}{4} \selectfont \centering \textbf{$E_\sigma$ IG}}   
    &
    {\fontsize{4}{4} \selectfont\centering \textbf{CV IG}}  &
    {\fontsize{4}{4} \selectfont \centering \textbf{$E_\mu$ GBP}}
    &
    {\fontsize{4}{4} \selectfont \centering \textbf{$E_\sigma$ GBP}}  
    &
    {\fontsize{4}{4} \selectfont \centering \textbf{CV GBP}}
    &
    {\fontsize{4}{4} \selectfont \centering \textbf{$E_\mu$ IG}} 
    &
    {\fontsize{4}{4} \selectfont \centering \textbf{$E_\sigma$ IG}}   
    &
    {\fontsize{4}{4} \selectfont\centering \textbf{CV IG}}  &
    {\fontsize{4}{4} \selectfont \centering \textbf{$E_\mu$ GBP}}
    &
    {\fontsize{4}{4} \selectfont \centering \textbf{$E_\sigma$ GBP}}  
    &
    {\fontsize{4}{4} \selectfont \centering \textbf{CV GBP}}
    \\

    \rotatebox{90}{\centering \textbf{\fontsize{4}{4} \selectfont Ensemble}}
    & 
    \includegraphics[width=1cm, height=1cm]{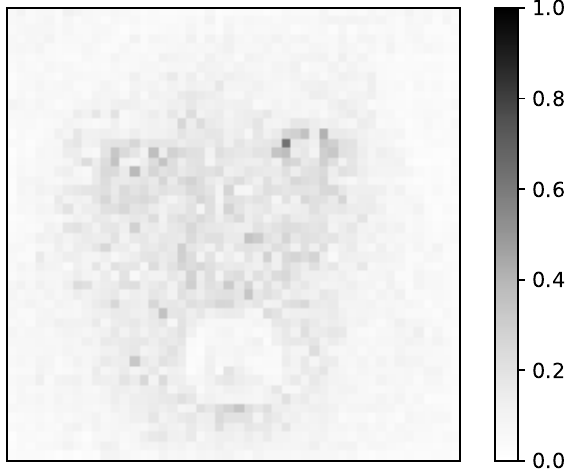}
    & 
    \includegraphics[width=1cm, height=1cm]{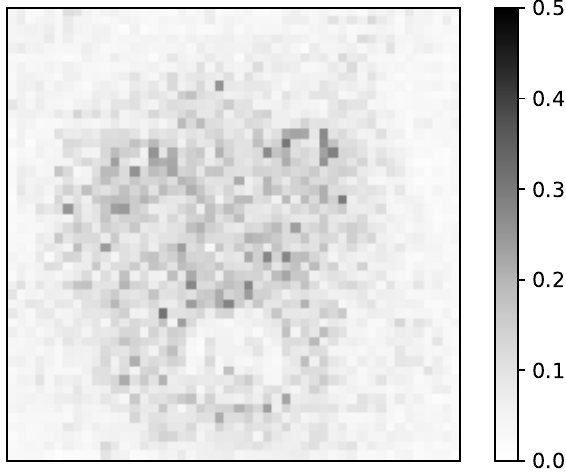}
    & 
    \includegraphics[width=1cm, height=1cm]{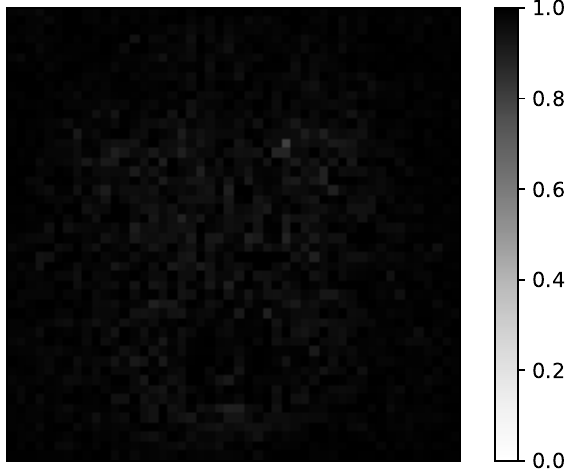}
    &
    \includegraphics[width=1cm, height=1cm]{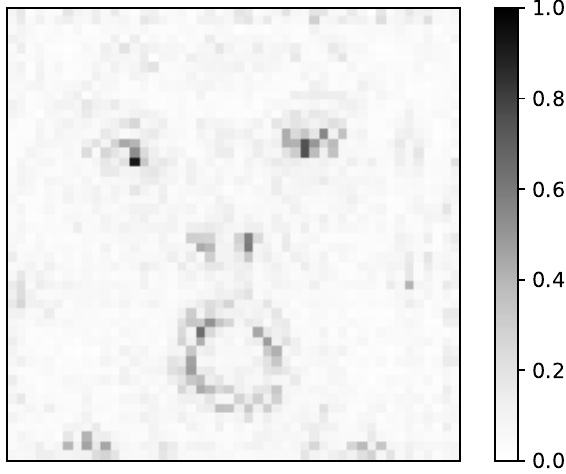}
    &  
    \includegraphics[width=1cm, height=1cm]{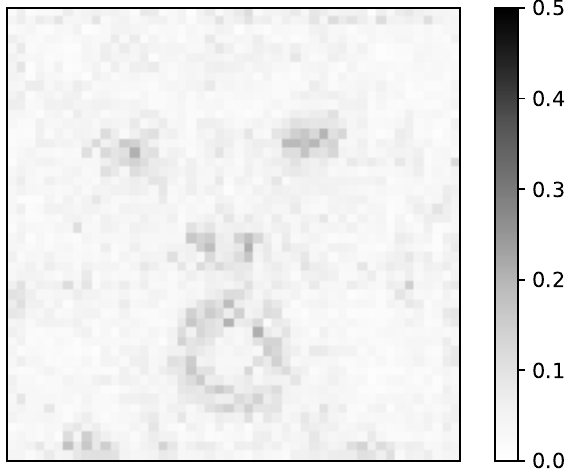}
    &
    \includegraphics[width=1cm, height=1cm]{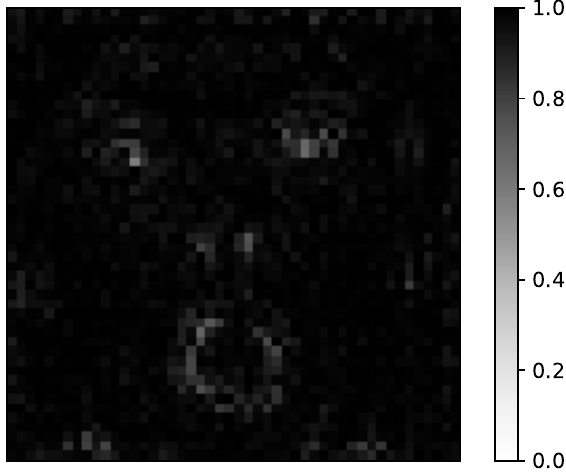}    
    &
    \includegraphics[width=1cm, height=1cm]{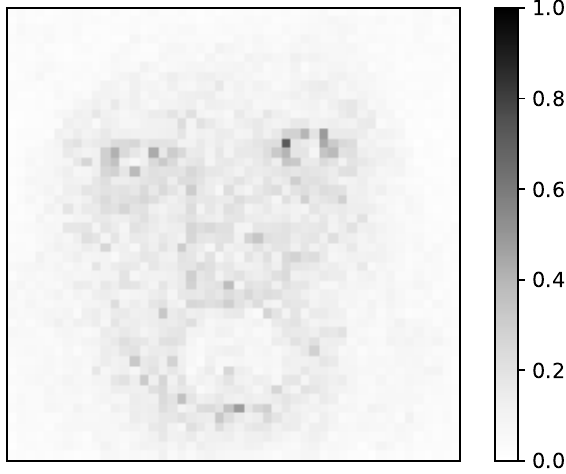}
    & 
    \includegraphics[width=1cm, height=1cm]{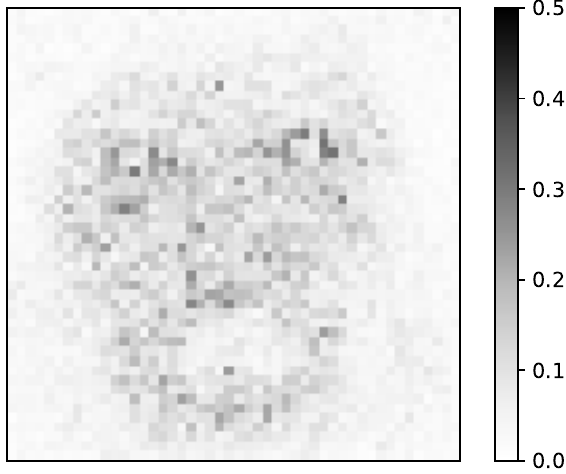}
    & 
    \includegraphics[width=1cm, height=1cm]{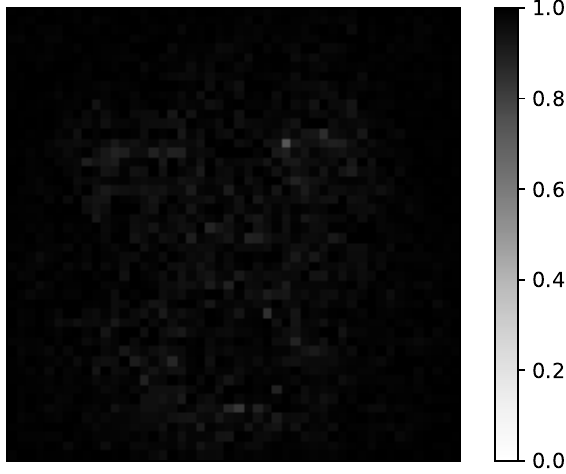}
    &
    \includegraphics[width=1cm, height=1cm]{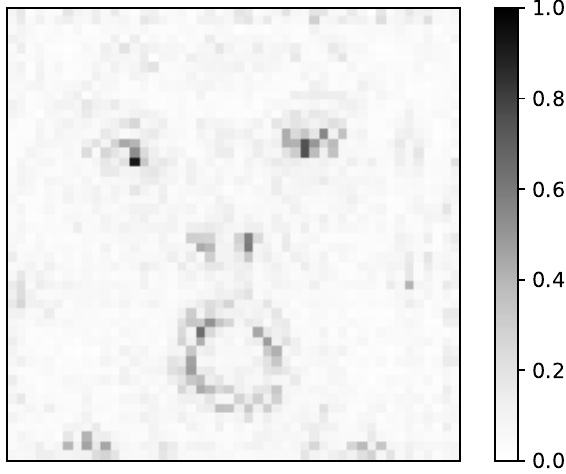}
    &  
    \includegraphics[width=1cm, height=1cm]{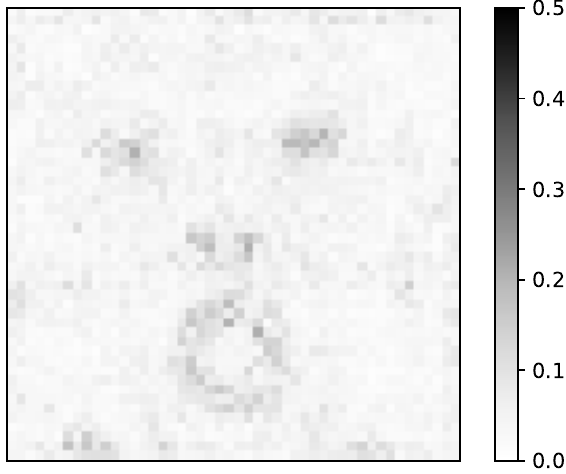}
    &
    \includegraphics[width=1cm, height=1cm]{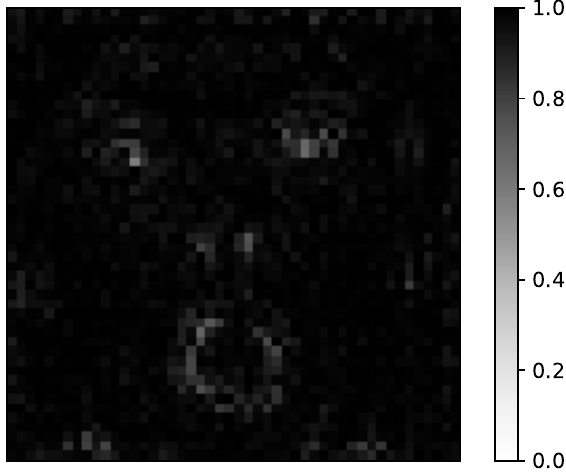}
    \\

    \rotatebox{90}{\centering \textbf{\fontsize{4}{4} \selectfont MC-D}}
    & 
    \includegraphics[width=1cm, height=1cm]{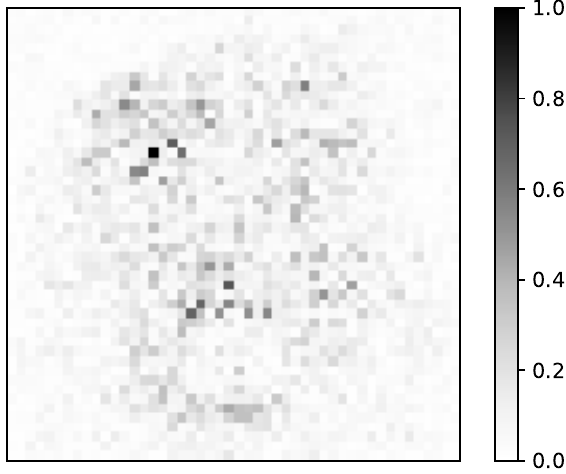}
    & 
    \includegraphics[width=1cm, height=1cm]{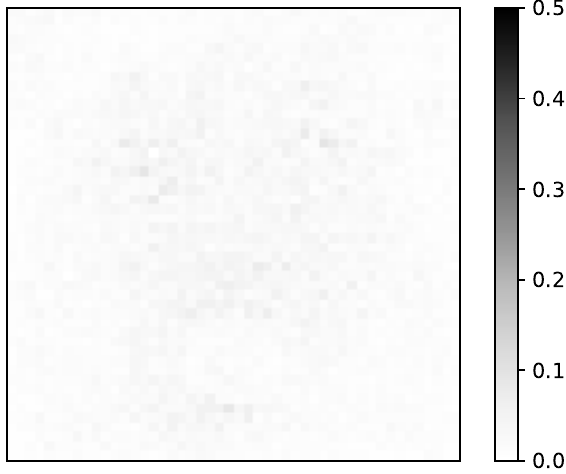}
    & 
    \includegraphics[width=1cm, height=1cm]{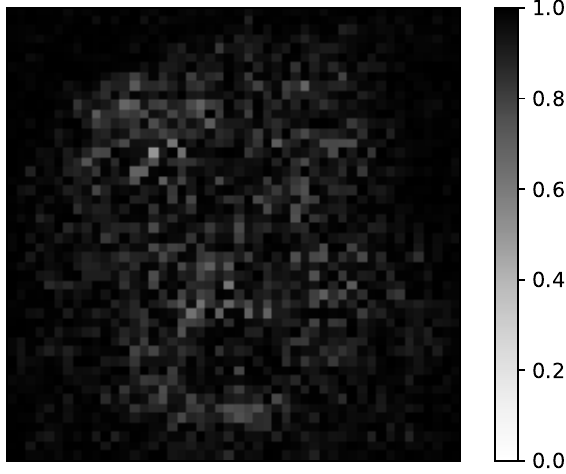}
    &
    \includegraphics[width=1cm, height=1cm]{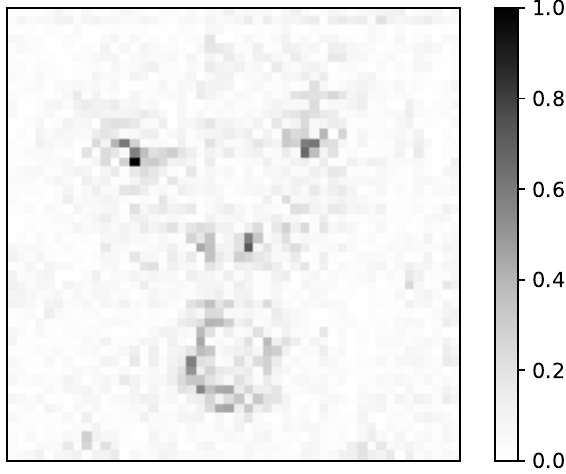}
    &  
    \includegraphics[width=1cm, height=1cm]{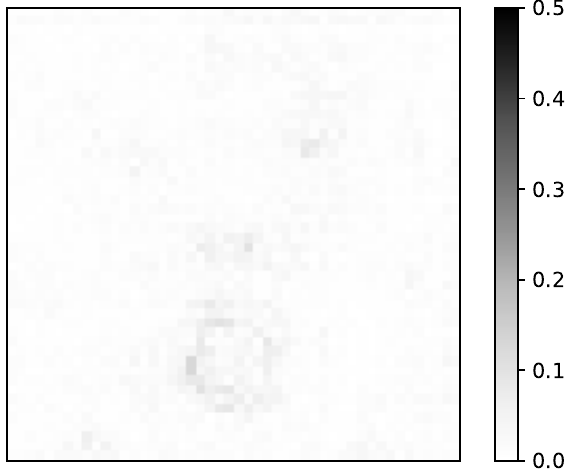}
    &
    \includegraphics[width=1cm, height=1cm]{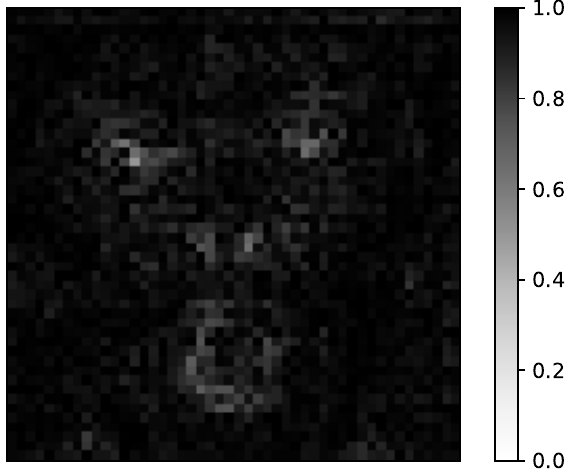}
    & 
    \includegraphics[width=1cm, height=1cm]{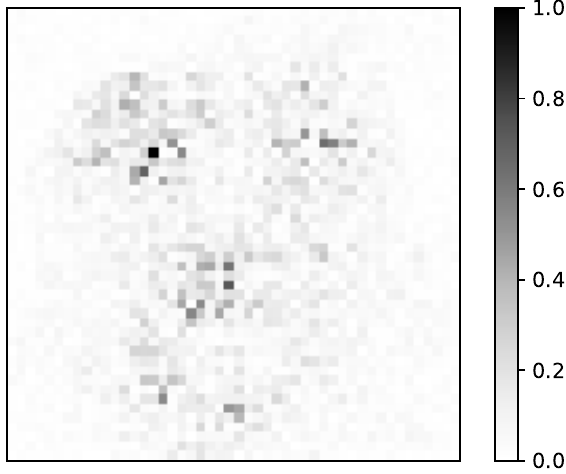}
    & 
    \includegraphics[width=1cm, height=1cm]{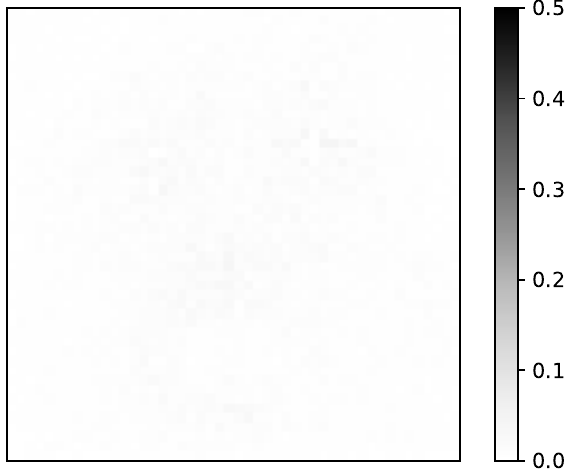}
    & 
    \includegraphics[width=1cm, height=1cm]{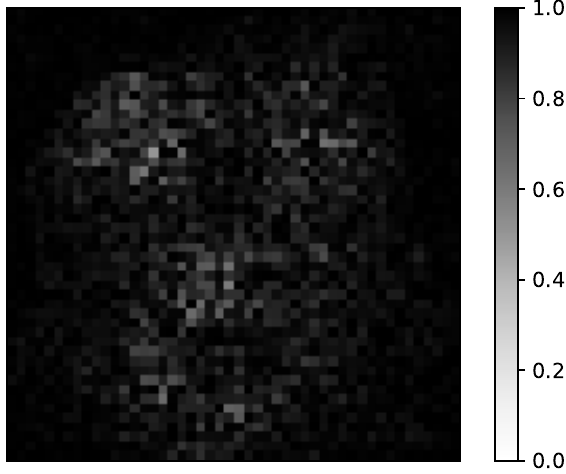}
    &
    \includegraphics[width=1cm, height=1cm]{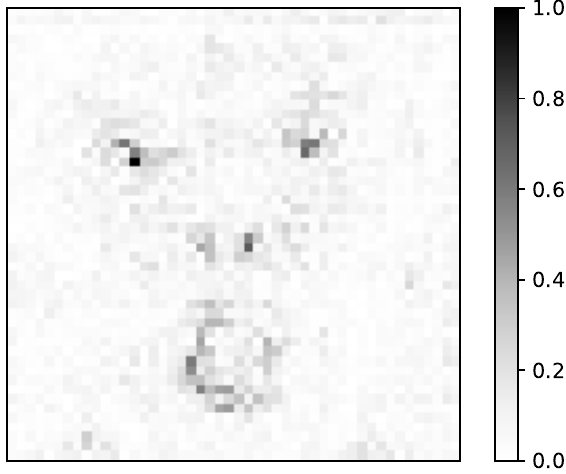}
    &  
    \includegraphics[width=1cm, height=1cm]{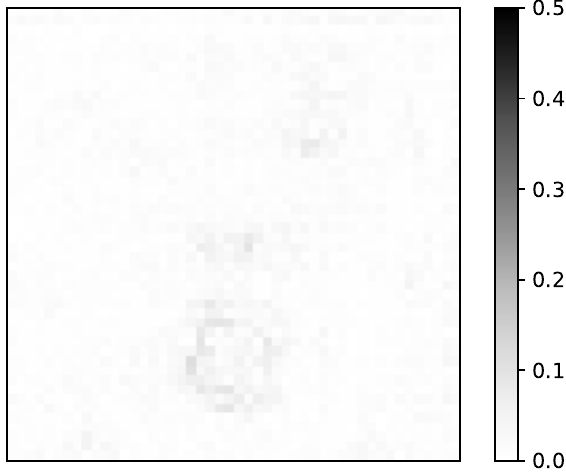}
    &
    \includegraphics[width=1cm, height=1cm]{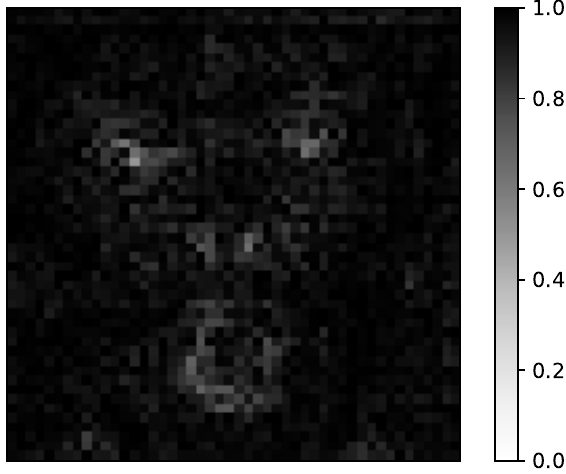}
    \\

    \rotatebox{90}{\centering \textbf{\fontsize{4}{4} \selectfont MC-DC}}
    & 
    \includegraphics[width=1cm, height=1cm]{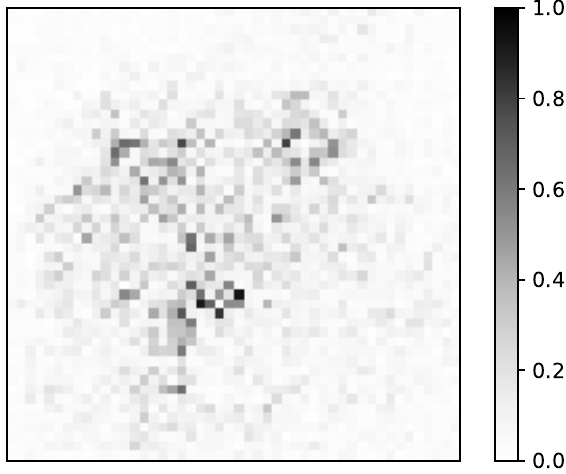}
    & 
    \includegraphics[width=1cm, height=1cm]{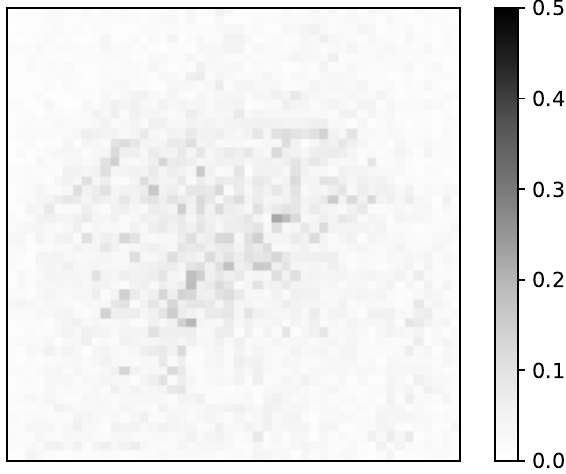}
    & 
    \includegraphics[width=1cm, height=1cm]{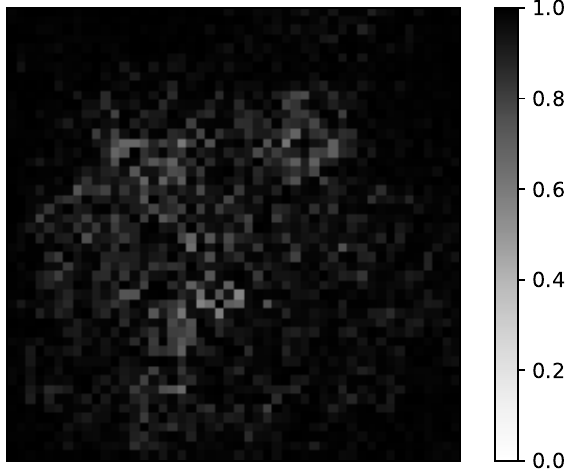}
    &
    \includegraphics[width=1cm, height=1cm]{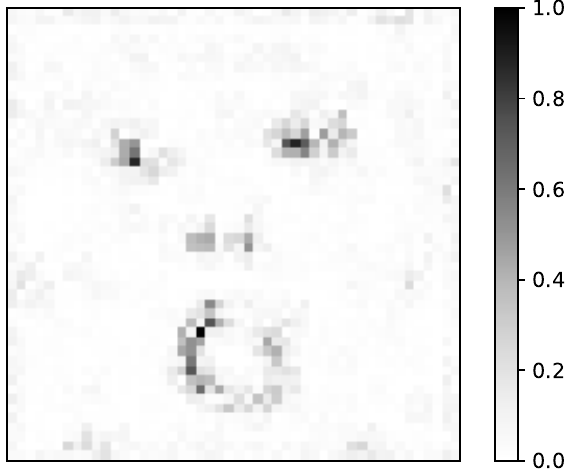}
    &  
    \includegraphics[width=1cm, height=1cm]{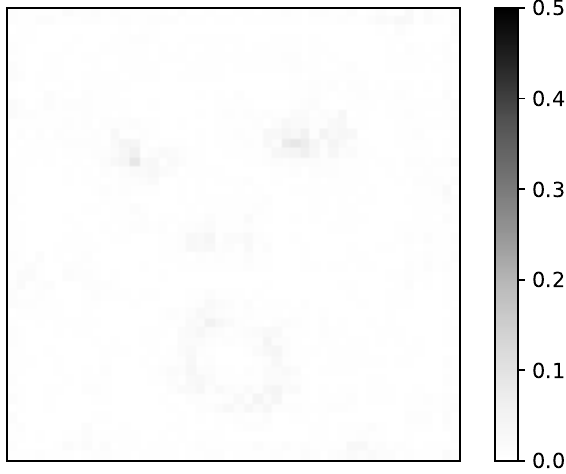}
    &
    \includegraphics[width=1cm, height=1cm]{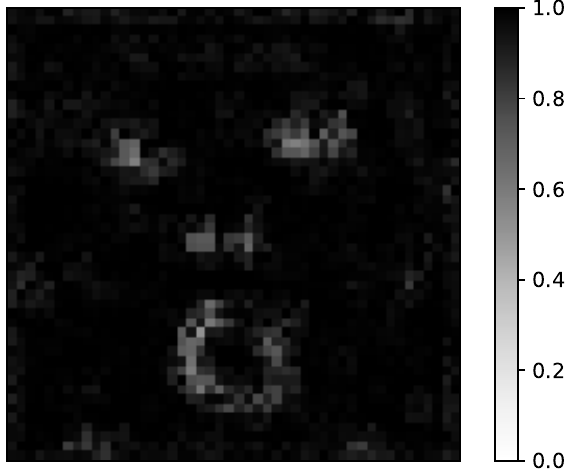}
    & 
    \includegraphics[width=1cm, height=1cm]{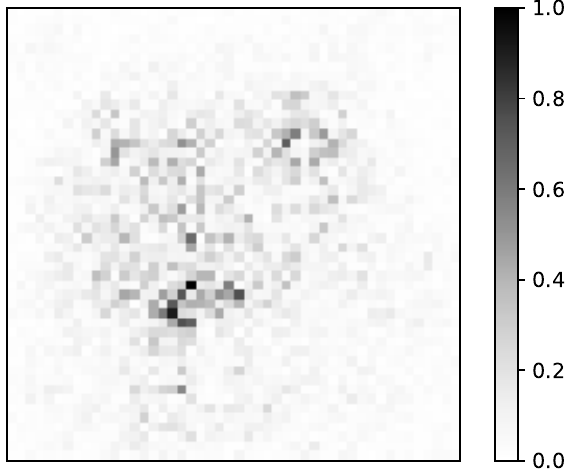}
    & 
    \includegraphics[width=1cm, height=1cm]{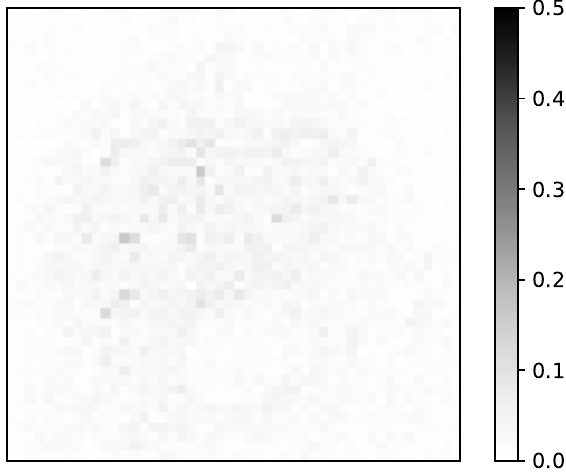}
    & 
    \includegraphics[width=1cm, height=1cm]{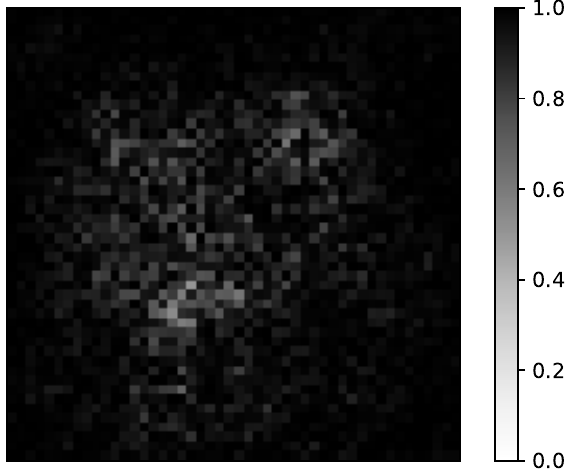}
    &
    \includegraphics[width=1cm, height=1cm]{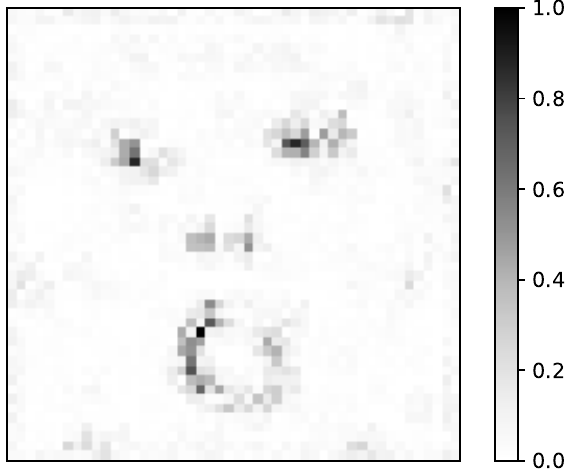}
    &  
    \includegraphics[width=1cm, height=1cm]{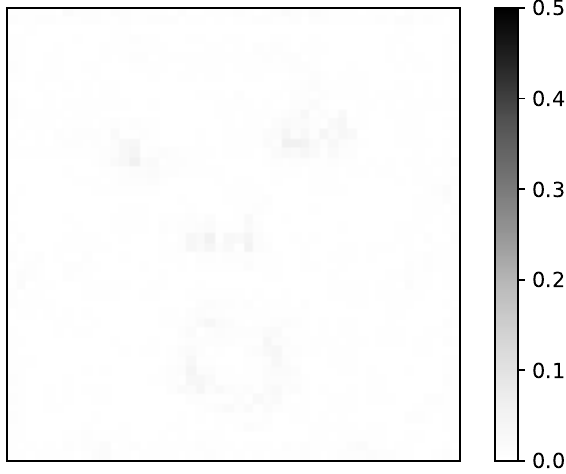}
    &
    \includegraphics[width=1cm, height=1cm]{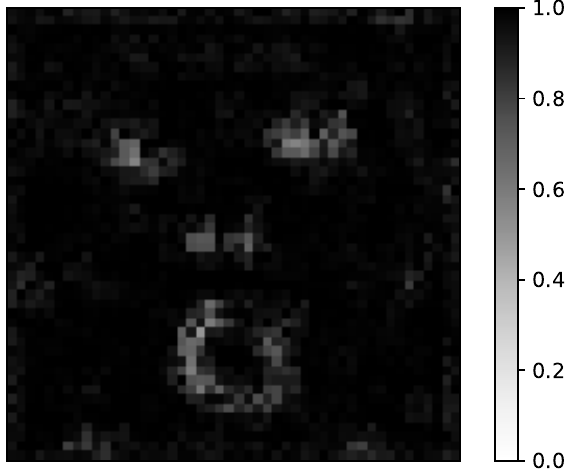}
    \\

    \rotatebox{90}{\centering \textbf{\fontsize{4}{4} \selectfont Flipout}}
    & 
    \includegraphics[width=1cm, height=1cm]{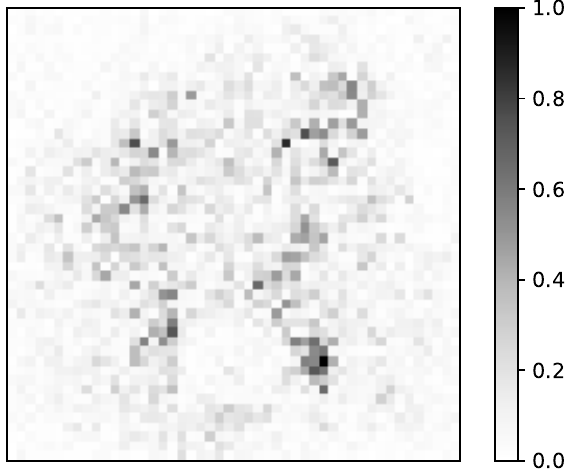}
    & 
    \includegraphics[width=1cm, height=1cm]{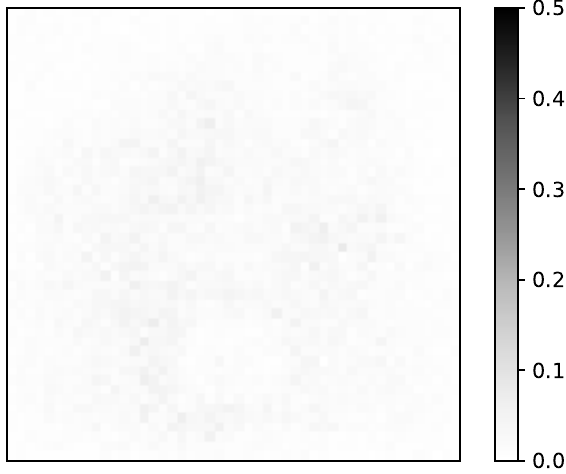}
    & 
    \includegraphics[width=1cm, height=1cm]{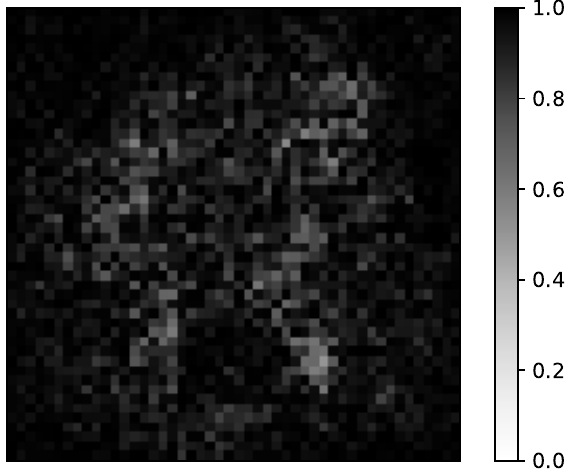}
    &
    \includegraphics[width=1cm, height=1cm]{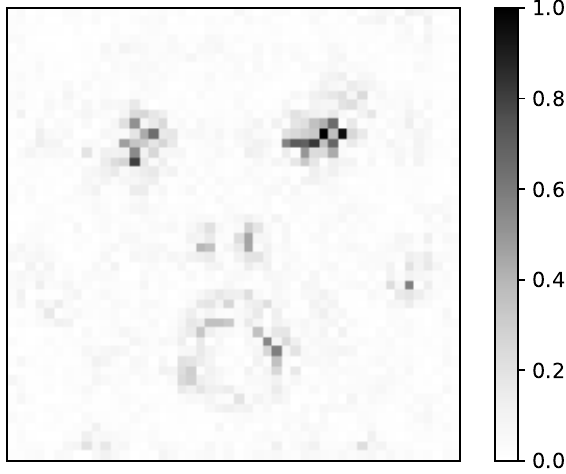}
    &  
    \includegraphics[width=1cm, height=1cm]{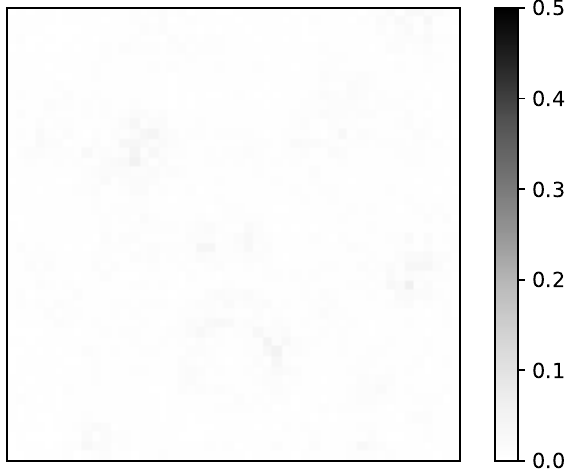}
    &
    \includegraphics[width=1cm, height=1cm]{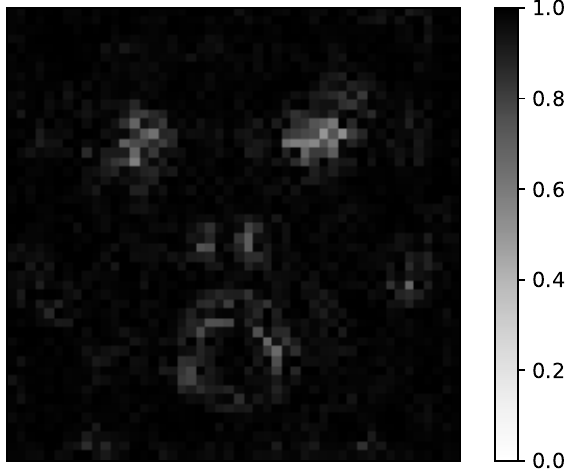}
    & 
    \includegraphics[width=1cm, height=1cm]{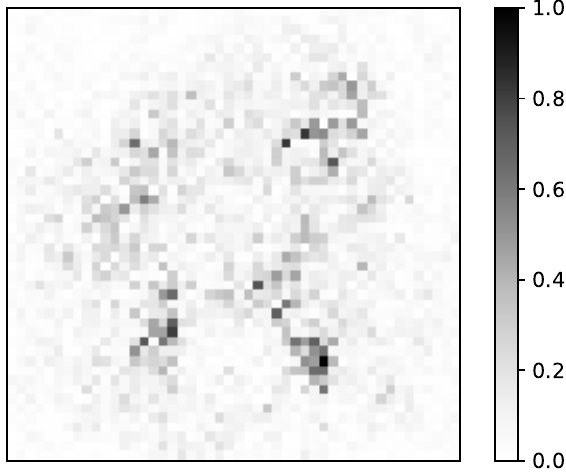}
    & 
    \includegraphics[width=1cm, height=1cm]{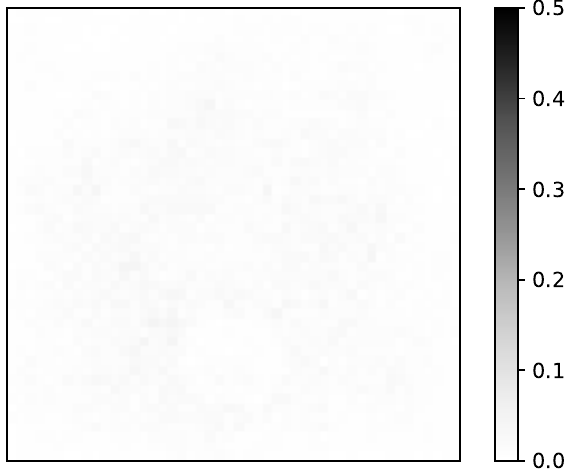}
    & 
    \includegraphics[width=1cm, height=1cm]{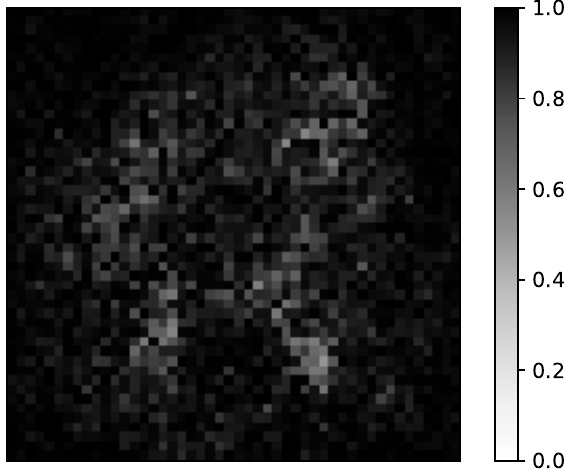}
    &
    \includegraphics[width=1cm, height=1cm]{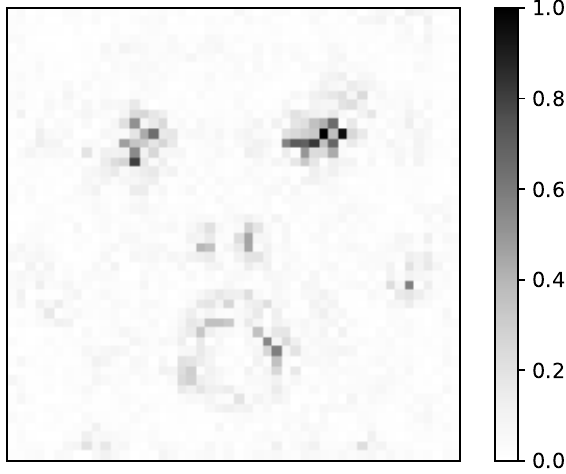}
    &  
    \includegraphics[width=1cm, height=1cm]{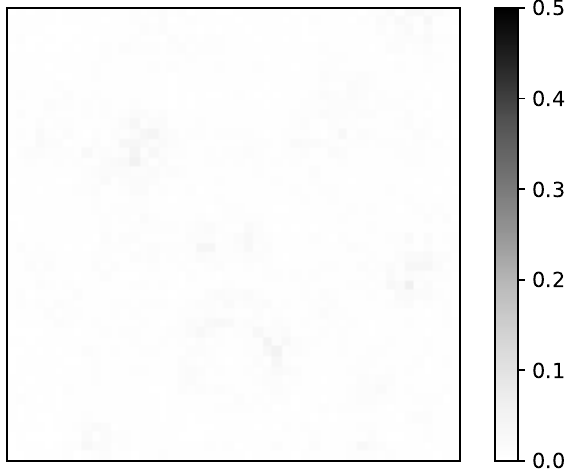}
    &
    \includegraphics[width=1cm, height=1cm]{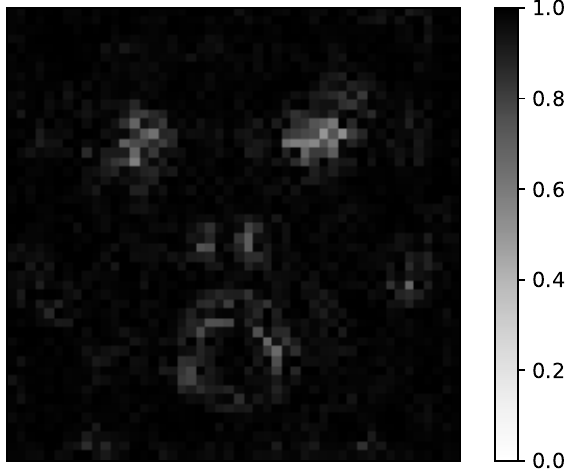}
    \\    
    \bottomrule

    \end{tabular}
    \caption{Visualization of the mean \textbf{($E_\mu$)}, the standard deviation \textbf{($E_\sigma$)}, and the coefficient of variation \textbf{($CV$)} heatmaps generated using the combinations of uncertainty estimation methods (Deep Ensemble, MC-Dropout, MC-DropConnect and Flipout) and explanation methods (Guided Backpropagation and Integrated Gradients). The uncertainty estimation methods are organized in individual rows and the explanation methods are aligned along the columns with the columns 1-3 representing the $\mu$, $\sigma$, and the $CV$ heatmaps for IG and the columns 4-6 representing the same for GBP.}
    \label{fer_variant_experiment_example_1}
\end{figure*}

The presentation of the results is similar to that in Section \ref{explanation_uncertainty_for_image_classification_result} for the same input image. The results are shown in the last four rows of Figure \ref{fer_variant_experiment_example_1}. 

\par
We expect the output of both the variants of explanation computation to differ \textit{only} when the network prediction \textit{does not} match the ground truth for the samples obtained from forward passes/ensemble models.
\par 
Similar to the observations with original gradient computation, we find that in Figure \ref{fer_variant_experiment_example_1} the explanations generated by GBP have higher certainty in comparison to the explanations generated by IG. Additionally, we notice the modifications applied to the gradient computation step do not drastically alter the generated explanation. A slight variation in the activation of different pixel regions can be observed in the case of IG, but it is not as prominent in GBP. 
\par 
A possible reason for no drastic changes in the explanation could be that the averaging effect of the multitude of explanation samples negates the impact of the modified gradient computation. It might be possible that the majority of the output prediction sample matches the ground truth thereby outweighing the effect of modified gradient computation associated with those incorrect output predictions. In such a case, where the model prediction increasingly matches the ground truth, the results of both the original gradient computation and the variation in gradient computation converge.

\subsection{Regression Results}
\begin{figure}[t]
    \centering
    \begin{subfigure}{0.01\textwidth}
        \rotatebox[origin=c]{90}{\fontsize{4}{4} \selectfont \textbf{GBP}}
        \vspace*{1cm}
    \end{subfigure}
    \begin{subfigure}{0.1\textwidth}
        \centering
        \textbf{\fontsize{4}{4} \selectfont Ensemble}\\
        \includegraphics[width=\linewidth]{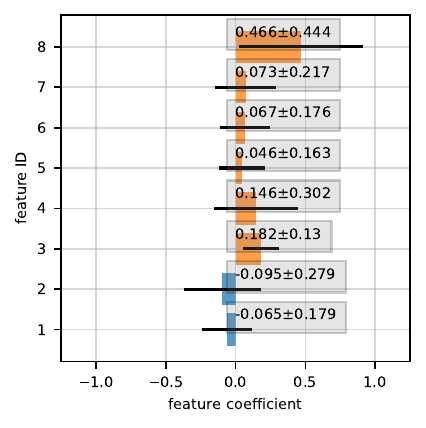}
    \end{subfigure}
    \hfill
    \begin{subfigure}{0.1\textwidth}
        \centering
        \textbf{\fontsize{4}{4} \selectfont MC-Dropout}\\
        \includegraphics[width=\linewidth]{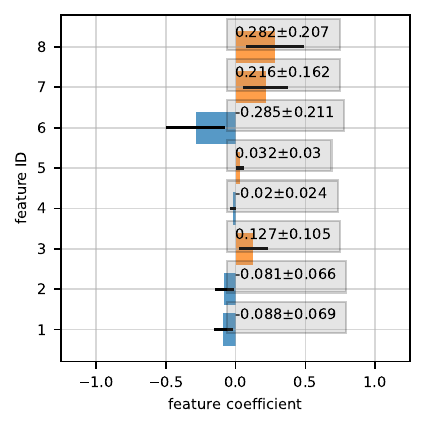}
    \end{subfigure}
    \hfill
    \begin{subfigure}{0.1\textwidth}
        \centering
        \textbf{\fontsize{4}{4} \selectfont MC-DropConnect}\\
        \includegraphics[width=\linewidth]{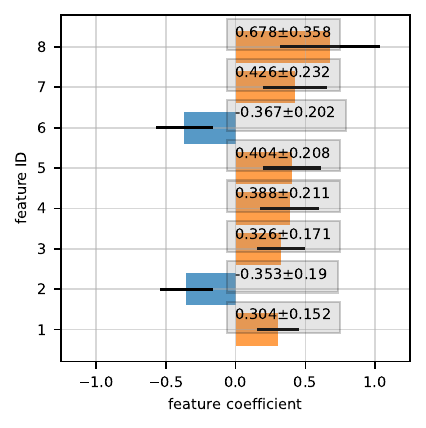}
    \end{subfigure}
    \hfill
    \begin{subfigure}{0.1\textwidth}
        \centering
        \textbf{\fontsize{4}{4} \selectfont Flipout}\\
        \includegraphics[width=\linewidth]{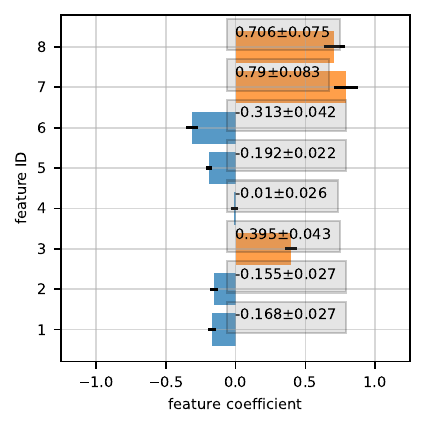}
    \end{subfigure}

    \centering
    \begin{subfigure}{0.01\textwidth}
        \rotatebox[origin=c]{90}{\fontsize{4}{4} \selectfont\textbf{LIME}}
        \vspace*{1cm}
    \end{subfigure}
    \begin{subfigure}{0.1\textwidth}
        \includegraphics[width=\linewidth]{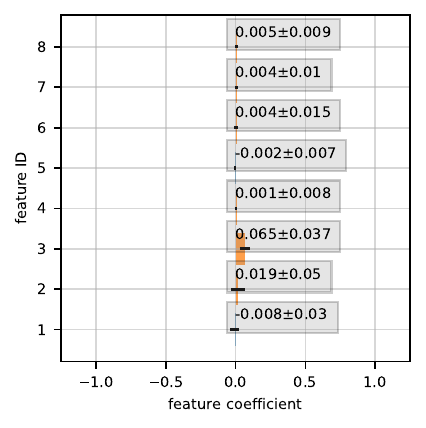}
    \end{subfigure}
    \hfill
    \begin{subfigure}{0.1\textwidth}
        \includegraphics[width=\linewidth]{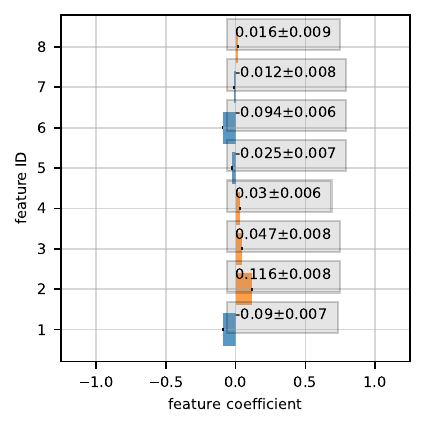}
    \end{subfigure}
    \hfill
    \begin{subfigure}{0.1\textwidth}
        \includegraphics[width=\linewidth]{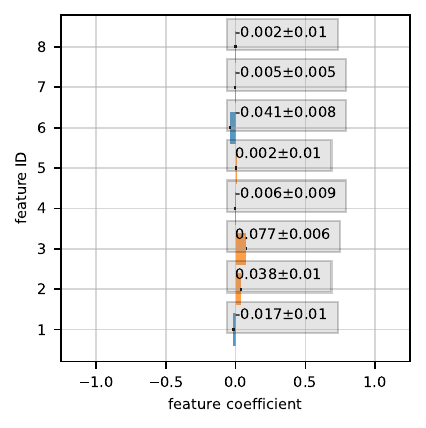}
    \end{subfigure}
    \hfill
    \begin{subfigure}{0.1\textwidth}
        \includegraphics[width=\linewidth]{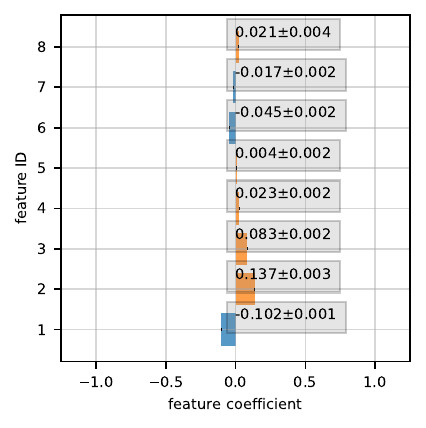}
    \end{subfigure}

    \centering
    \begin{subfigure}{0.01\textwidth}
        \rotatebox[origin=c]{90}{\fontsize{4}{4} \selectfont \textbf{GBP}}
        \vspace*{1cm}
    \end{subfigure}
    \begin{subfigure}{0.1\textwidth}
        \includegraphics[width=\linewidth]{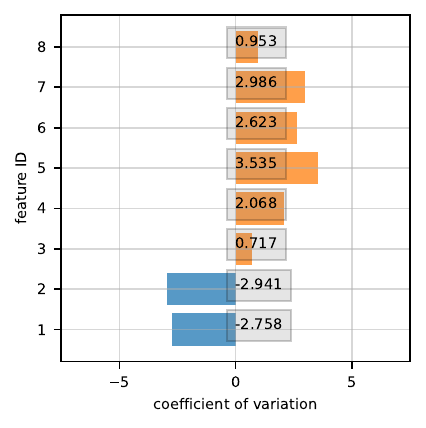}
    \end{subfigure}
    \hfill
    \begin{subfigure}{0.1\textwidth}
        \includegraphics[width=\linewidth]{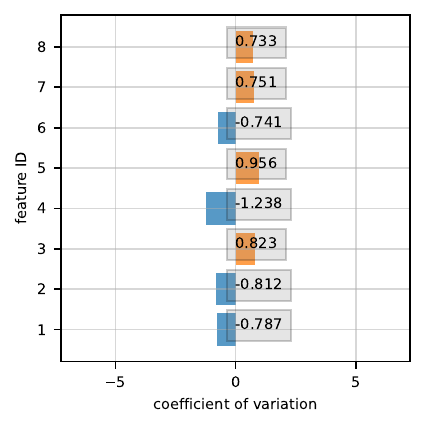}
    \end{subfigure}
    \hfill
    \begin{subfigure}{0.1\textwidth}
        \includegraphics[width=\linewidth]{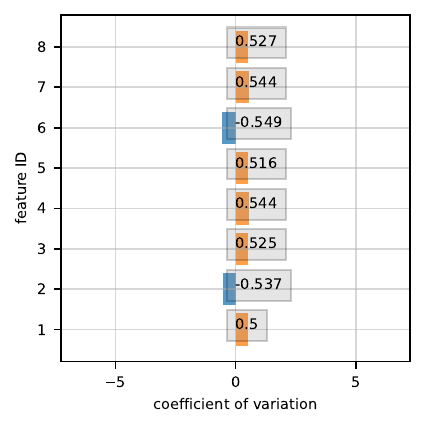}
    \end{subfigure}
    \hfill
    \begin{subfigure}{0.1\textwidth}
        \includegraphics[width=\linewidth]{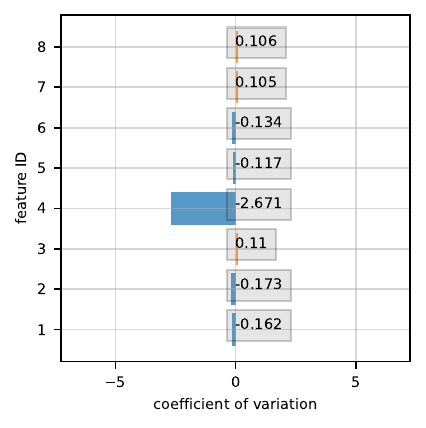}
    \end{subfigure}

    \centering
    \begin{subfigure}{0.01\textwidth}
        \rotatebox[origin=c]{90}{\fontsize{4}{4} \selectfont \textbf{LIME}}
        \vspace*{1cm}
    \end{subfigure}
    \begin{subfigure}{0.1\textwidth}
        \includegraphics[width=\linewidth]{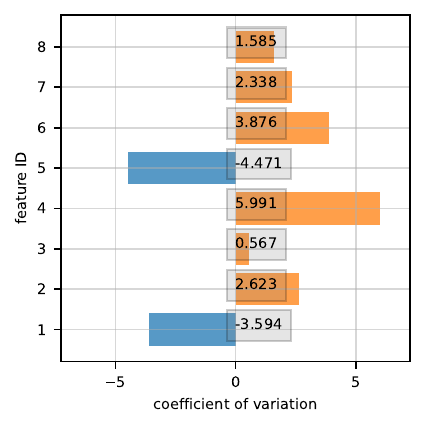}
    \end{subfigure}
    \hfill
    \begin{subfigure}{0.1\textwidth}
        \includegraphics[width=\linewidth]{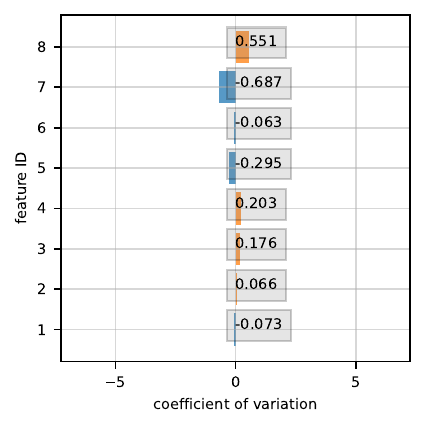}
    \end{subfigure}
    \hfill
    \begin{subfigure}{0.1\textwidth}
        \includegraphics[width=\linewidth]{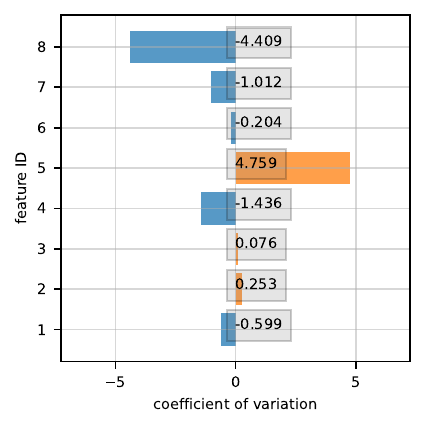}
    \end{subfigure}
    \hfill
    \begin{subfigure}{0.1\textwidth}
        \includegraphics[width=\linewidth]{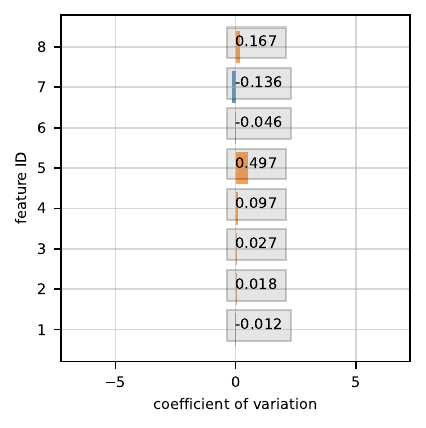}
    \end{subfigure}

    \begin{subfigure}{0.25\textwidth}
        \centering
        \includegraphics[width=\textwidth]{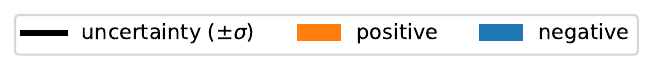}
    \end{subfigure}
    \caption{The explanations for the regression task with the mean and the standard deviation explanation (rows 1-2) and the coefficient of variation explanations (rows 3-4). The rows  1 and 3 represent the outputs of GBP and the rows 2 and 4 represent LIME. From left to right, the columns represent as follows: (i) Deep Ensemble (ii) MC-Dropout (iii) MC-DropConnect (iv) Flipout. The input features mapping in the plots is as follows: [1: longitude, 2: latitude, 3: housing\ median\ age, 4: total\ rooms, 5: total\ bedrooms, 6: population, 7: households, 8: median\ income].}
    \label{experiment_4_result_ms_cv_1}
\end{figure}

As the modality of input data changes for the tabular regression task, we change the depiction of the explanation as well. We visualize feature importance as bar charts for this task. The length of the bars shows the mean explanation and the error bars depict the standard deviation explanation. Figure \ref{experiment_4_result_ms_cv_1} depicts the output obtained on a data point taken from the California Housing Dataset.

\par 
It can be observed that the mean explanations generated by LIME generally have a lower activation as compared to GBP. This indicates less agreement amongst the constituent explanations and in turn higher uncertainty in the explanation generated by LIME. 

Furthermore, we find that for LIME, the value of the coefficient of variation is slightly higher for several features, especially in the case of Deep Ensemble and MC-DropConnect. This implies that these explanations are highly uncertain and should be used with caution. This can be attributed to disagreement among the constituent explanations for that particular feature. The same is observed in the case of GBP for a couple of features, (to a smaller degree as compared to LIME), especially for the case of Deep Ensembles.

\section{Conclusions and Future Work}
\label{sec:conclusions_and_future_work}

In this section, we discuss the conclusions and answer the questions formulated in the previous section of this paper. 
\par 
RQ1: We have demonstrated that additional combinations of uncertainty estimation and explanation methods can be implemented efficiently to generate explanation uncertainty by using our proposed approach. RQ2: We identified that the mean, the standard deviation, and the coefficient of variation can be used for concise analysis and representation of the resultant explanation distribution. 
The features having a high mean, low standard deviation, and subsequently a low coefficient of variation are generally more likely to have influenced the model output.
RQ3: To analyze the quality of explanation distributions, we proposed and computed a modified version of the pixel insertion/deletion metric. We find that a higher value for pixel insertion and a lower value for pixel deletion are the characteristics of a good explanation.
RQ4: From the analysis of the explanation representations, we deduce that any combination of uncertainty estimation with GBP (as an explanation method) can produce useful and highly confident explanation distributions. This can be supplemented by the observation that GBP tends to generate less noisy explanation heatmaps with higher mean and lower standard deviation for relevant features.

We expect that explanation uncertainty becomes a standard feature of saliency explanations as used by researchers and practitioners, same as it should be for uncertainty estimation methods, so a machine learning model can produce a prediction, with an uncertainty estimate, and an explanation, also with its uncertainty estimate, for a complete set of information required for trustworthiness.

\par
We hope that our proposed pipeline can be utilized to verify new developments in uncertainty estimation and explanation approaches. This work can be extended by incorporating a human-based evaluation of the explanations generated by algorithms. Having such feedback from users could provide valuable insights for improvements in the explanation methods.

\section*{Broader Impact Statement}

Both uncertainty estimation and saliency explanations have broad impact on society, as machine learning models are increasingly being used in real-world settings, where humans could be harmed, and uncertainty estimation produces better calibrated output uncertainties to detect when a model makes incorrect or ambiguous predictions. Similarly for saliency explanations, when explanation uncertainty is produced by using an uncertainty estimation method, this would allow human users to obtain additional information in order to decide if the explanation should be trusted or not.

In both cases, output uncertainty and saliency explanations can be misleading, and always additional validation is necessary when used by human users. This paper did not make a user study to evaluate the quality of explanations and uncertainty from a human perspective, we leave this for future work.

\clearpage
{
    \small
    \bibliographystyle{ieeenat_fullname}
    \bibliography{main}
}

\clearpage
\setcounter{page}{1}
\appendix 

\section{Additional Model Training Details}
\label{model_training}

The specifications for the model training have been provided in Tables \ref{cifar_minivgg} to Table \ref{hparam}.

    \begin{table}[h]
    \centering
    \begin{tabular} 
    {p{0.10\textwidth} 
    p{0.10\textwidth}
    p{0.10\textwidth}
    p{0.10\textwidth}
    }
    
    \toprule
    \textbf{Hyperparams.} & 
    \textbf{CIFAR-10} &
    \textbf{FER+} &
    \textbf{CHD}
    \\
    \toprule
    optimizer & adam & adam & RMSProp\\ 
    loss & cross-entropy & cross-entropy & MSE \\
    metric & accuracy & accuracy & MAE\\
    learning rate & 0.001 & 0.001 & 0.001\\ 
    train samples & 50000 & 28559 & 12750\\
    validation samples & 9500 & 3579 & 4250\\ 
    test samples & 500 & 3573 & 4250\\ 
    train batch size & 2048 & 2048 & 4096\\
    validation batch size & 2048 & 2048 & 4096\\ 
     
    \bottomrule
    \end{tabular}
    \caption{Hyperparameter used to train the miniVGG16 on CIFAR-10, FER+ and California Housing Dataset (CHD).  MSE: Mean Squared Error, MAE: Mean Absolute Error.}
    \label{cifar_minivgg}
    \end{table}

    \begin{table}[h]
    \centering
    \begin{tabular}
    {p{0.15\textwidth} 
    p{0.10\textwidth}
    p{0.10\textwidth}}
    
    \toprule
    {\centering \textbf{Augmentation}} & 
    {\centering \textbf{CIFAR-10}} &
    {\centering \textbf{FER+}} \\
    \toprule
    width shift range  & 0.1 & 0.1\\ 
    height shift range & 0.1 & 0.1\\ 
    shear range & 0.1 & -\\
    zoom range  &  0.1 & 0.1\\
    horizontal flip  & True & True\\ 
    rescale & 1/255 & 1/255 \\ 
    rotation range & - & 30 \\
     
    \bottomrule
    \end{tabular} 
    \caption{Data augmentation applied to CIFAR-10 and FER+.}
    \label{data_augmentation_cifar_minivgg}
    \end{table}

    \begin{table}[h]
    \centering
    \begin{tabular}{p{0.150\textwidth} 
    p{0.175\textwidth}}
    \toprule
    \textbf{Hyperparameter} & \textbf{Search Space}  \\ 
    \toprule
    \# of units in a layer & \{1, 2, 4, 6, 8\}  \\ 
    \# of layers & \{1, 2, 3, 4, 5\}\\
    \# of epochs & \{10, 50, 100, 150\}  \\
    optimizer & \{\lq{SGD}\rq, \lq{RMSProp}\rq,
       \lq Adam\rq\} \\
    \bottomrule
    \end{tabular}
    \caption{Search space used for the hyperparameter tuning conducted for the MLP.}
    \label{hparam}
    \end{table}

    \begin{figure}[t]
    \centering
    \includegraphics[width=0.475\textwidth]{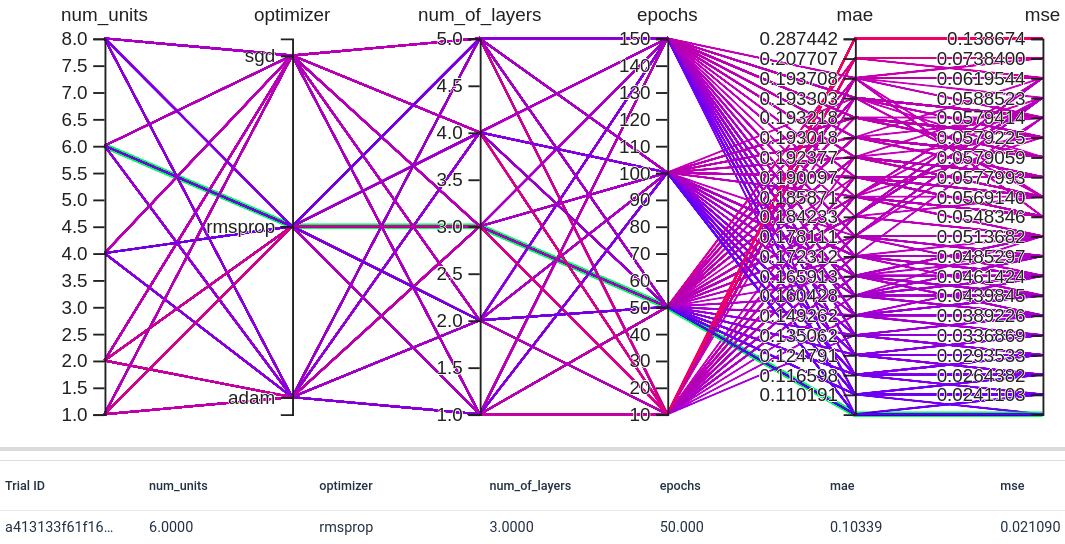}
    \caption{Parallel plot depicting the results of hyperparameter search conducted for creating an MLP. The optimum values of the hyperparameters are provided in the table below the plot.}
    \label{chd_hyperparameter_optimization_mlp_parallel}
    \end{figure}

    \begin{figure}[t]
    \centering
    \includegraphics[width=0.475\textwidth]{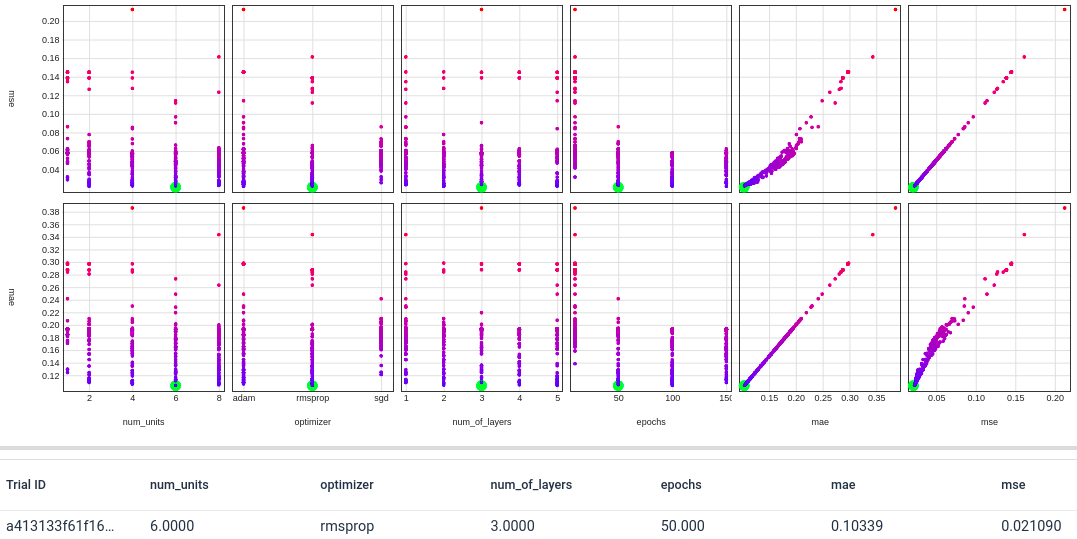}
    \caption{Scatter plot depicting the results of hyperparameter search conducted for creating an MLP. The optimum values of the hyperparameters are provided in the table below the plot.}
    \label{chd_hyperparameter_optimization_mlp_scatter}
    \end{figure}

\section{Additional Insertion/Deletion Curves}

\begin{figure*}
    \centering
    \begin{tabular}{p{0.09\textwidth}p{0.22\textwidth}p{0.20\textwidth}p{0.20\textwidth}p{0.20\textwidth}}
         &  \footnotesize \textbf{Ensemble}  & \footnotesize \textbf{MC-Dropout} & \footnotesize \textbf{MC-DropConnect} & \footnotesize \textbf{Flipout} \\
    \end{tabular}
    
    \begin{subfigure}{0.01\textwidth}
        \rotatebox[origin=c]{90}{\textbf{\fontsize{5}{4} \selectfont $\mu$-GBP}}
        \vspace*{0.5cm}
    \end{subfigure}
    \begin{subfigure}{0.11\textwidth}
        \centering
        \textbf{\fontsize{5}{4} \selectfont Deletion}\\
        \includegraphics[width=\linewidth]%
        {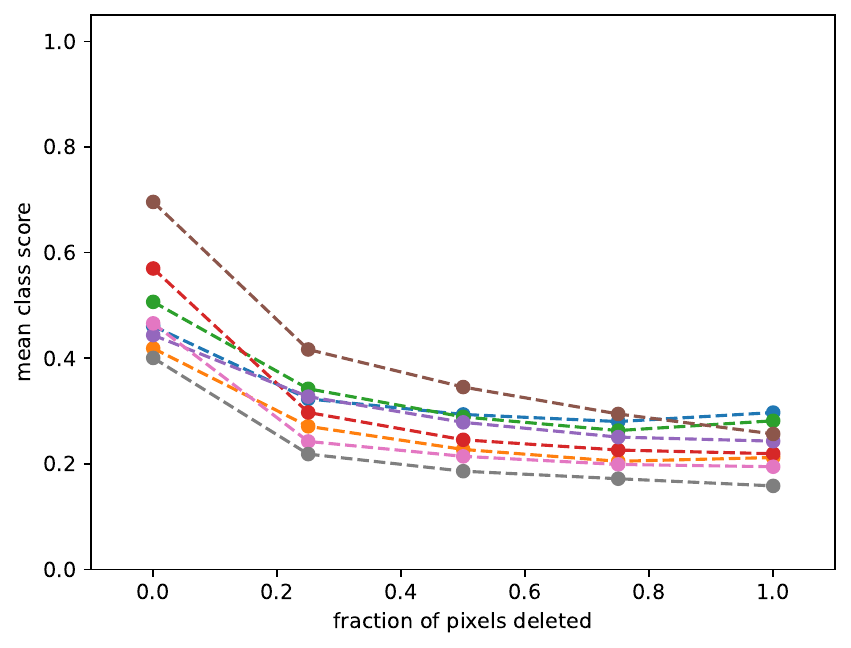}
    \end{subfigure}
    \begin{subfigure}{0.11\textwidth}
        \centering
        \textbf{\fontsize{5}{4} \selectfont Insertion}
        \includegraphics[width=\linewidth]%
        {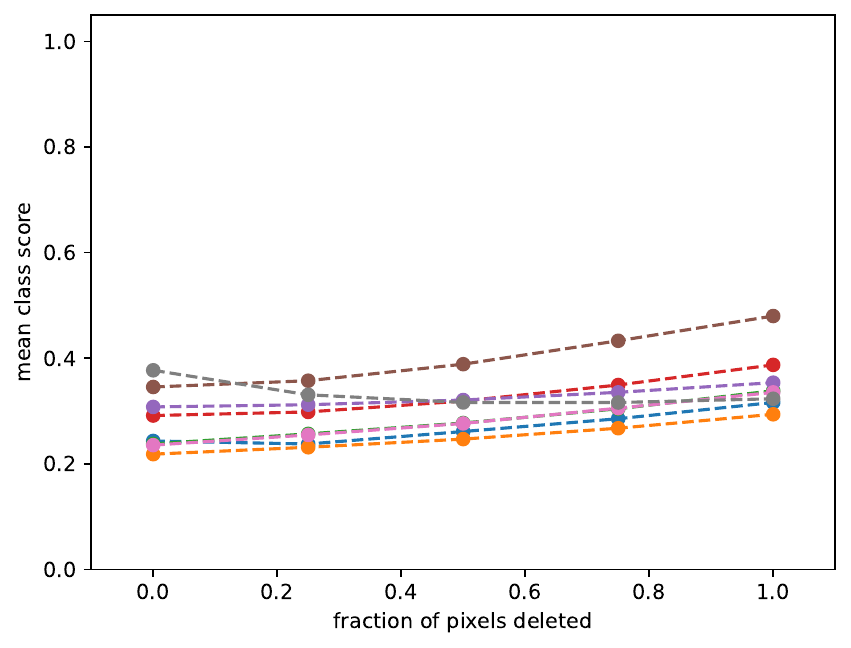}

    \end{subfigure}
    \begin{subfigure}{0.11\textwidth}
        \centering
        \textbf{\fontsize{5}{4} \selectfont Deletion}
        \includegraphics[width=\linewidth]%
        {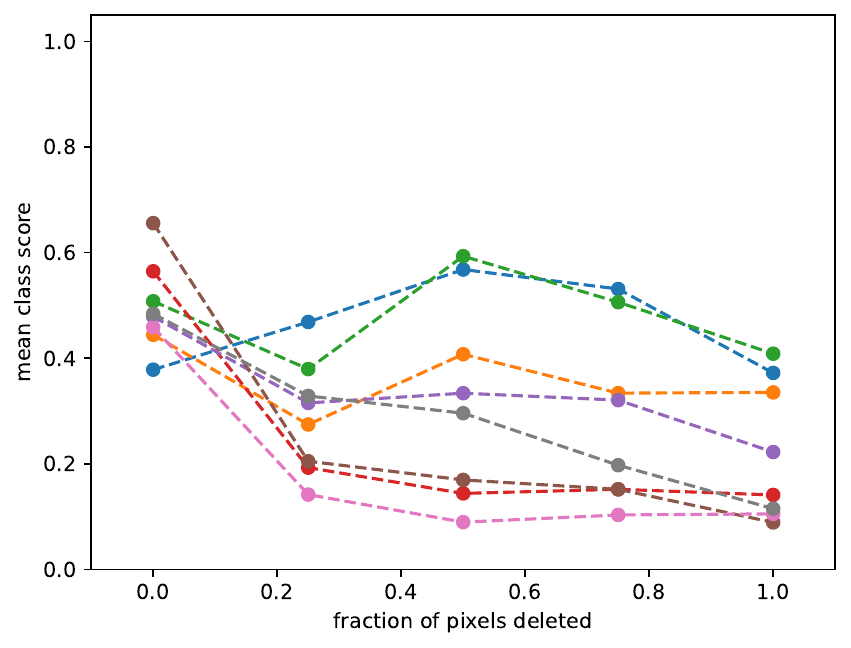}
    \end{subfigure}
    \begin{subfigure}{0.11\textwidth}
        \centering
        \textbf{\fontsize{5}{4} \selectfont Insertion}
        \includegraphics[width=\linewidth]%
        {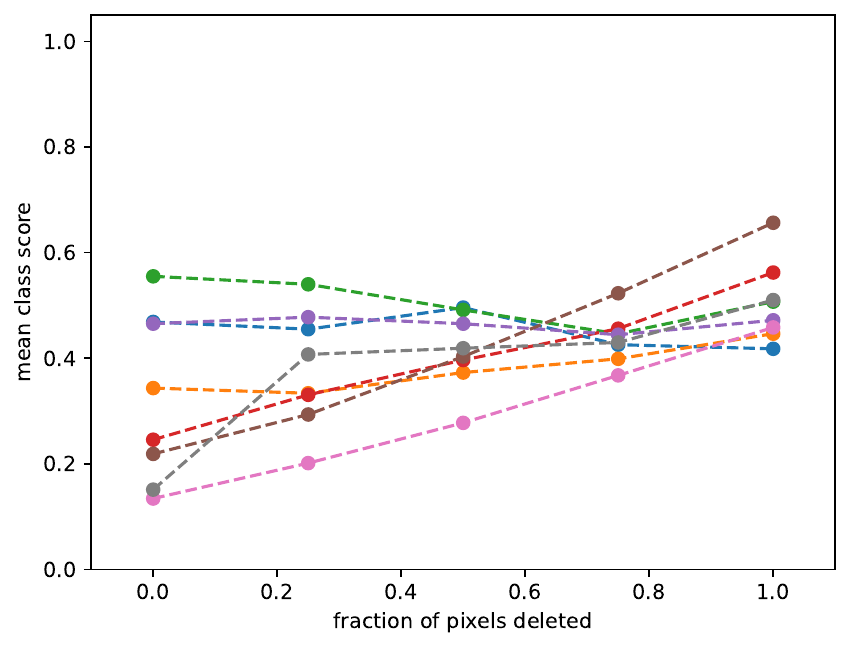}
    \end{subfigure}
    \begin{subfigure}{0.11\textwidth}
        \centering
        \textbf{\fontsize{5}{4} \selectfont Deletion}
        \includegraphics[width=\linewidth]%
        {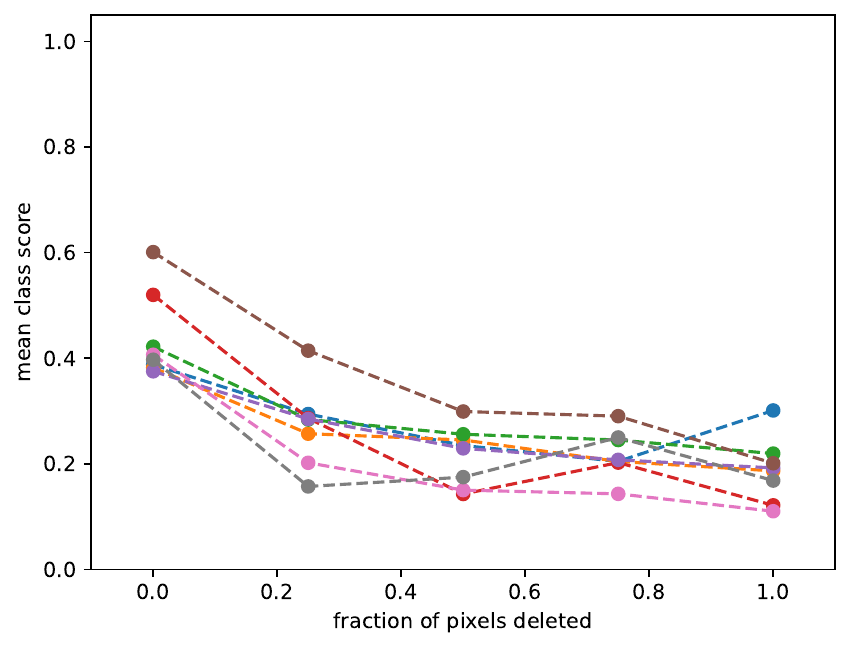}
    \end{subfigure}
    \begin{subfigure}{0.11\textwidth}
        \centering
        \textbf{\fontsize{5}{4} \selectfont Insertion}
        \includegraphics[width=\linewidth]%
        {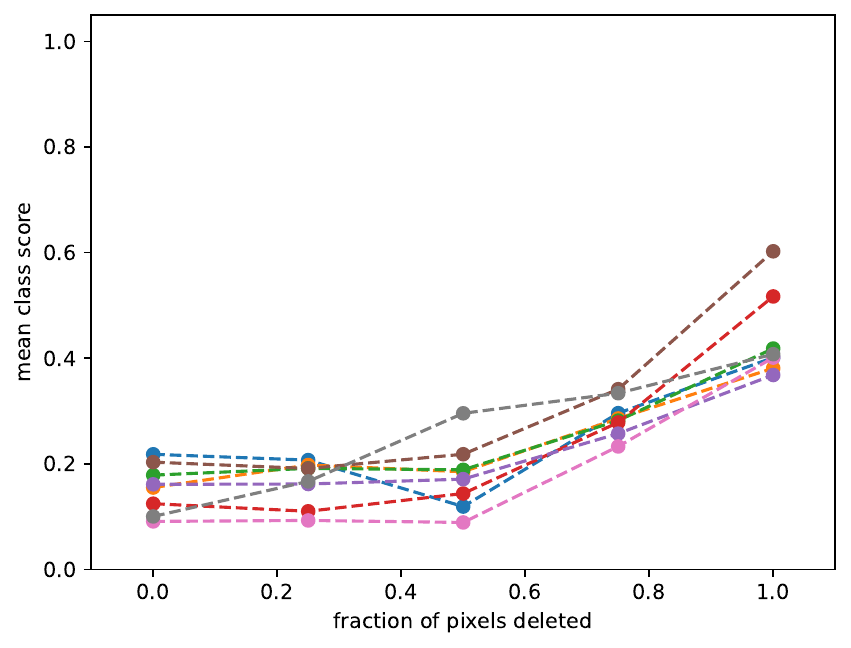}
    \end{subfigure}
    \begin{subfigure}{0.11\textwidth}
        \centering
        \textbf{\fontsize{5}{4} \selectfont Deletion}
        \includegraphics[width=\linewidth]%
        {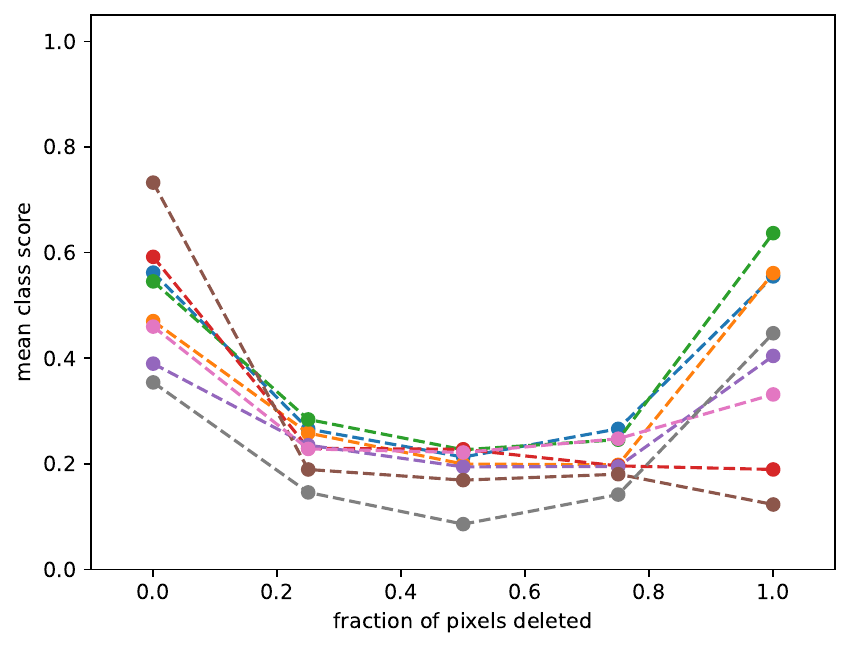}
    \end{subfigure}
    \begin{subfigure}{0.11\textwidth}
        \centering
        \textbf{\fontsize{5}{4} \selectfont Insertion}
        \includegraphics[width=\linewidth]%
        {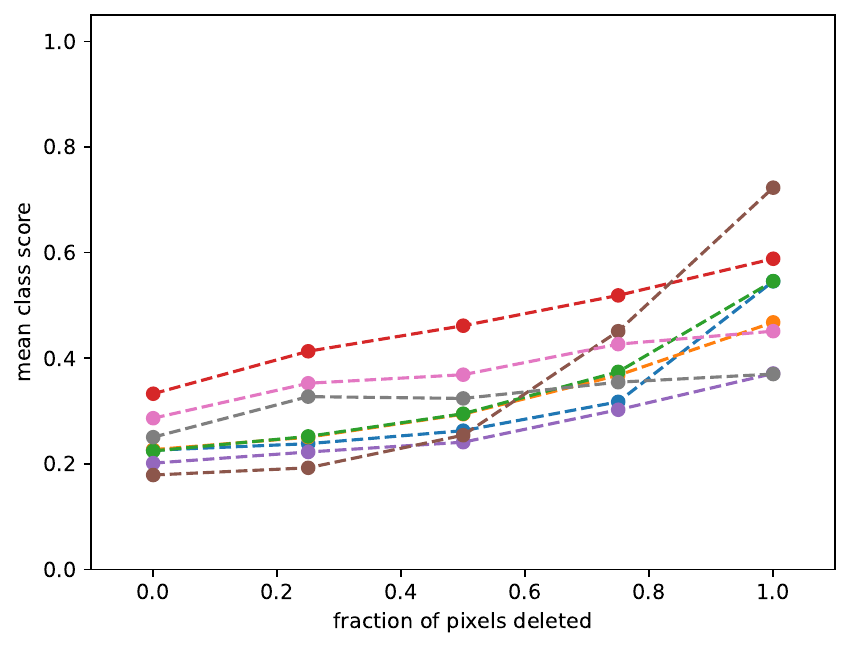}
    \end{subfigure} \\
    
    \begin{subfigure}{0.01\textwidth}
        \rotatebox[origin=c]{90}{\textbf{\fontsize{5}{4} \selectfont $\sigma$-GBP}}
        \vspace*{0.5cm}
    \end{subfigure}
    \begin{subfigure}{0.11\textwidth}
        \includegraphics[width=\linewidth]%
        {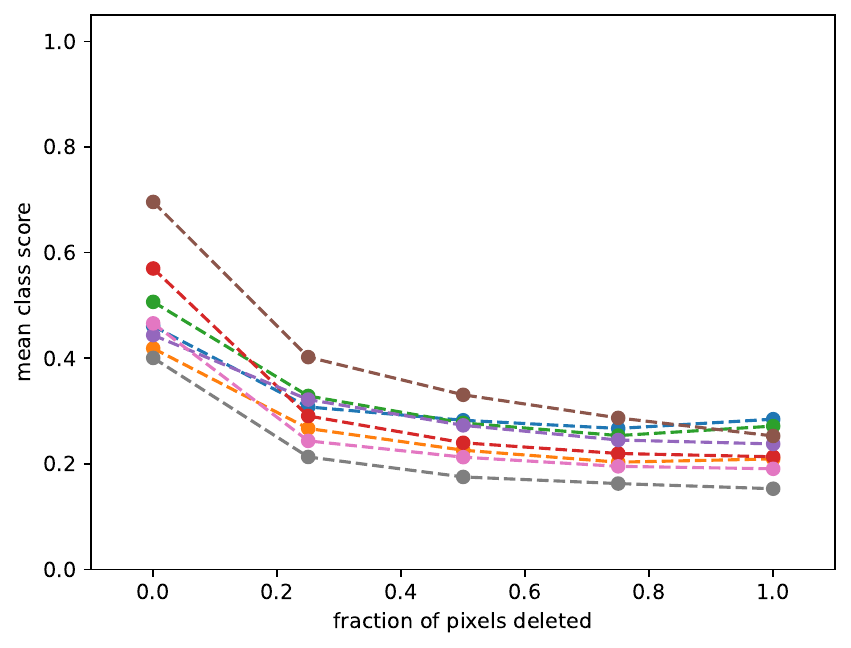}
    \end{subfigure}
    \begin{subfigure}{0.11\textwidth}
        \includegraphics[width=\linewidth]%
        {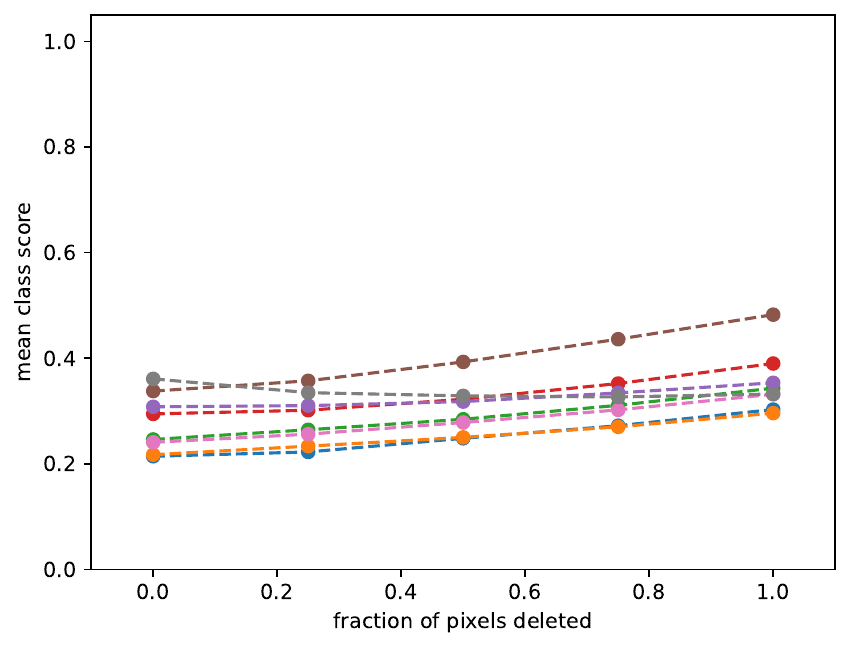}
    \end{subfigure}
    \begin{subfigure}{0.11\textwidth}
        \includegraphics[width=\linewidth]%
        {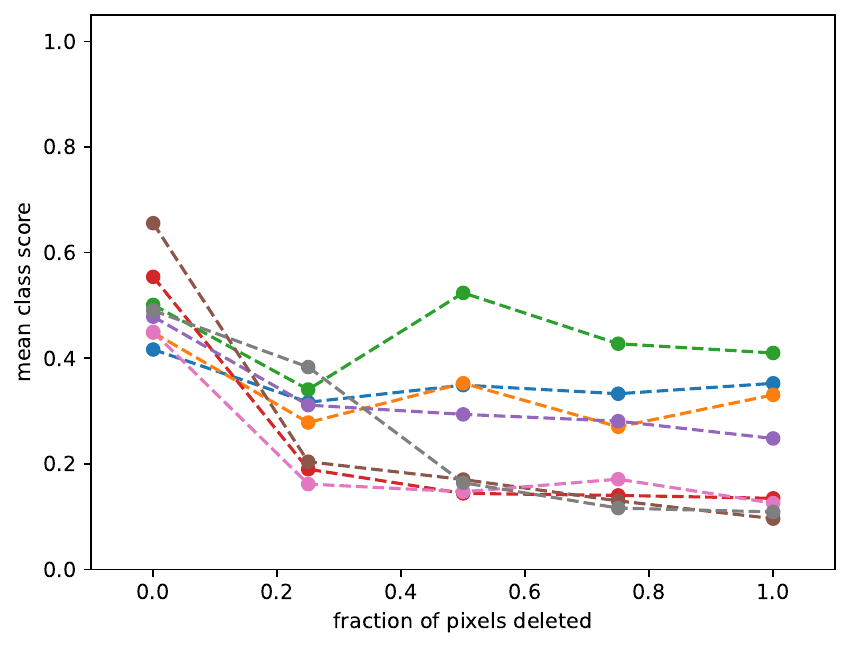}
    \end{subfigure}
    \begin{subfigure}{0.11\textwidth}
        \includegraphics[width=\linewidth]%
        {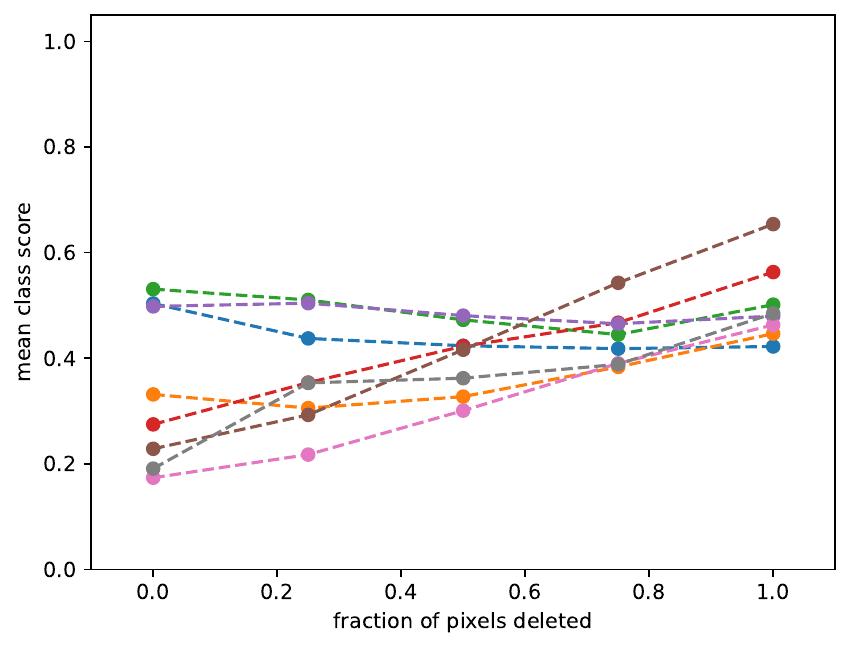}
    \end{subfigure}
    \begin{subfigure}{0.11\textwidth}
        \includegraphics[width=\linewidth]%
        {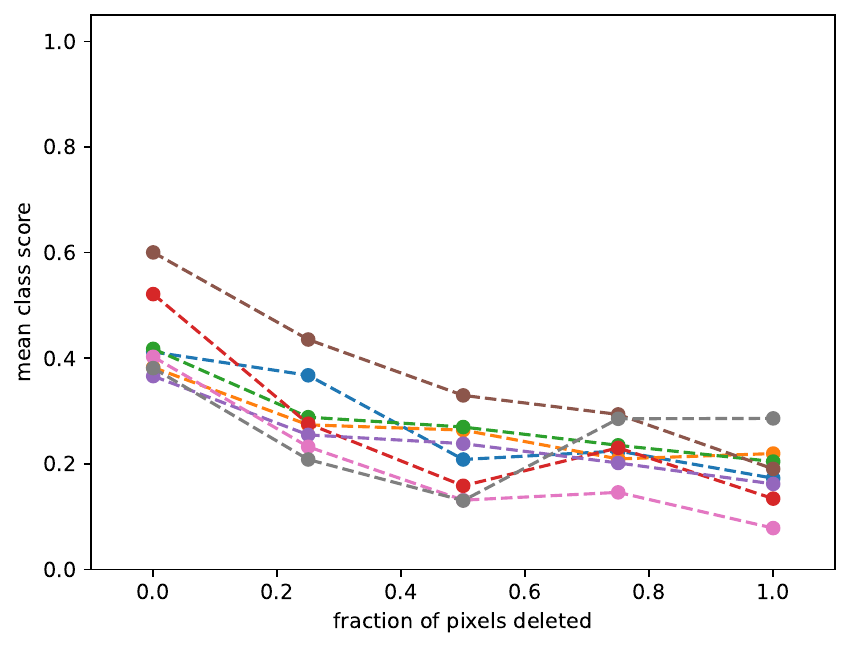}
    \end{subfigure}
    \begin{subfigure}{0.11\textwidth}
        \includegraphics[width=\linewidth]%
        {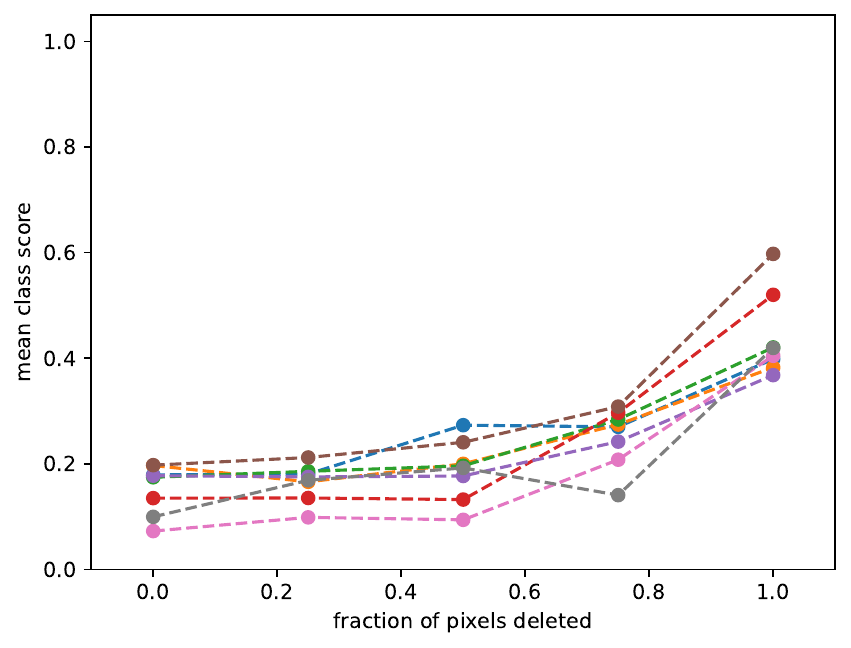}
    \end{subfigure}
    \begin{subfigure}{0.11\textwidth}
        \includegraphics[width=\linewidth]%
        {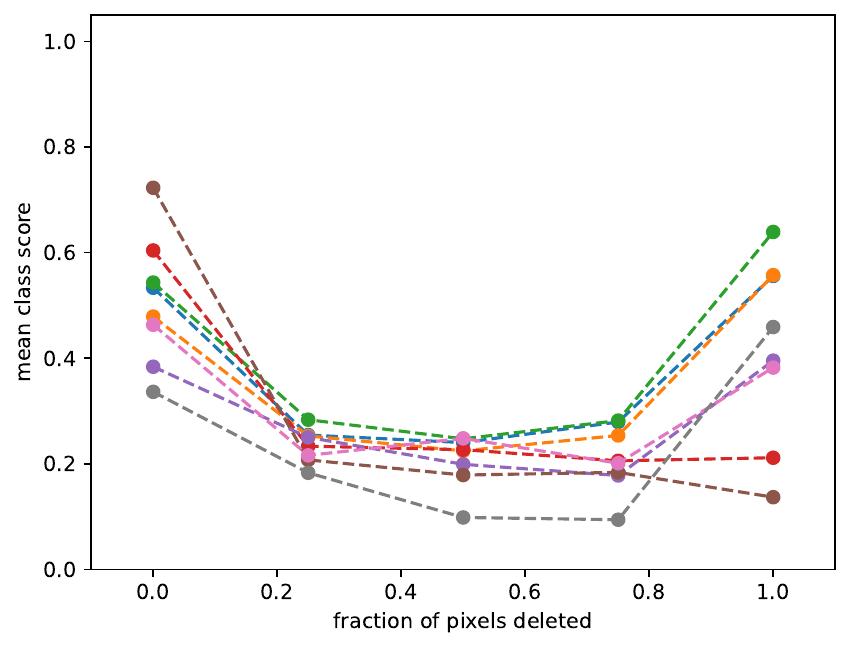}
    \end{subfigure}
    \begin{subfigure}{0.11\textwidth}
        \includegraphics[width=\linewidth]%
        {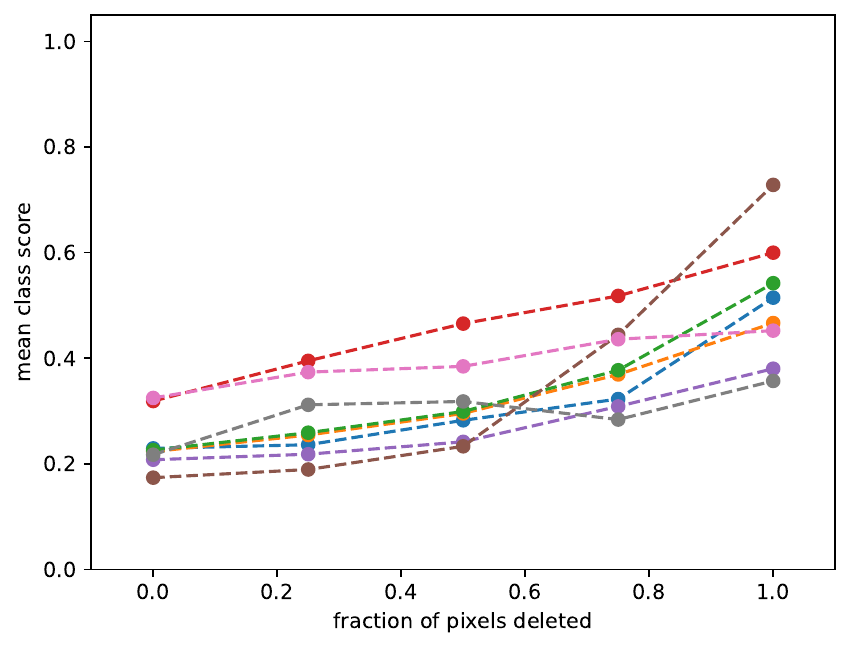}
    \end{subfigure}
    \\

    \begin{subfigure}{0.01\textwidth}
        \rotatebox[origin=c]{90}{\textbf{\fontsize{5}{4} \selectfont $\mu$-IG}}
        \vspace*{0.5cm}
    \end{subfigure}
    \begin{subfigure}{0.11\textwidth}
        \includegraphics[width=\linewidth]%
        {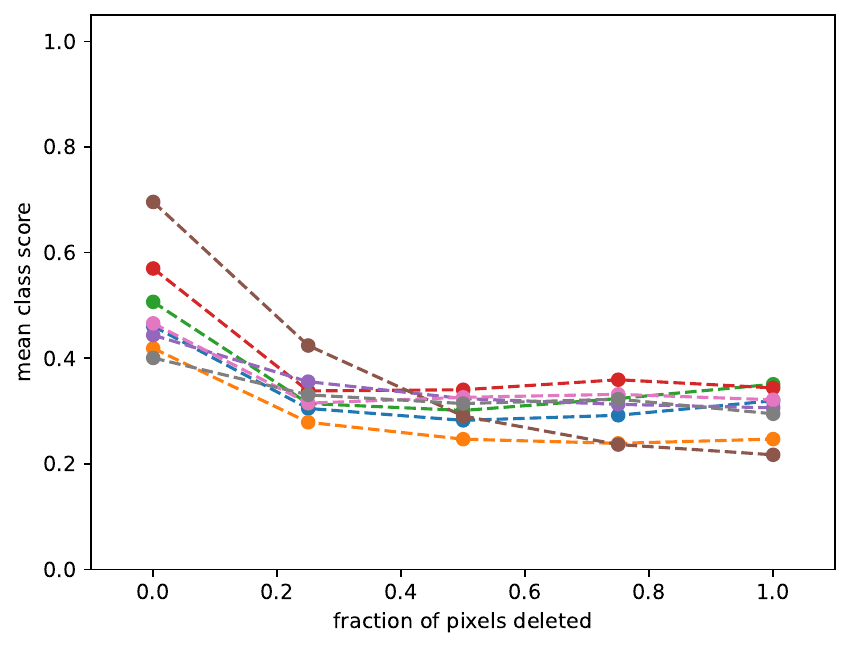}
    \end{subfigure}
    \begin{subfigure}{0.11\textwidth}
        \includegraphics[width=\linewidth]%
        {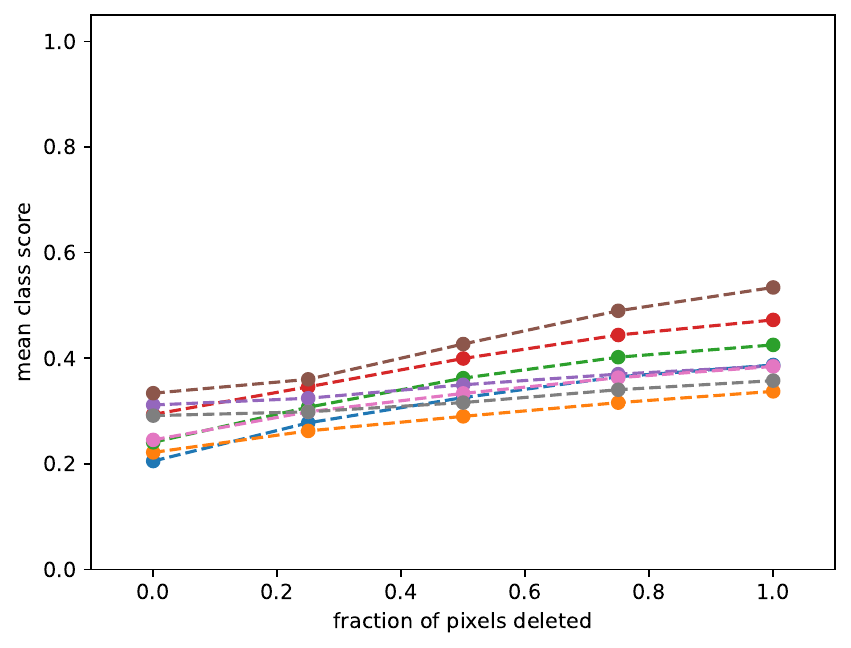}
    \end{subfigure}
    \begin{subfigure}{0.11\textwidth}
        \includegraphics[width=\linewidth]%
        {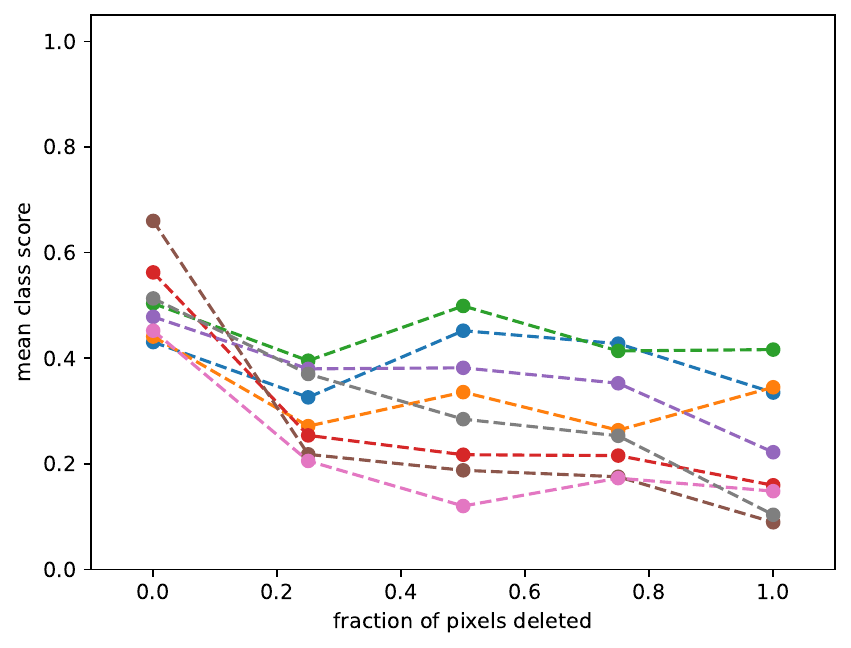}
    \end{subfigure}
    \begin{subfigure}{0.11\textwidth}
        \includegraphics[width=\linewidth]%
        {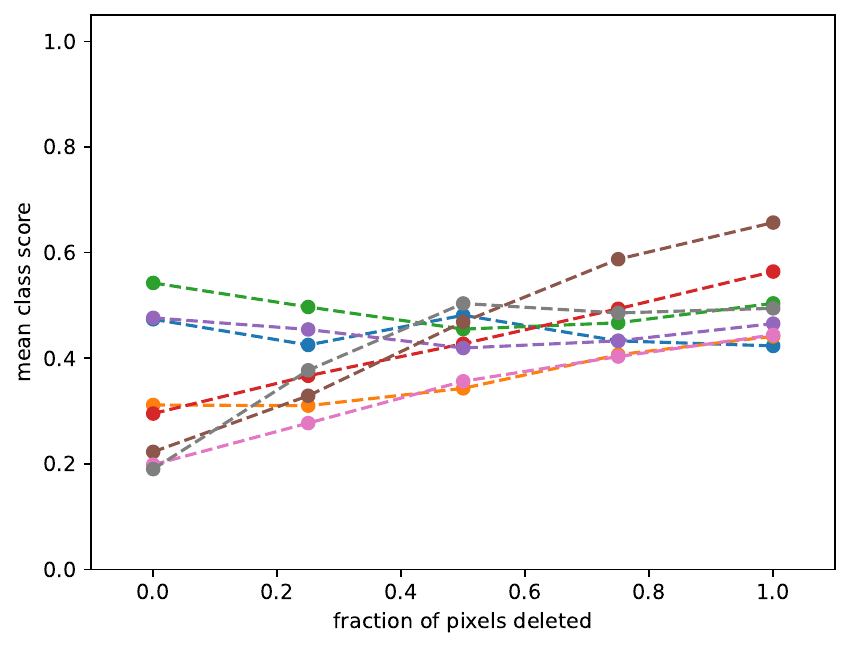}
    \end{subfigure}
    \begin{subfigure}{0.11\textwidth}
        \includegraphics[width=\linewidth]%
        {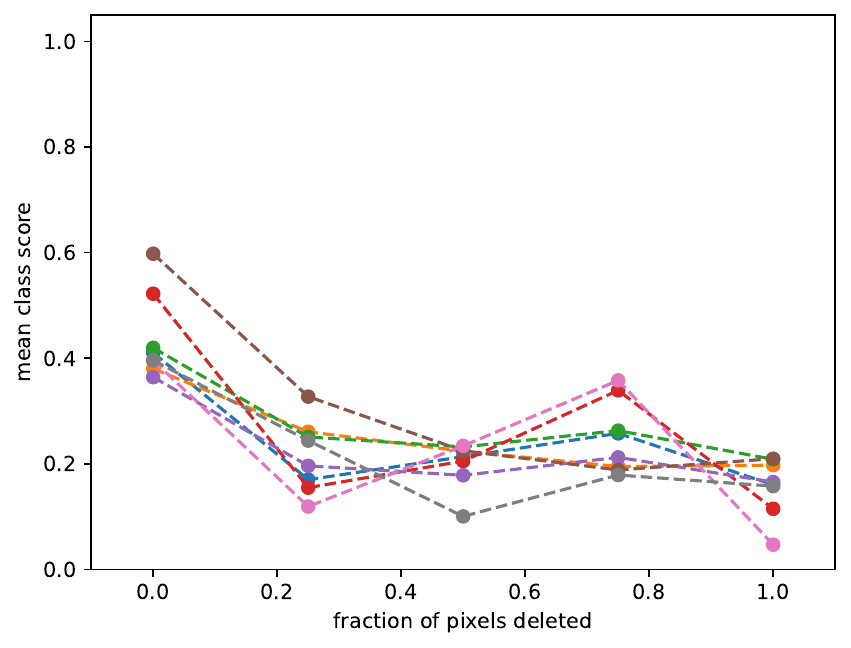}
    \end{subfigure}
    \begin{subfigure}{0.11\textwidth}
        \includegraphics[width=\linewidth]%
        {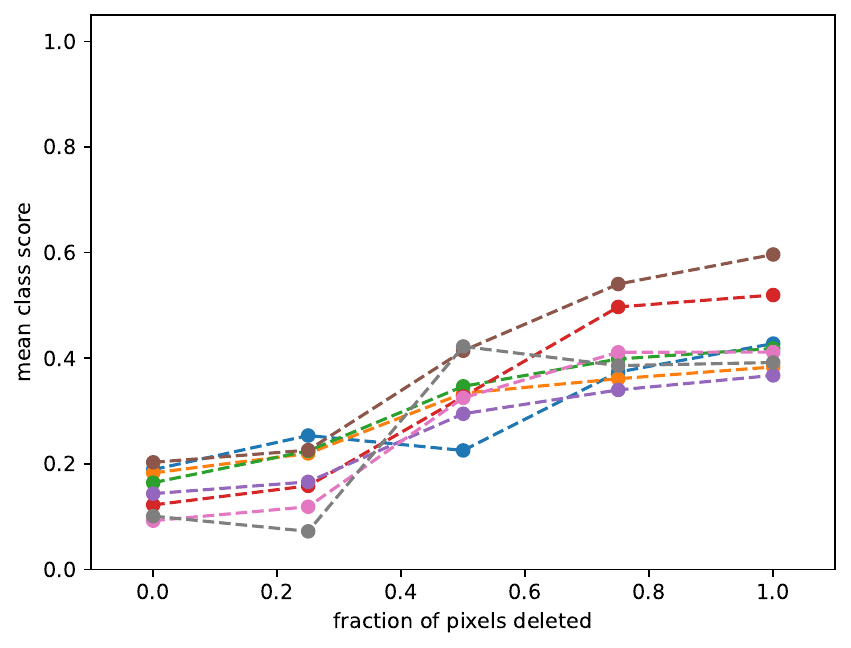}
    \end{subfigure}
    \begin{subfigure}{0.11\textwidth}
        \includegraphics[width=\linewidth]%
        {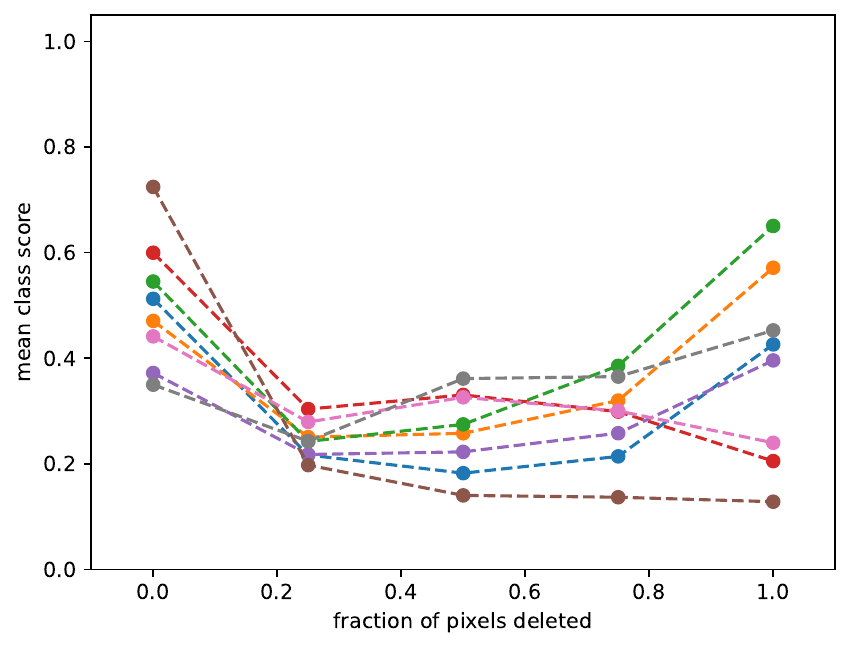}
    \end{subfigure}
    \begin{subfigure}{0.11\textwidth}
        \includegraphics[width=\linewidth]%
        {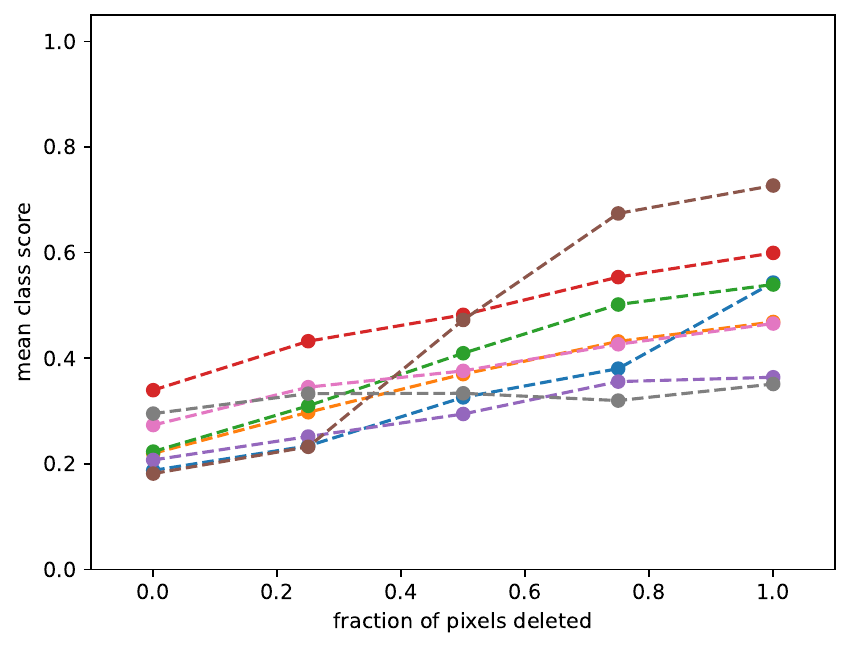}
    \end{subfigure}
    \\
    \begin{subfigure}{0.01\textwidth}
        \rotatebox[origin=c]{90}{\textbf{\fontsize{5}{4} \selectfont $\sigma$-IG}}
        \vspace*{0.5cm}
    \end{subfigure}
    \begin{subfigure}{0.11\textwidth}
        \includegraphics[width=\linewidth]%
        {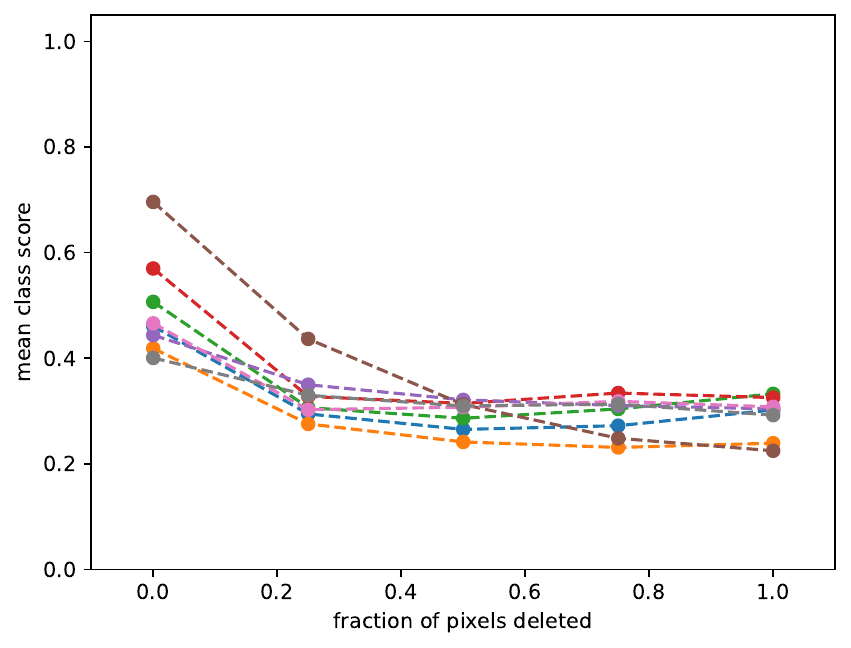}
    \end{subfigure}
    \begin{subfigure}{0.11\textwidth}
        \includegraphics[width=\linewidth]%
        {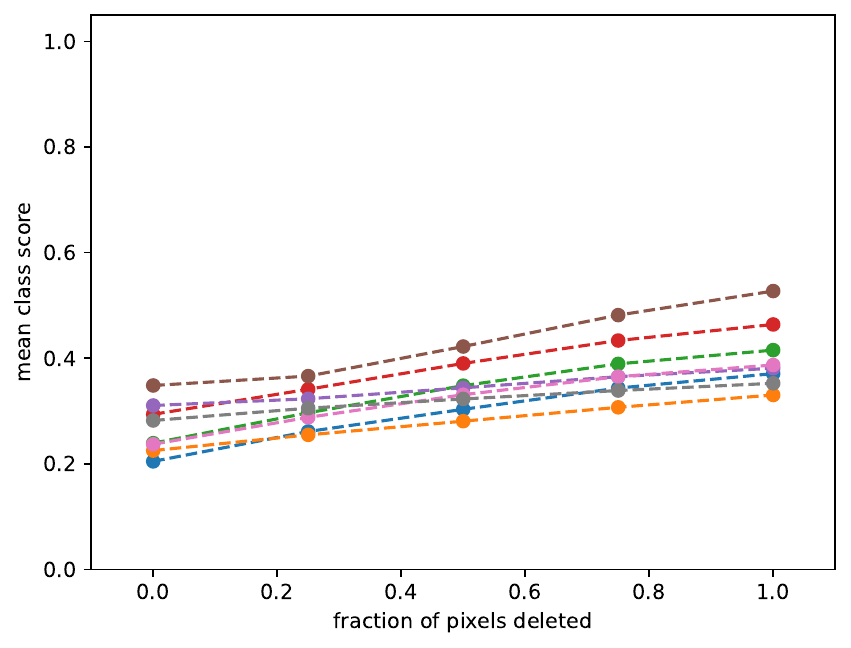}
    \end{subfigure}
    \begin{subfigure}{0.11\textwidth}
        \includegraphics[width=\linewidth]%
        {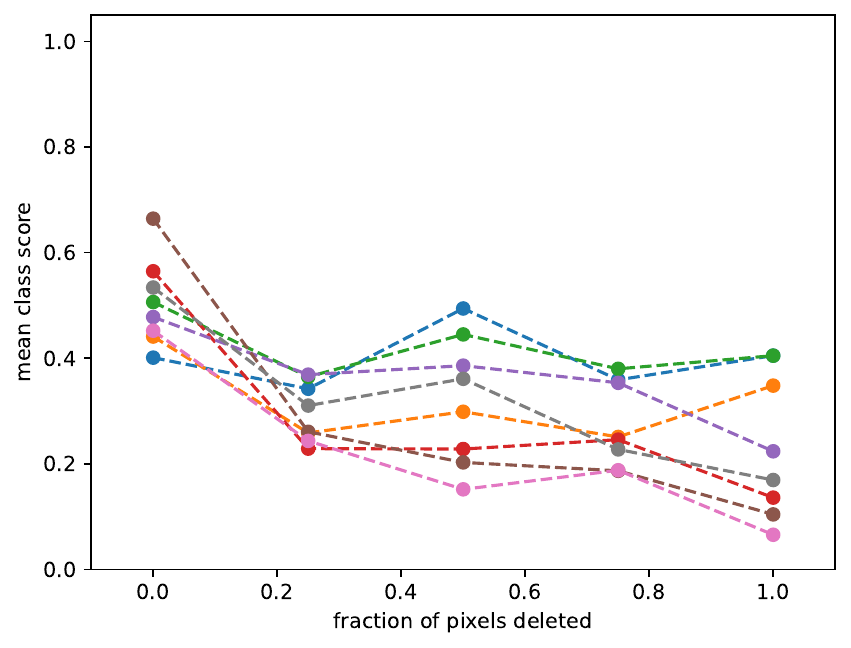}
    \end{subfigure}
    \begin{subfigure}{0.11\textwidth}
        \includegraphics[width=\linewidth]%
        {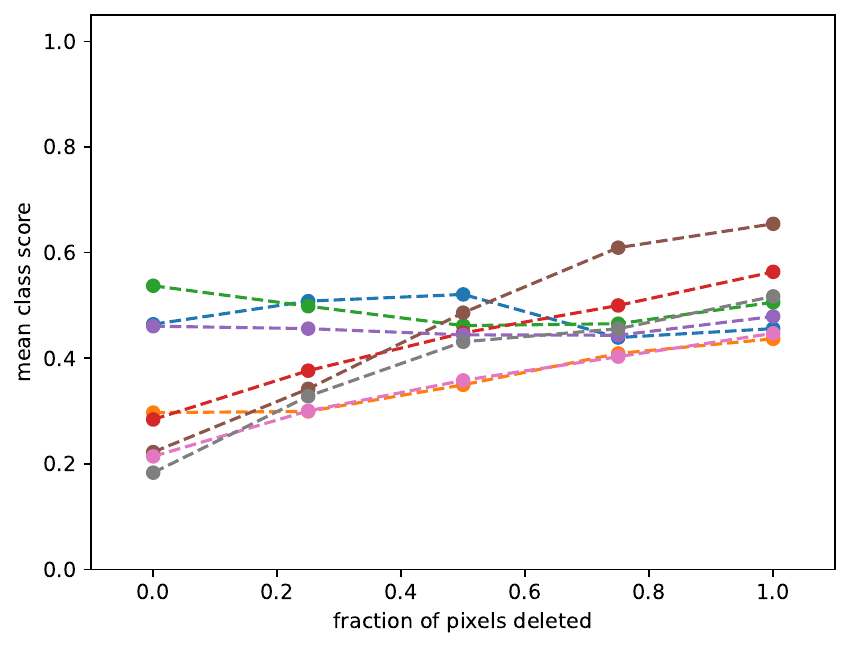}
    \end{subfigure}
    \begin{subfigure}{0.11\textwidth}
        \includegraphics[width=\linewidth]%
        {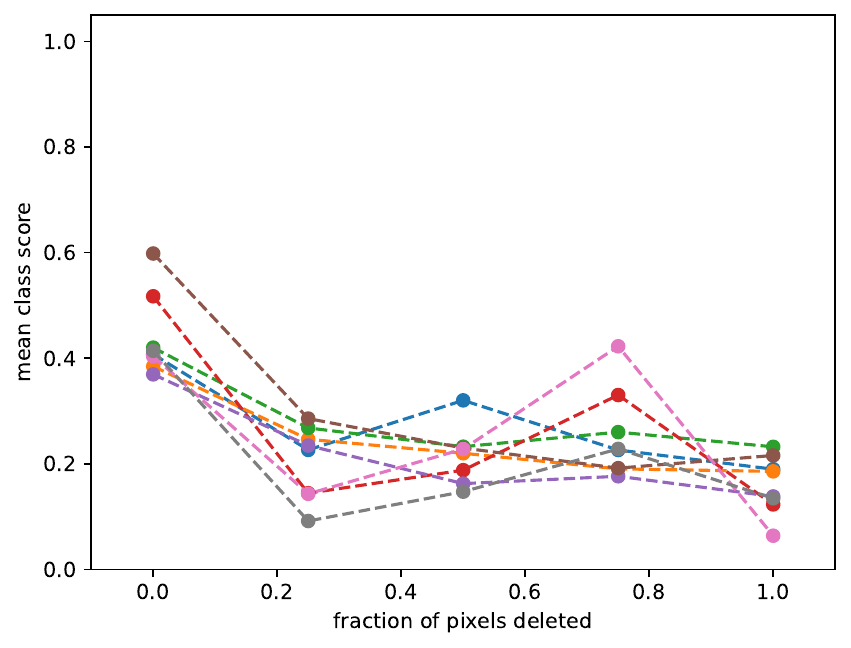}
    \end{subfigure}
    \begin{subfigure}{0.11\textwidth}
        \includegraphics[width=\linewidth]%
        {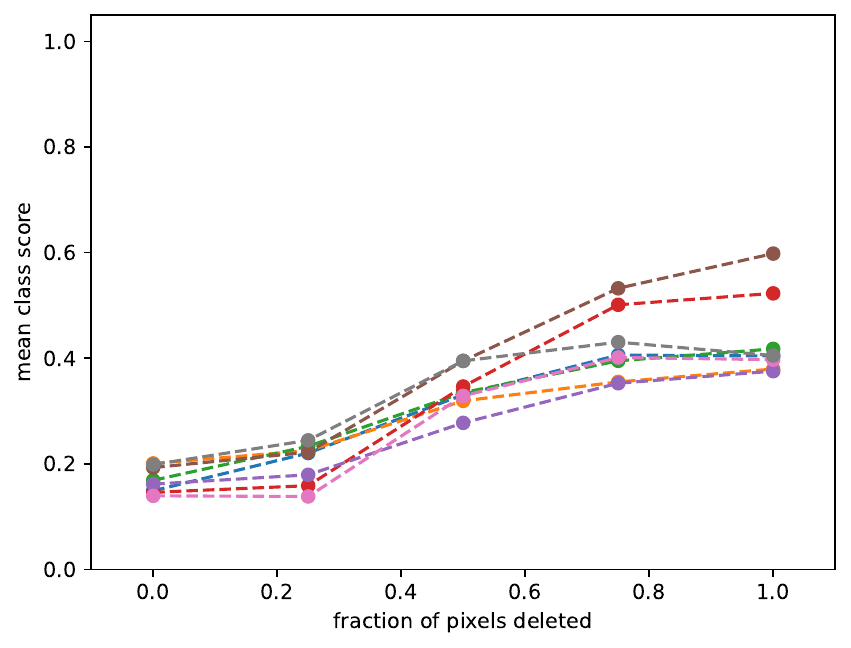}
    \end{subfigure}
    \begin{subfigure}{0.11\textwidth}
        \includegraphics[width=\linewidth]%
        {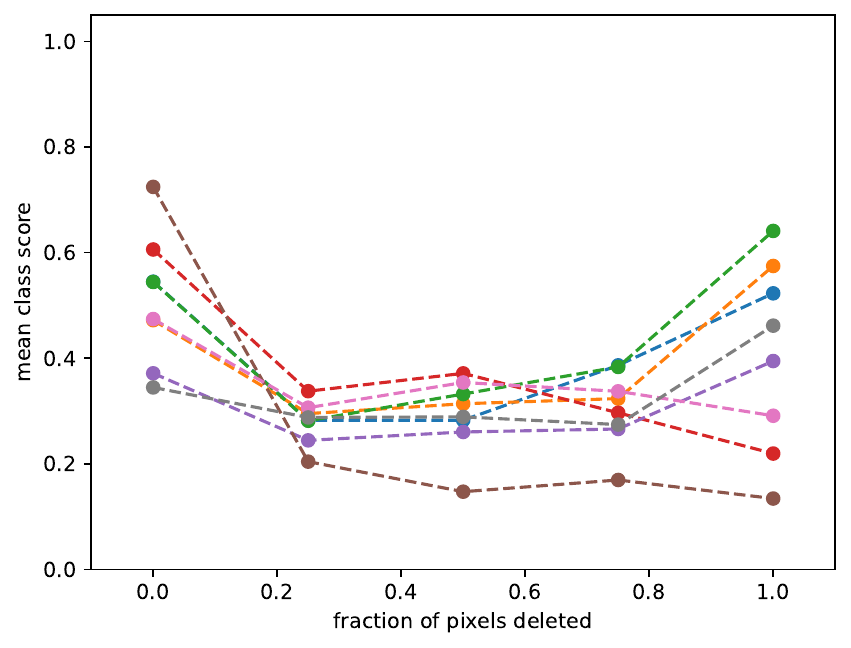}
    \end{subfigure}
    \begin{subfigure}{0.11\textwidth}
        \includegraphics[width=\linewidth]%
        {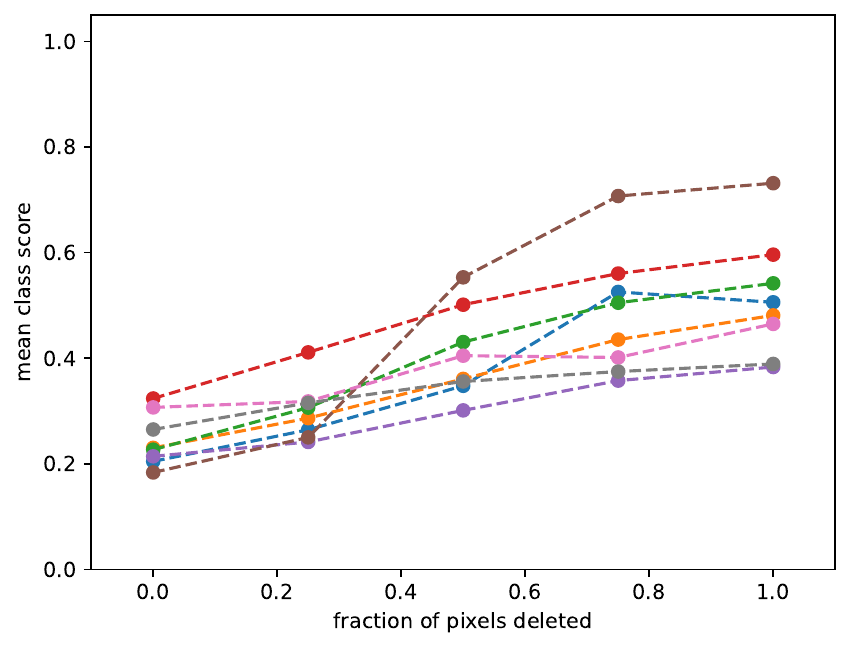}
    \end{subfigure}
    \\

    \begin{subfigure}{\textwidth}
        \centering
        \includegraphics[width=\textwidth]{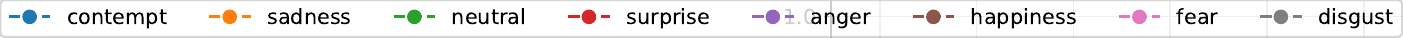}
    \end{subfigure}

    \caption{Pixel flipping curves for Deep Ensemble (columns 1-2), MC-Dropout (columns 3-4), MC-DropConnect (columns 5-6), and Flipout (columns 7-8) for FER+. The columns 1, 3, 5, and 7 depict the deletion curves whereas the columns 2, 4, 6, and 8 depict the insertion curves. From top to bottom, the rows show the following heatmaps and the explanation methods they correspond to: (i) $\mu$-GBP (ii) $\sigma$-GBP (iii) $\mu$-IG (iv) $\sigma$-IG.}
    \label{experiment_2_fde_fdo_fdc_fdf}
\end{figure*}

The Area Under Curve (AUCs) for the pixel insertion and deletion metrics for CIFAR-10 and FER+ are provided in Table \ref{cifar_fer_experiment_2}. Additionally, the class specific pixel insertion and deletion curves for FER+ are shown in Figure \ref{experiment_2_fde_fdo_fdc_fdf}.

\begin{figure*}[h]
    \centering
    \begin{tabular}{p{0.09\textwidth}p{0.22\textwidth}p{0.20\textwidth}p{0.20\textwidth}p{0.20\textwidth}}
         &  \footnotesize \textbf{Ensemble}  & \footnotesize \textbf{MC-Dropout} & \footnotesize \textbf{MC-DropConnect} & \footnotesize \textbf{Flipout} \\
    \end{tabular}
    
    \begin{subfigure}{0.01\textwidth}
        \rotatebox[origin=c]{90}{\textbf{\fontsize{5}{4} \selectfont $\mu$-GBP}}
        \vspace*{0.5cm}
    \end{subfigure}
    \begin{subfigure}{0.11\textwidth}
        \centering
        \textbf{\fontsize{5}{4} \selectfont Deletion}\\
        \includegraphics[width=\linewidth]
        {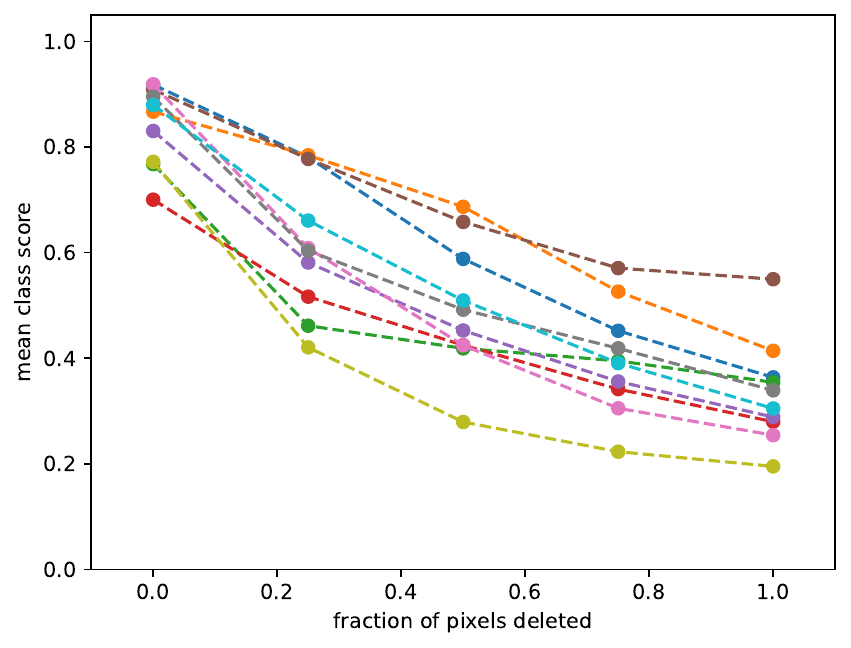}
    \end{subfigure}
    \begin{subfigure}{0.11\textwidth}
        \centering
        \textbf{\fontsize{5}{4} \selectfont Insertion}
        \includegraphics[width=\linewidth]%
        {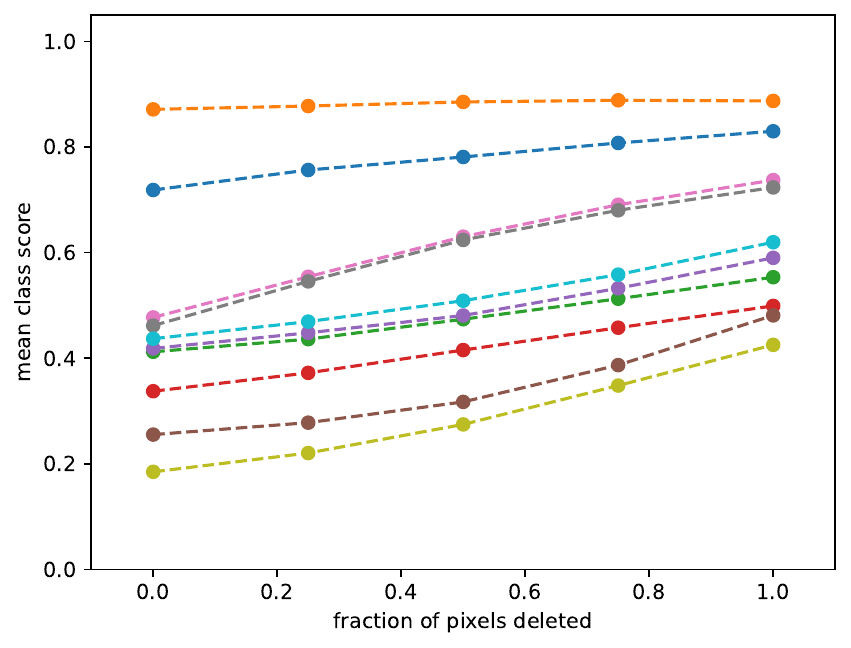}
    \end{subfigure}
    \begin{subfigure}{0.11\textwidth}
        \centering
        \textbf{\fontsize{5}{4} \selectfont Deletion}
        \includegraphics[width=\linewidth]%
        {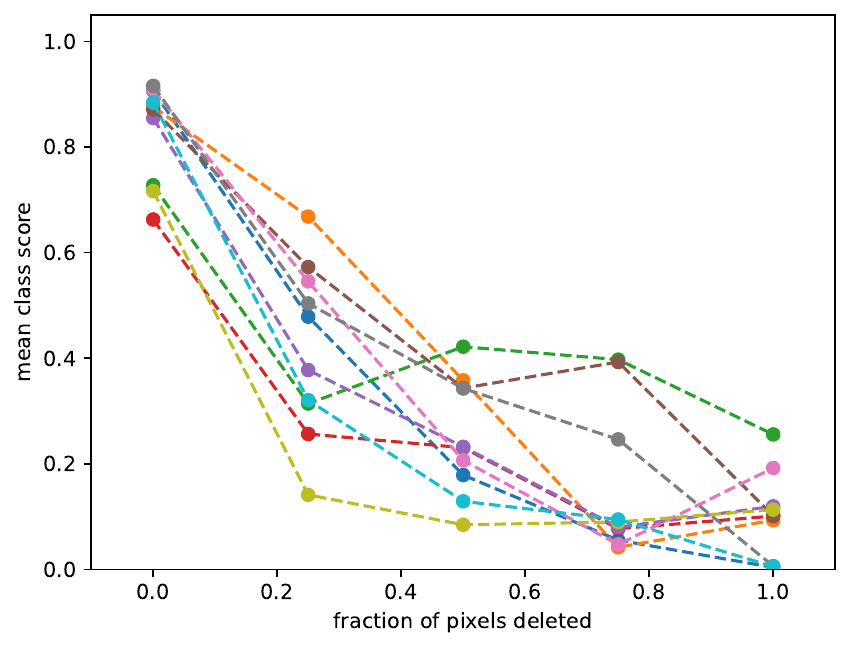}
    \end{subfigure}
    \begin{subfigure}{0.11\textwidth}
        \centering
        \textbf{\fontsize{5}{4} \selectfont Insertion}
        \includegraphics[width=\linewidth]%
        {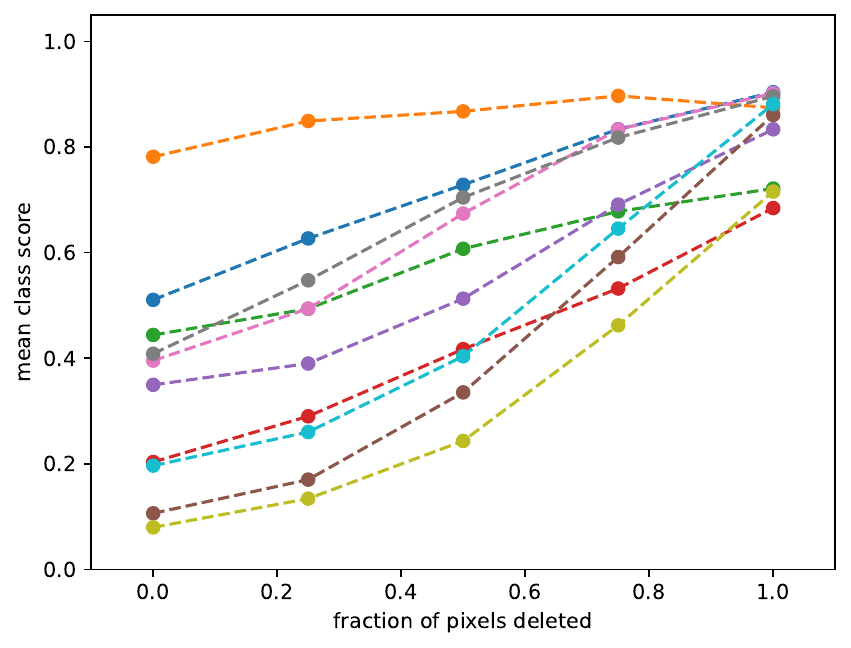}
    \end{subfigure}
    \begin{subfigure}{0.11\textwidth}
        \centering
        \textbf{\fontsize{5}{4} \selectfont Deletion}
        \includegraphics[width=\linewidth]%
        {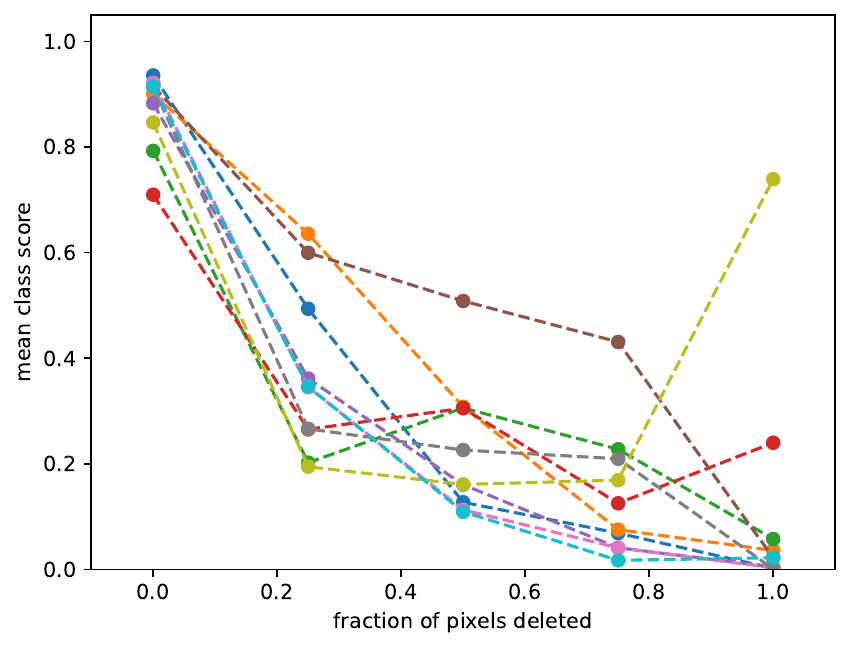}
    \end{subfigure}
    \begin{subfigure}{0.11\textwidth}
        \centering
        \textbf{\fontsize{5}{4} \selectfont Insertion}
        \includegraphics[width=\linewidth]%
        {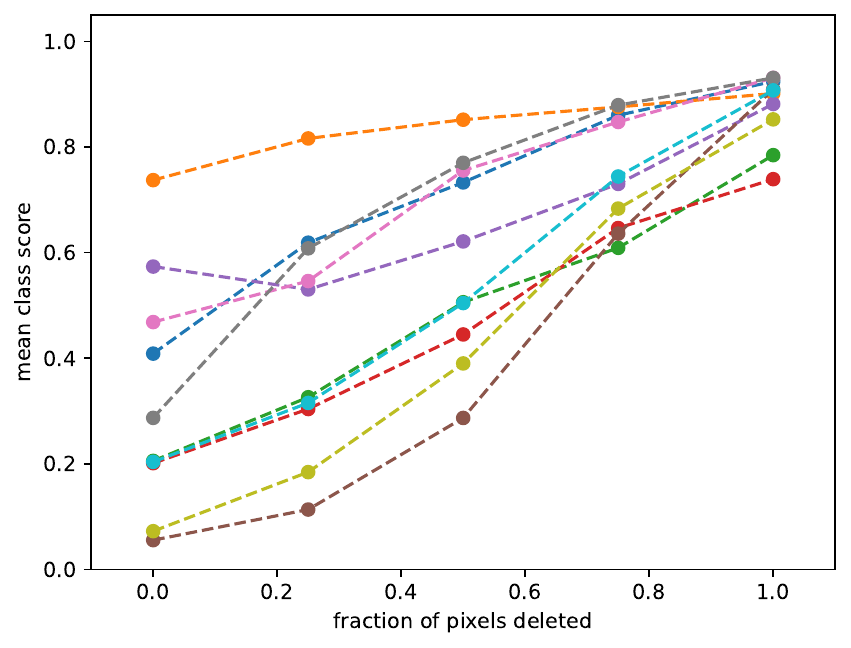}
    \end{subfigure}
    \begin{subfigure}{0.11\textwidth}
        \centering
        \textbf{\fontsize{5}{4} \selectfont Deletion}
        \includegraphics[width=\linewidth]%
        {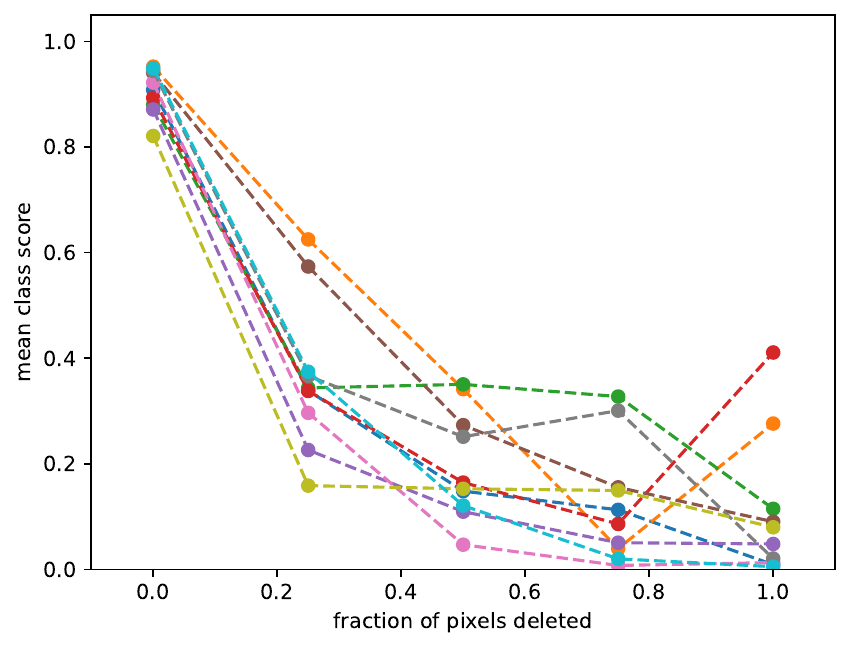}
    \end{subfigure}
    \begin{subfigure}{0.11\textwidth}
        \centering
        \textbf{\fontsize{5}{4} \selectfont Insertion}
        \includegraphics[width=\linewidth]%
        {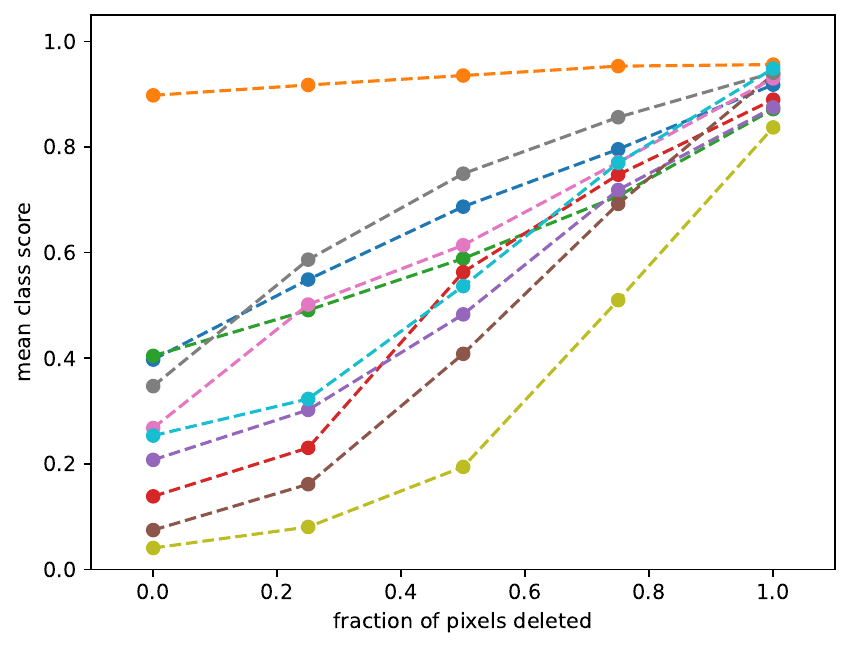}
    \end{subfigure} \\
    
    \begin{subfigure}{0.01\textwidth}
        \rotatebox[origin=c]{90}{\textbf{\fontsize{5}{4} \selectfont $\sigma$-GBP}}
        \vspace*{0.5cm}
    \end{subfigure}
    \begin{subfigure}{0.11\textwidth}
        \includegraphics[width=\linewidth]{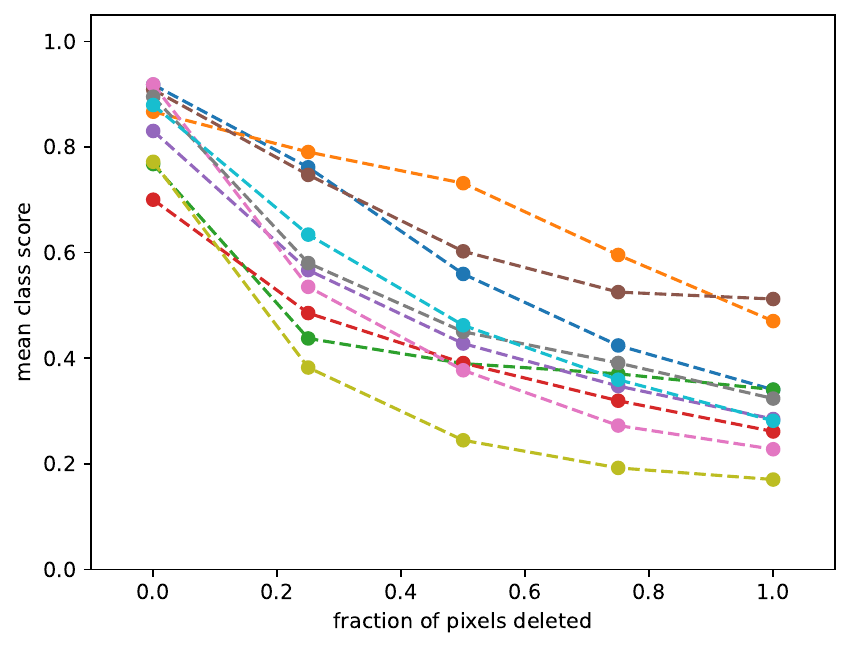}
    \end{subfigure}
    \begin{subfigure}{0.11\textwidth}
        \includegraphics[width=\linewidth]%
        {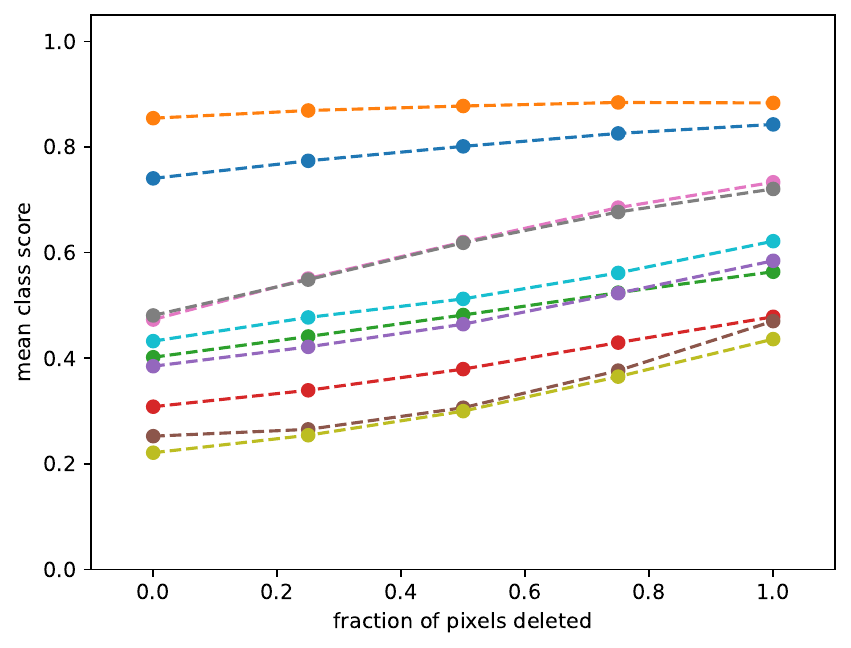}
    \end{subfigure}
    \begin{subfigure}{0.11\textwidth}
        \includegraphics[width=\linewidth]%
        {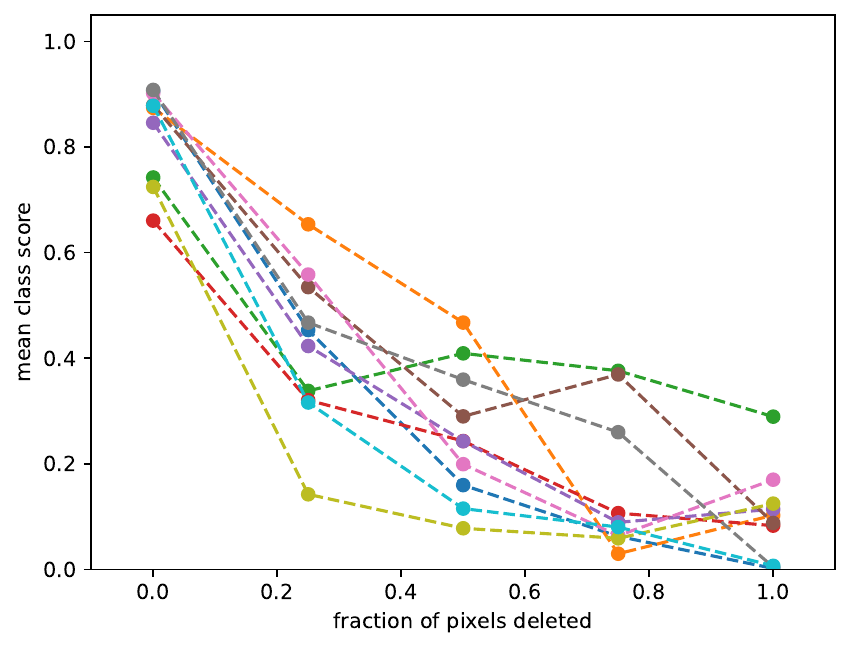}
    \end{subfigure}
    \begin{subfigure}{0.11\textwidth}
        \includegraphics[width=\linewidth]%
        {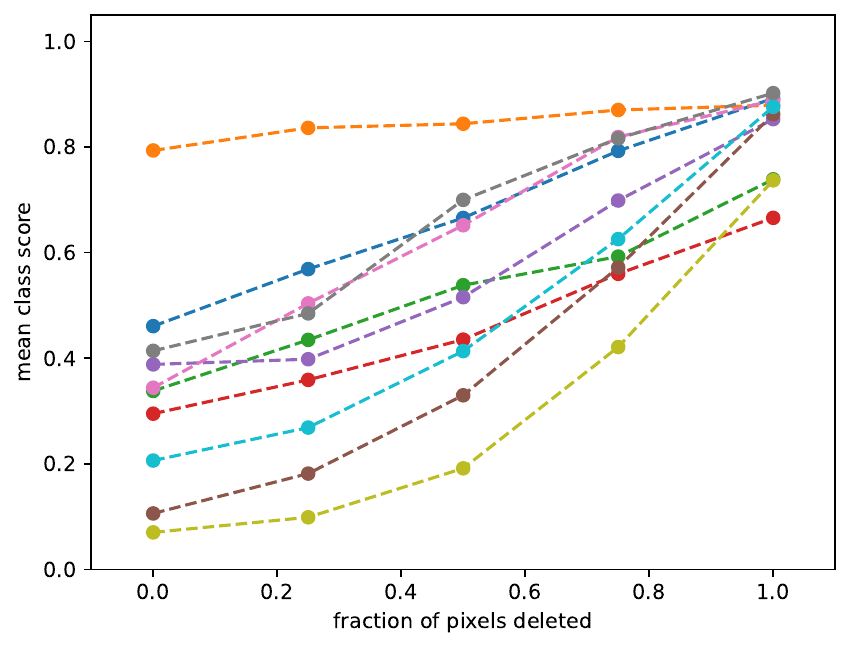}
    \end{subfigure}
    \begin{subfigure}{0.11\textwidth}
        \includegraphics[width=\linewidth]%
        {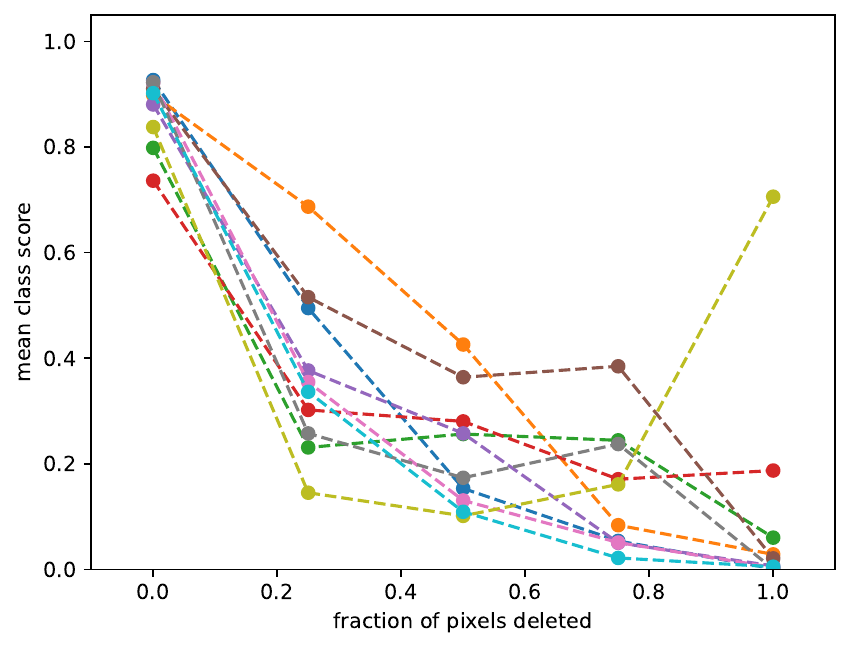}
    \end{subfigure}
    \begin{subfigure}{0.11\textwidth}
        \includegraphics[width=\linewidth]%
        {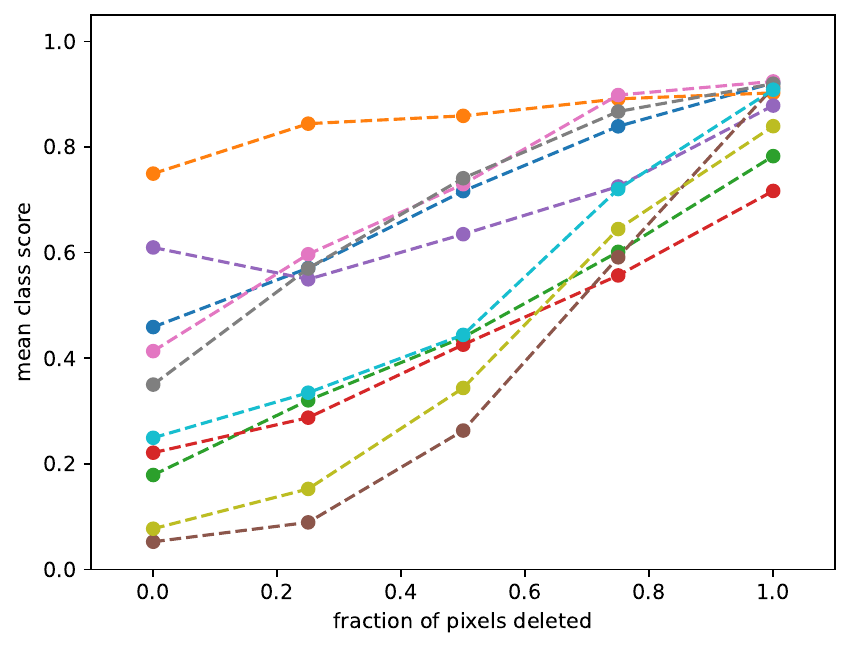}
    \end{subfigure}
    \begin{subfigure}{0.11\textwidth}
        \includegraphics[width=\linewidth]%
        {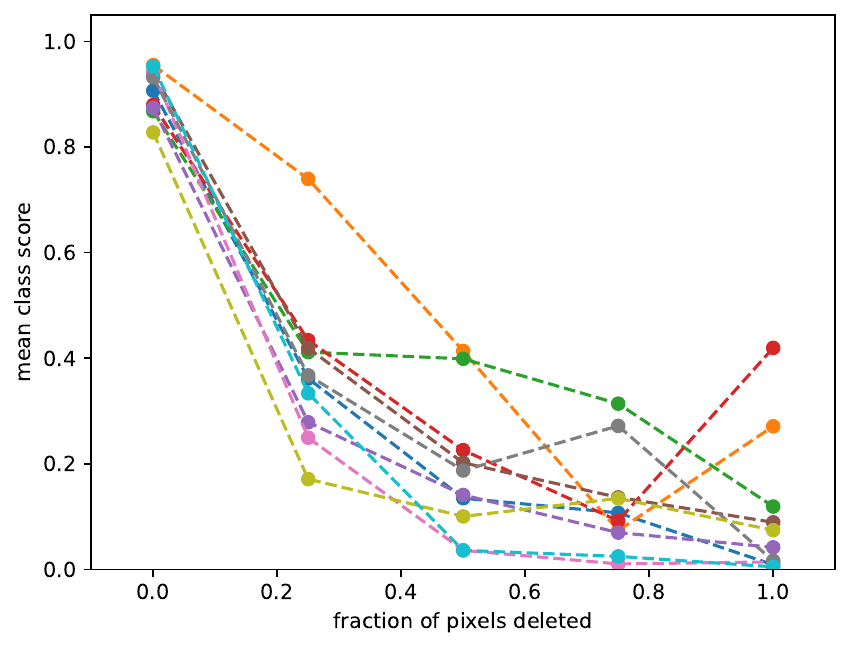}
    \end{subfigure}
    \begin{subfigure}{0.11\textwidth}
        \includegraphics[width=\linewidth]%
        {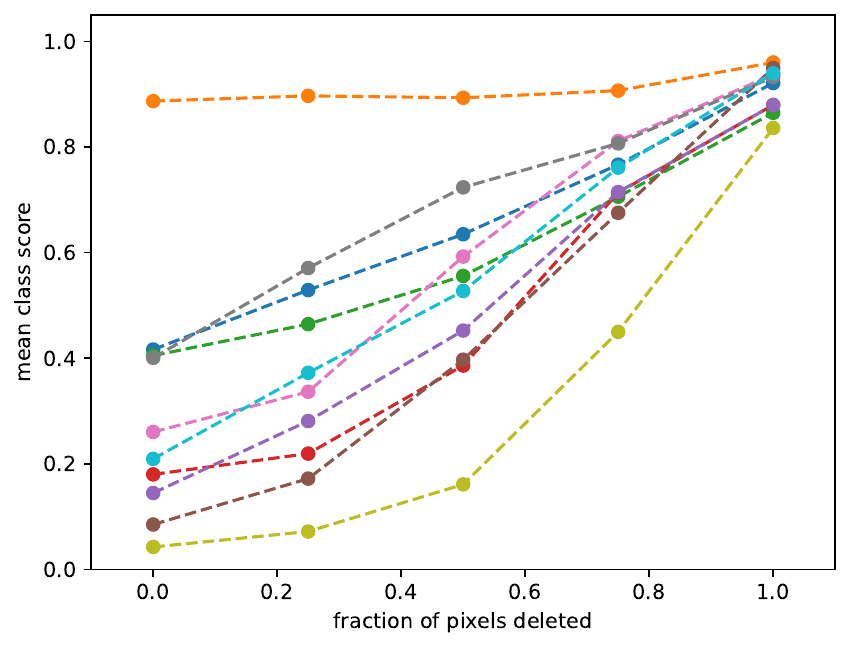}
    \end{subfigure}
    \\

    \begin{subfigure}{0.01\textwidth}
        \rotatebox[origin=c]{90}{\textbf{\fontsize{5}{4} \selectfont $\mu$-IG}}
        \vspace*{0.5cm}
    \end{subfigure}
    \begin{subfigure}{0.11\textwidth}
        \includegraphics[width=\linewidth]
        {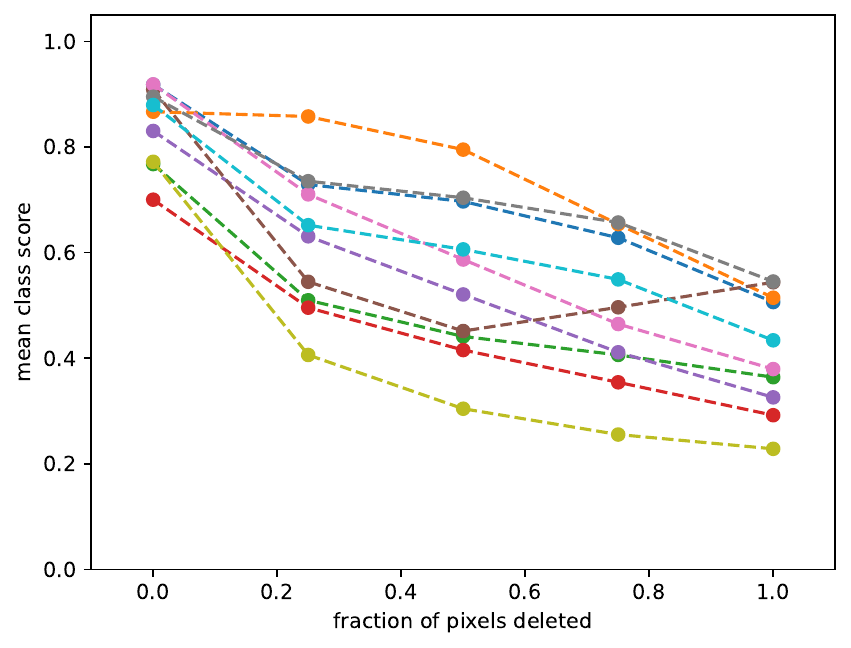}
    \end{subfigure}
    \begin{subfigure}{0.11\textwidth}
        \includegraphics[width=\linewidth]%
        {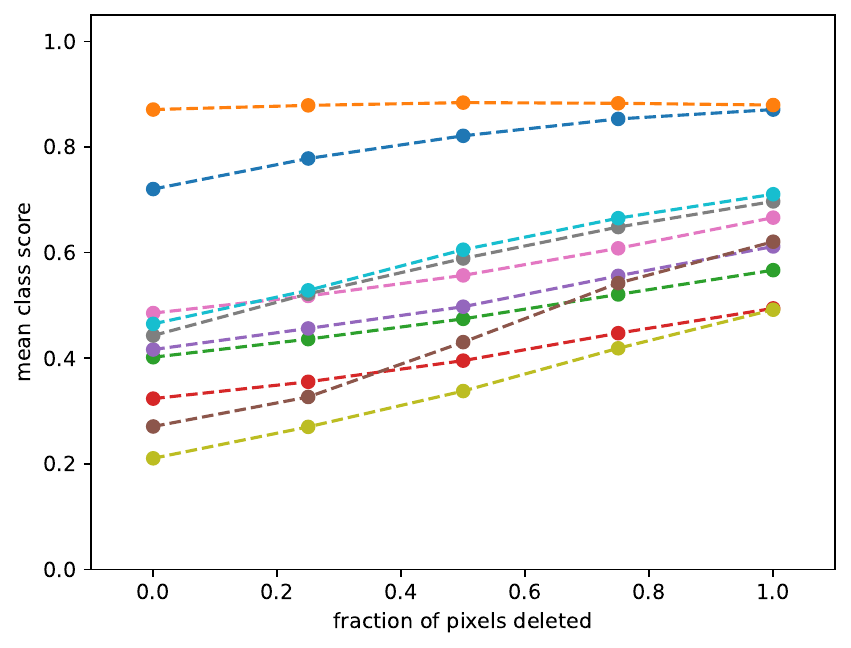}
    \end{subfigure}
    \begin{subfigure}{0.11\textwidth}
        \includegraphics[width=\linewidth]%
        {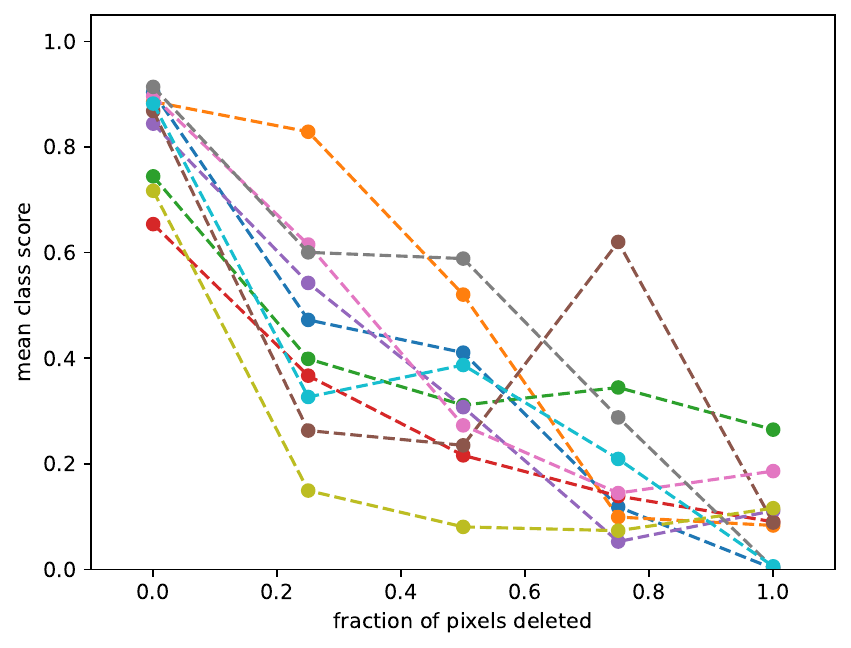}
    \end{subfigure}
    \begin{subfigure}{0.11\textwidth}
        \includegraphics[width=\linewidth]%
        {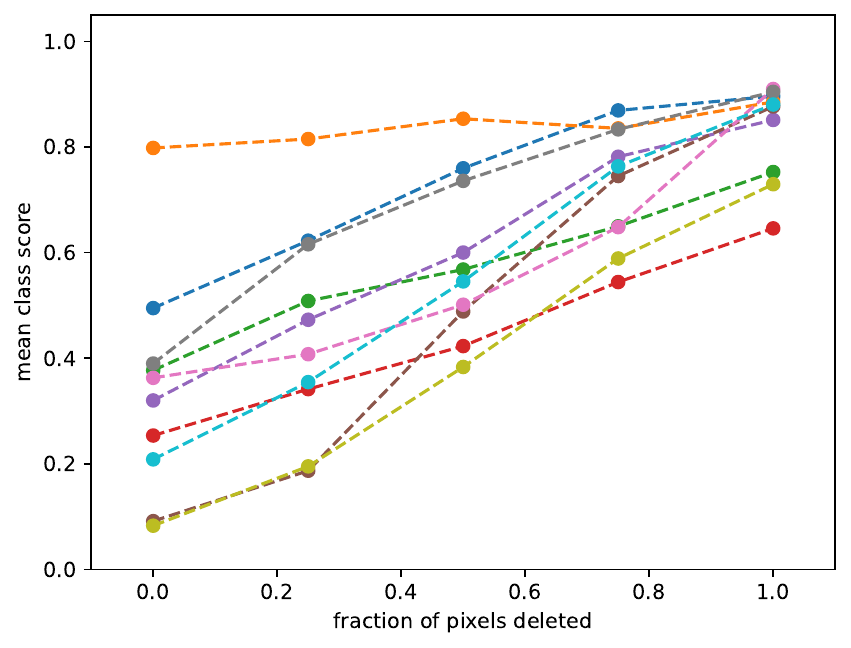}
    \end{subfigure}
    \begin{subfigure}{0.11\textwidth}
        \includegraphics[width=\linewidth]%
        {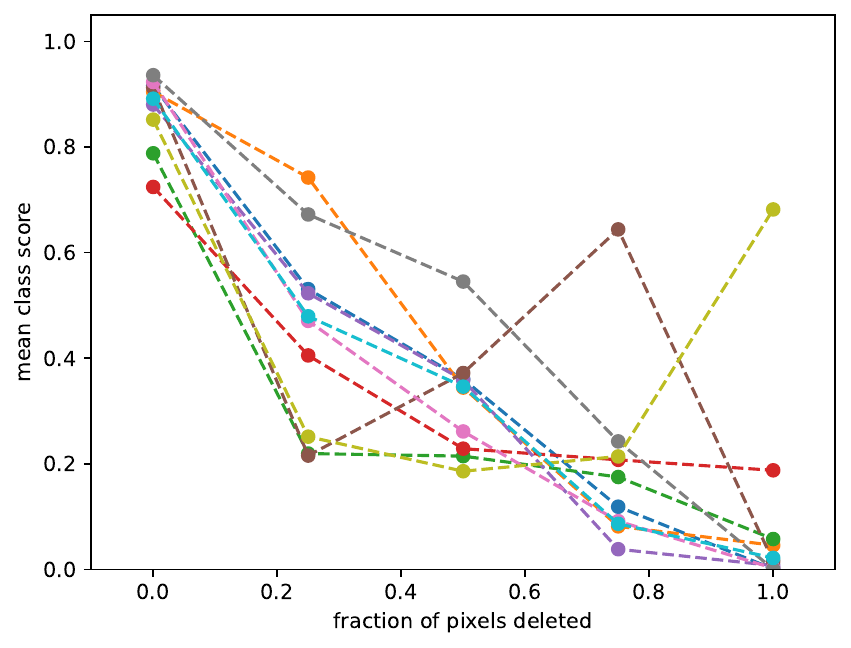}
    \end{subfigure}
    \begin{subfigure}{0.11\textwidth}
        \includegraphics[width=\linewidth]%
        {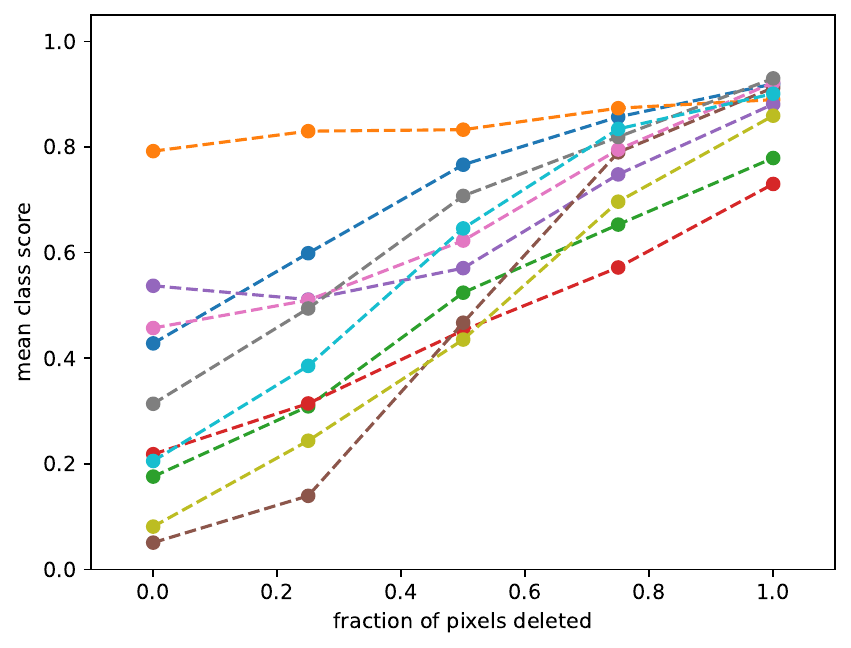}
    \end{subfigure}
    \begin{subfigure}{0.11\textwidth}
        \includegraphics[width=\linewidth]%
        {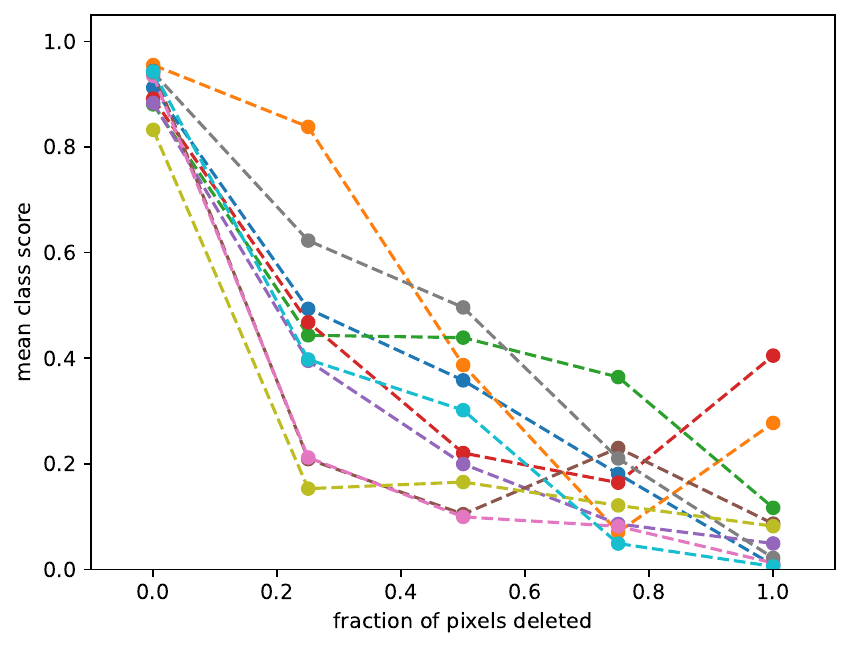}
    \end{subfigure}
    \begin{subfigure}{0.11\textwidth}
        \includegraphics[width=\linewidth]%
        {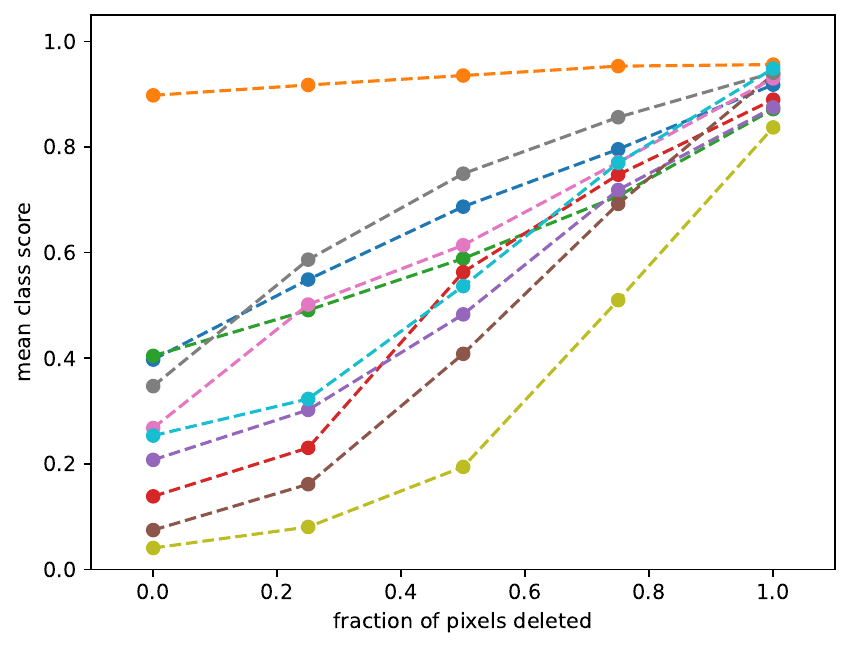}
    \end{subfigure}
    \\
    \begin{subfigure}{0.01\textwidth}
        \rotatebox[origin=c]{90}{\textbf{\fontsize{5}{4} \selectfont $\sigma$-IG}}
        \vspace*{0.5cm}
    \end{subfigure}
    \begin{subfigure}{0.11\textwidth}
        \includegraphics[width=\linewidth]
        {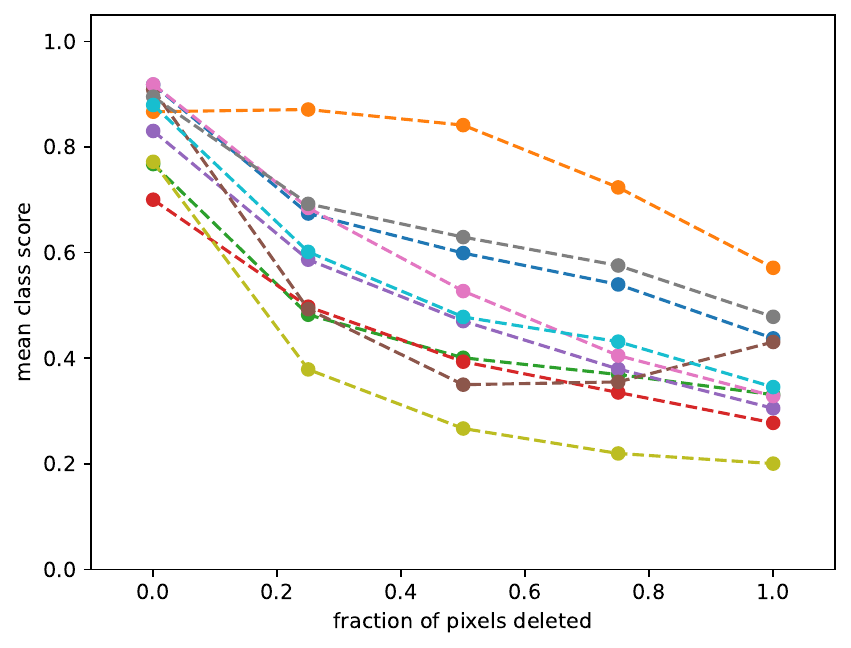}
    \end{subfigure}
    \begin{subfigure}{0.11\textwidth}
        \includegraphics[width=\linewidth]%
        {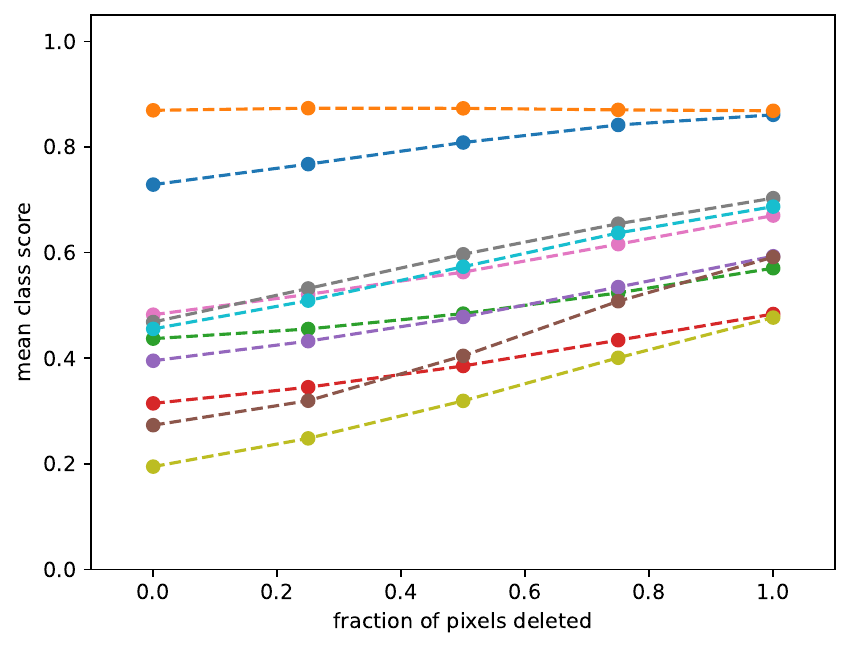}
    \end{subfigure}
    \begin{subfigure}{0.11\textwidth}
        \includegraphics[width=\linewidth]%
        {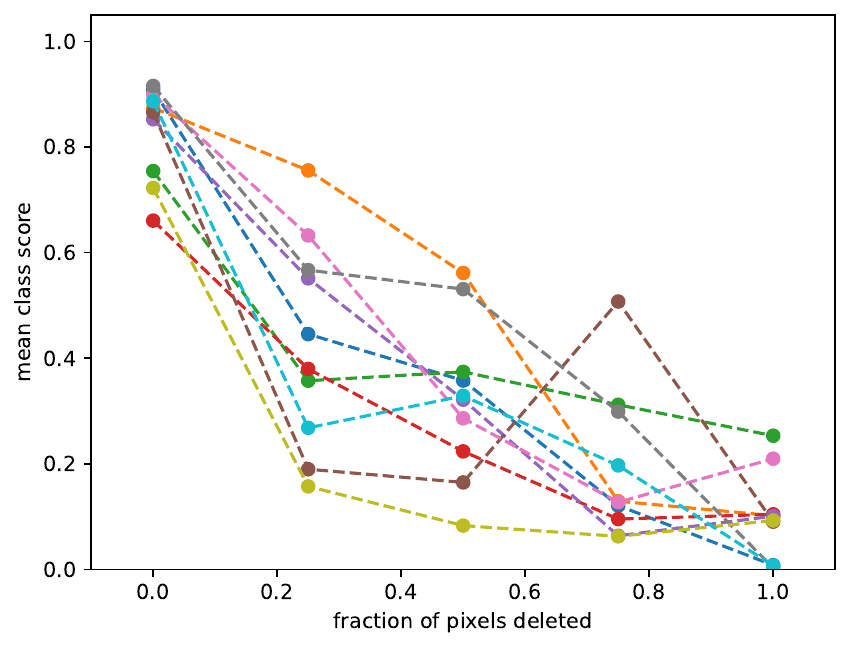}
    \end{subfigure}
    \begin{subfigure}{0.11\textwidth}
        \includegraphics[width=\linewidth]%
        {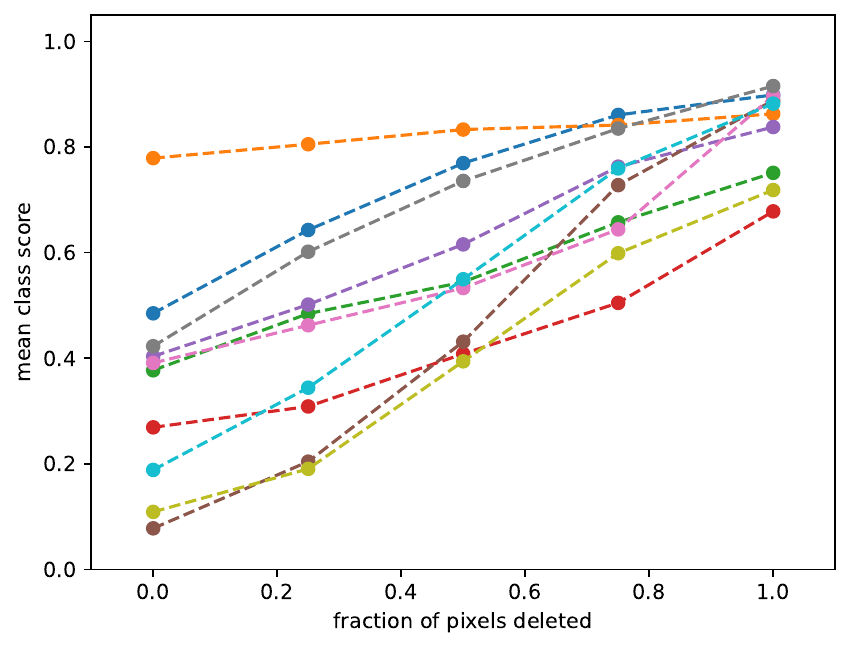}
    \end{subfigure}
    \begin{subfigure}{0.11\textwidth}
        \includegraphics[width=\linewidth]%
        {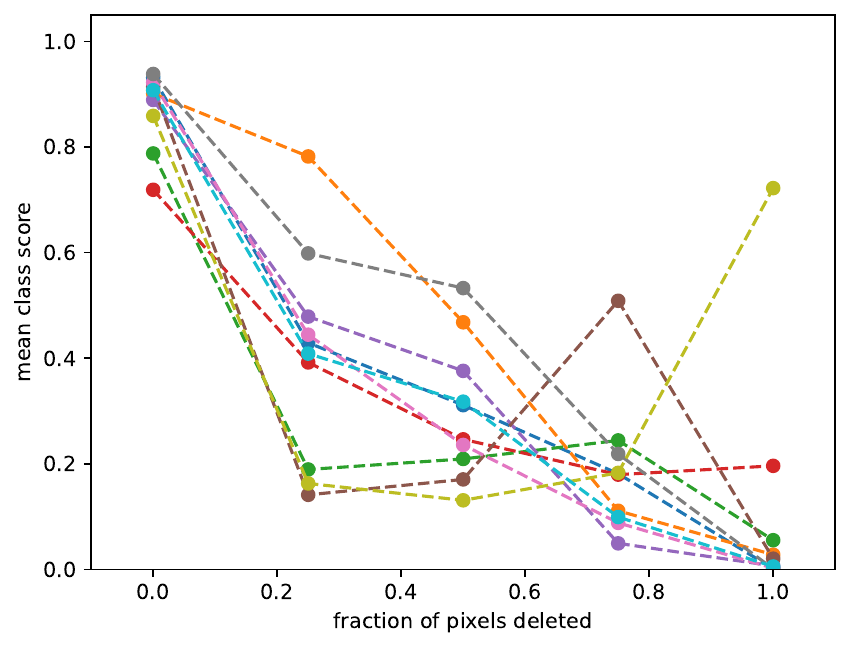}
    \end{subfigure}
    \begin{subfigure}{0.11\textwidth}
        \includegraphics[width=\linewidth]%
        {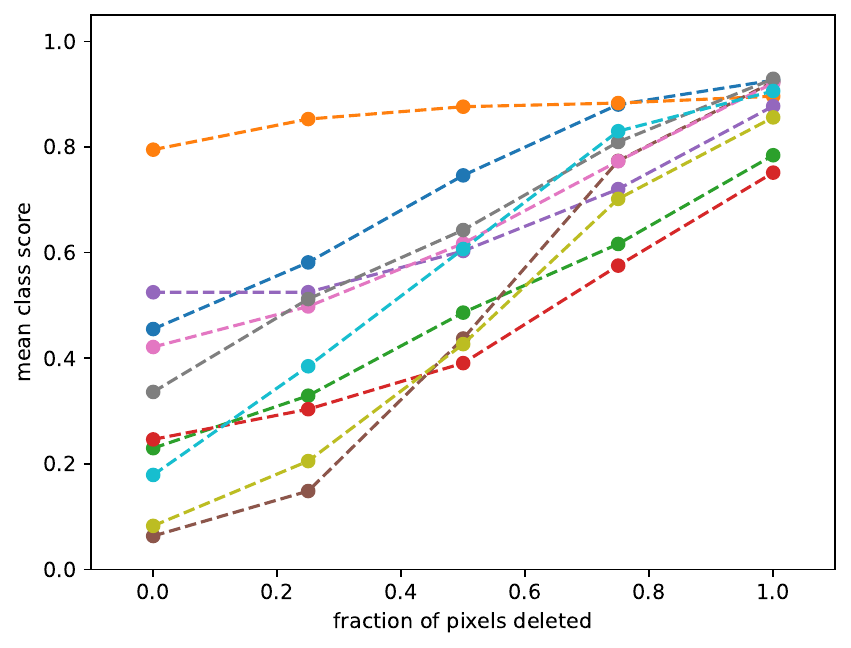}
    \end{subfigure}
    \begin{subfigure}{0.11\textwidth}
        \includegraphics[width=\linewidth]%
        {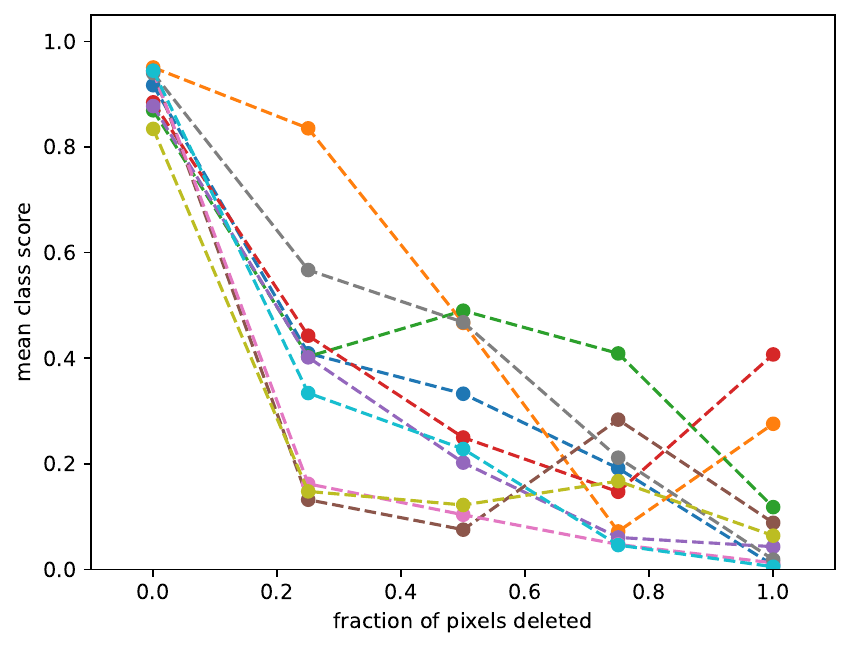}
    \end{subfigure}
    \begin{subfigure}{0.11\textwidth}
        \includegraphics[width=\linewidth]%
        {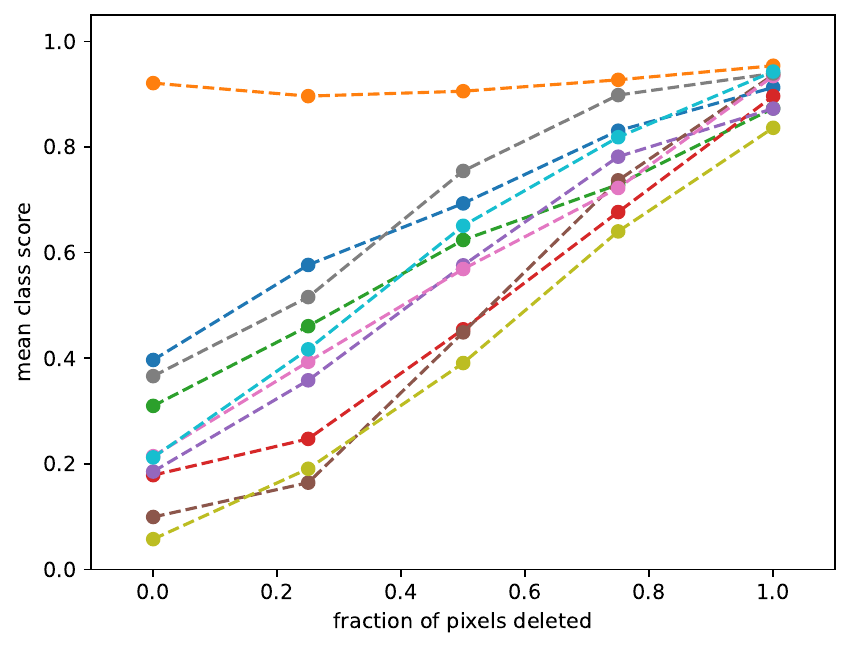}
    \end{subfigure}
    \\
    \begin{subfigure}{\textwidth}
        \centering
        \includegraphics[width=\textwidth]{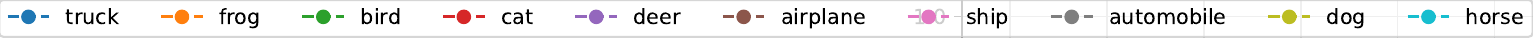}
    \end{subfigure}

    \caption{Pixel flipping curves for Deep Ensemble (columns 1-2), MC-Dropout (columns 3-4), MC-DropConnect (columns 5-6), and Flipout (columns 7-8) for CIFAR-10. The columns 1, 3, 5, and 7 depict the deletion curves whereas the columns 2, 4, 6, and 8 depict the insertion curves. From top to bottom, the rows show the following heatmaps and the explanation methods they correspond to: (i) $\mu$-GBP (ii) $\sigma$-GBP (iii) $\mu$-IG (iv) $\sigma$-IG.}
    \label{experiment_2_cde_cdo_cdc_cdf}
\end{figure*}

\begin{table*}
\setlength{\tabcolsep}{0.5pt} %
\renewcommand{\tiny}{\normalsize} %
\begin{tabularx}{\linewidth}{>{\centering\arraybackslash}X *{72}{l}}%
& \multicolumn{40}{l}{\fontsize{6}{4} \textbf{CIFAR10}} & \multicolumn{14}{l}{\fontsize{6}{4} \textbf{FER}+}\\
\toprule
\fontsize{6}{4} \selectfont  & 
\multicolumn{4}{c}{ \textbf{\fontsize{5}{4} \selectfont \makecell{truck\\(+  -)}}} & 
\multicolumn{4}{c}{ \textbf{\fontsize{5}{4} \selectfont \makecell{frog\\(+  -)}}} & 
\multicolumn{4}{c}{ \textbf{\fontsize{5}{4} \selectfont \makecell{bird\\(+  -)}}} & 
\multicolumn{4}{c}{ \textbf{\fontsize{5}{4} \selectfont \makecell{cat\\(+  -)}}} & 
\multicolumn{4}{c}{ \textbf{\fontsize{5}{4} \selectfont \makecell{deer\\(+  -)}}} & 
\multicolumn{4}{c}{ \textbf{\fontsize{5}{4} \selectfont \makecell{plane\\(+  -)}}} & 
\multicolumn{4}{c}{ \textbf{\fontsize{5}{4} \selectfont \makecell{ship\\(+  -)}}} & 
\multicolumn{4}{c}{ \textbf{\fontsize{5}{4} \selectfont \makecell{auto\\(+  -)}}} &
\multicolumn{4}{c}{ \textbf{\fontsize{5}{4} \selectfont \makecell{dog\\(+  -)}}} & 
\multicolumn{4}{c}{ \textbf{\fontsize{5}{4} \selectfont \makecell{horse\\(+  -)}}}   & 
\multicolumn{4}{c}{ \textbf{\fontsize{5}{4} \selectfont \makecell{contempt\\(+  -)}}} & 
\multicolumn{4}{c}{ \textbf{\fontsize{5}{4} \selectfont \makecell{sadness\\(+  -)}}} & 
\multicolumn{4}{c}{ \textbf{\fontsize{5}{4} \selectfont \makecell{neutral\\(+  -)}}} & 
\multicolumn{4}{c}{ \textbf{\fontsize{5}{4} \selectfont \makecell{surprise\\(+  -)}}} & 
\multicolumn{4}{c}{ \textbf{\fontsize{5}{4} \selectfont \makecell{anger\\(+  -)}}} & 
\multicolumn{4}{c}{ \textbf{\fontsize{5}{4} \selectfont \makecell{happy\\(+  -)}}} & 
\multicolumn{4}{c}{ \textbf{\fontsize{5}{4} \selectfont \makecell{fear\\(+  -)}}} & 
\multicolumn{4}{c}{ \textbf{\fontsize{5}{4} \selectfont \makecell{disgust\\(+  -)}}}
\\
\bottomrule
\end{tabularx}
\caption{Class-wise AUCs for the pixel flipping plots for CIFAR-10 (columns 1-10) and FER+ (columns 11-18). The \textbf{+} and the \textbf{-} denote the pixel insertion and pixel deletion metrics respectively. The combination of the uncertainty estimation methods (Deep Ensemble, MC-Dropout, MC-DropConnect and Flipout, explanation methods (Guided BackPropagation (\textbf{GBP}) and Integrated Gradients (\textbf{IG})), and type of explanation heatmap (mean (\textbf{$\mu$}) and standard deviation (\textbf{$\sigma$})) are listed along the rows. }
\label{cifar_fer_experiment_2}
\end{table*}

\section{Additional Example Explanations}

\begin{figure}[t]
    \centering
    \begin{varwidth}{0.5\textwidth}
    
\begin{tabular}
    {
    p{0.001\textwidth}
    p{0.1\textwidth} %
    p{0.1\textwidth}
    p{0.1\textwidth}
    p{0.1\textwidth}
    p{0.1\textwidth}
    p{0.1\textwidth} %
    p{0.1\textwidth} %
    }
    \multicolumn{1}{p{0.001\textwidth}}{\textbf{}}   
    &
    \multicolumn{1}{p{0.105\textwidth}}{} 
    &
    \multicolumn{1}{p{0.105\textwidth}}{}   
    &
    \multicolumn{1}{p{0.105\textwidth}}{}  &
    \multicolumn{1}{p{0.105\textwidth}}{}
    &
    \multicolumn{1}{p{0.105\textwidth}}{}  
    &
    \multicolumn{1}{p{0.105\textwidth}}{} &
    \multicolumn{1}{p{0.105\textwidth}}{} 
    \\

    \rotatebox{90}{\centering \fontsize{6}{4} \textbf{Input}}
    &
    \includegraphics[width=1cm, height=1cm]
    {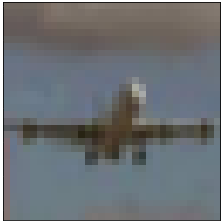}
    & 
    \rotatebox{90}{\centering \textbf{}}
    & 
    \rotatebox{90}{\centering \textbf{}}
    & 
    \rotatebox{90}{\centering \textbf{}}
    &
    \rotatebox{90}{\centering \textbf{}}
    &  
    \rotatebox{90}{\centering \textbf{}}
    &
    \rotatebox{90}{\centering \textbf{}}
    \\

    \multicolumn{1}{p{0.001\linewidth}}{}   
    &
    \multicolumn{1}{p{0.105\linewidth}}{\fontsize{6}{4} \selectfont \centering \textbf{$E_\mu$ IG}} 
    &
    \multicolumn{1}{p{0.105\linewidth}}{\fontsize{6}{4} \selectfont \centering \textbf{$E_\sigma$ IG}}   
    &
    \multicolumn{1}{p{0.105\linewidth}}{\fontsize{6}{4} \selectfont\centering \textbf{CV IG}}  &
    \multicolumn{1}{p{0.105\linewidth}}{\fontsize{6}{4} \selectfont \centering \textbf{$E_\mu$ GBP}}
    &
    \multicolumn{1}{p{0.105\linewidth}}{\fontsize{6}{4} \selectfont \centering \textbf{$E_\sigma$ GBP}}  
    &
    \multicolumn{1}{p{0.105\linewidth}}{\fontsize{6}{4} \selectfont \centering \textbf{CV GBP}}
    \\
\end{tabular}

    \captionsetup{justification=centering}
    \caption{Example explanations of an airplane from CIFAR-10. }
    \label{cifar_experiment_example_2}
    \end{varwidth}
    \end{figure}

\begin{figure}[t]
    \centering
    \begin{varwidth}{0.5\textwidth}
    
    \begin{tabular}
    {
    p{0.001\textwidth}
    p{0.1\textwidth} %
    p{0.1\textwidth}
    p{0.1\textwidth}
    p{0.1\textwidth}
    p{0.1\textwidth}
    p{0.1\textwidth} %
    p{0.1\textwidth} %
    }
    \multicolumn{1}{p{0.001\textwidth}}{\textbf{}}   
    &
    \multicolumn{1}{p{0.105\textwidth}}{} 
    &
    \multicolumn{1}{p{0.105\textwidth}}{}   
    &
    \multicolumn{1}{p{0.105\textwidth}}{}  &
    \multicolumn{1}{p{0.105\textwidth}}{}
    &
    \multicolumn{1}{p{0.105\textwidth}}{}  
    &
    \multicolumn{1}{p{0.105\textwidth}}{} &
    \multicolumn{1}{p{0.105\textwidth}}{} 
    \\

    \rotatebox{90}{\centering \fontsize{6}{4} \textbf{Input}}
    &
    \includegraphics[width=1cm, height=1cm]
    {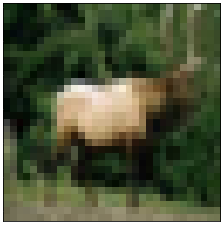}
    & 
    \rotatebox{90}{\centering \textbf{}}
    & 
    \rotatebox{90}{\centering \textbf{}}
    & 
    \rotatebox{90}{\centering \textbf{}}
    &
    \rotatebox{90}{\centering \textbf{}}
    &  
    \rotatebox{90}{\centering \textbf{}}
    &
    \rotatebox{90}{\centering \textbf{}}
    \\

    \multicolumn{1}{p{0.001\linewidth}}{}   
    &
    \multicolumn{1}{p{0.105\linewidth}}{\fontsize{6}{4} \selectfont \centering \textbf{$E_\mu$ IG}} 
    &
    \multicolumn{1}{p{0.105\linewidth}}{\fontsize{6}{4} \selectfont \centering \textbf{$E_\sigma$ IG}}   
    &
    \multicolumn{1}{p{0.105\linewidth}}{\fontsize{6}{4} \selectfont\centering \textbf{CV IG}}  &
    \multicolumn{1}{p{0.105\linewidth}}{\fontsize{6}{4} \selectfont \centering \textbf{$E_\mu$ GBP}}
    &
    \multicolumn{1}{p{0.105\linewidth}}{\fontsize{6}{4} \selectfont \centering \textbf{$E_\sigma$ GBP}}  
    &
    \multicolumn{1}{p{0.105\linewidth}}{\fontsize{6}{4} \selectfont \centering \textbf{CV GBP}}
    \\
\end{tabular}

    \captionsetup{justification=centering}
    \caption{Example explanations of a deer from CIFAR-10. }
    \label{cifar_experiment_example_3}
    \end{varwidth}
    \end{figure}

\begin{figure}
    \centering
    \begin{varwidth}{0.5\textwidth}
    
    \begin{tabular}
    {
    p{0.001\textwidth}
    p{0.1\textwidth} %
    p{0.1\textwidth}
    p{0.1\textwidth}
    p{0.1\textwidth}
    p{0.1\textwidth}
    p{0.1\textwidth} %
    p{0.1\textwidth} %
    }
    \multicolumn{1}{p{0.001\textwidth}}{\textbf{}}   
    &
    \multicolumn{1}{p{0.105\textwidth}}{} 
    &
    \multicolumn{1}{p{0.105\textwidth}}{}   
    &
    \multicolumn{1}{p{0.105\textwidth}}{}  &
    \multicolumn{1}{p{0.105\textwidth}}{}
    &
    \multicolumn{1}{p{0.105\textwidth}}{}  
    &
    \multicolumn{1}{p{0.105\textwidth}}{} &
    \multicolumn{1}{p{0.105\textwidth}}{} 
    \\

    \rotatebox{90}{\centering \fontsize{6}{4} \textbf{Input}}
    &
    \includegraphics[width=1cm, height=1cm]
    {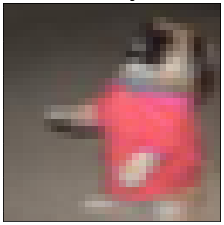}
    & 
    \rotatebox{90}{\centering \textbf{}}
    & 
    \rotatebox{90}{\centering \textbf{}}
    & 
    \rotatebox{90}{\centering \textbf{}}
    &
    \rotatebox{90}{\centering \textbf{}}
    &  
    \rotatebox{90}{\centering \textbf{}}
    &
    \rotatebox{90}{\centering \textbf{}}
    \\

    \multicolumn{1}{p{0.001\linewidth}}{}   
    &
    \multicolumn{1}{p{0.105\linewidth}}{\fontsize{6}{4} \selectfont \centering \textbf{$E_\mu$ IG}} 
    &
    \multicolumn{1}{p{0.105\linewidth}}{\fontsize{6}{4} \selectfont \centering \textbf{$E_\sigma$ IG}}   
    &
    \multicolumn{1}{p{0.105\linewidth}}{\fontsize{6}{4} \selectfont\centering \textbf{CV IG}}  &
    \multicolumn{1}{p{0.105\linewidth}}{\fontsize{6}{4} \selectfont \centering \textbf{$E_\mu$ GBP}}
    &
    \multicolumn{1}{p{0.105\linewidth}}{\fontsize{6}{4} \selectfont \centering \textbf{$E_\sigma$ GBP}}  
    &
    \multicolumn{1}{p{0.105\linewidth}}{\fontsize{6}{4} \selectfont \centering \textbf{CV GBP}}
    \\
\end{tabular}

    \captionsetup{justification=centering}
    \caption{Example explanations of a dog taken from CIFAR-10. }
    \label{cifar_experiment_example_4}
    \end{varwidth}
    \end{figure}

\begin{figure}
    \centering
    \begin{varwidth}{0.5\textwidth}
    
    \begin{tabular}
    {
    p{0.001\textwidth}
    p{0.1\textwidth} %
    p{0.1\textwidth}
    p{0.1\textwidth}
    p{0.1\textwidth}
    p{0.1\textwidth}
    p{0.1\textwidth} %
    p{0.1\textwidth} %
    }
    \multicolumn{1}{p{0.001\textwidth}}{\textbf{}}   
    &
    \multicolumn{1}{p{0.105\textwidth}}{} 
    &
    \multicolumn{1}{p{0.105\textwidth}}{}   
    &
    \multicolumn{1}{p{0.105\textwidth}}{}  &
    \multicolumn{1}{p{0.105\textwidth}}{}
    &
    \multicolumn{1}{p{0.105\textwidth}}{}  
    &
    \multicolumn{1}{p{0.105\textwidth}}{} &
    \multicolumn{1}{p{0.105\textwidth}}{} 
    \\

    \rotatebox{90}{\centering \fontsize{6}{4} \textbf{Input}}
    &
    \includegraphics[width=1cm, height=1cm]{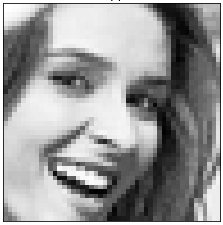}
    & 
    \rotatebox{90}{\centering \textbf{}}
    & 
    \rotatebox{90}{\centering \textbf{}}
    & 
    \rotatebox{90}{\centering \textbf{}}
    &
    \rotatebox{90}{\centering \textbf{}}
    &  
    \rotatebox{90}{\centering \textbf{}}
    &
    \rotatebox{90}{\centering \textbf{}}
    \\

    \multicolumn{1}{p{0.001\linewidth}}{}   
    &
    \multicolumn{1}{p{0.105\linewidth}}{\fontsize{6}{4} \selectfont \centering \textbf{$E_\mu$ IG}} 
    &
    \multicolumn{1}{p{0.105\linewidth}}{\fontsize{6}{4} \selectfont \centering \textbf{$E_\sigma$ IG}}   
    &
    \multicolumn{1}{p{0.105\linewidth}}{\fontsize{6}{4} \selectfont\centering \textbf{CV IG}}  &
    \multicolumn{1}{p{0.105\linewidth}}{\fontsize{6}{4} \selectfont \centering \textbf{$E_\mu$ GBP}}
    &
    \multicolumn{1}{p{0.105\linewidth}}{\fontsize{6}{4} \selectfont \centering \textbf{$E_\sigma$ GBP}}  
    &
    \multicolumn{1}{p{0.105\linewidth}}{\fontsize{6}{4} \selectfont \centering \textbf{CV GBP}}
    \\
\end{tabular}

    \captionsetup{justification=centering}
    \caption{Example explanations of a happy person from FER+. }
    \label{fer_experiment_example_1}
    \end{varwidth}
    \end{figure}

\begin{figure}
    \centering
    \begin{varwidth}{0.5\textwidth}
    
    \begin{tabular}
    {
    p{0.001\textwidth}
    p{0.1\textwidth} %
    p{0.1\textwidth}
    p{0.1\textwidth}
    p{0.1\textwidth}
    p{0.1\textwidth}
    p{0.1\textwidth} %
    p{0.1\textwidth} %
    }
    \multicolumn{1}{p{0.001\textwidth}}{\textbf{}}   
    &
    \multicolumn{1}{p{0.105\textwidth}}{} 
    &
    \multicolumn{1}{p{0.105\textwidth}}{}   
    &
    \multicolumn{1}{p{0.105\textwidth}}{}  &
    \multicolumn{1}{p{0.105\textwidth}}{}
    &
    \multicolumn{1}{p{0.105\textwidth}}{}  
    &
    \multicolumn{1}{p{0.105\textwidth}}{} &
    \multicolumn{1}{p{0.105\textwidth}}{} 
    \\

    \rotatebox{90}{\centering \fontsize{6}{4} \textbf{Input}}
    &
    \includegraphics[width=1cm, height=1cm]{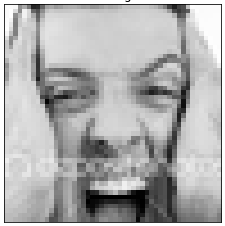}
    & 
    \rotatebox{90}{\centering \textbf{}}
    & 
    \rotatebox{90}{\centering \textbf{}}
    & 
    \rotatebox{90}{\centering \textbf{}}
    &
    \rotatebox{90}{\centering \textbf{}}
    &  
    \rotatebox{90}{\centering \textbf{}}
    &
    \rotatebox{90}{\centering \textbf{}}
    \\

    \multicolumn{1}{p{0.001\linewidth}}{}   
    &
    \multicolumn{1}{p{0.105\linewidth}}{\fontsize{6}{4} \selectfont \centering \textbf{$E_\mu$ IG}} 
    &
    \multicolumn{1}{p{0.105\linewidth}}{\fontsize{6}{4} \selectfont \centering \textbf{$E_\sigma$ IG}}   
    &
    \multicolumn{1}{p{0.105\linewidth}}{\fontsize{6}{4} \selectfont\centering \textbf{CV IG}}  &
    \multicolumn{1}{p{0.105\linewidth}}{\fontsize{6}{4} \selectfont \centering \textbf{$E_\mu$ GBP}}
    &
    \multicolumn{1}{p{0.105\linewidth}}{\fontsize{6}{4} \selectfont \centering \textbf{$E_\sigma$ GBP}}  
    &
    \multicolumn{1}{p{0.105\linewidth}}{\fontsize{6}{4} \selectfont \centering \textbf{CV GBP}}
    \\
\end{tabular}

    \captionsetup{justification=centering}
    \caption{Example explanations of an angry person from FER+. }
    \label{fer_experiment_example_3}
    \end{varwidth}
    \end{figure}

\begin{figure}
    \centering
    \begin{varwidth}{0.5\textwidth}
    
    \begin{tabular}
    {
    p{0.001\textwidth}
    p{0.1\textwidth} %
    p{0.1\textwidth}
    p{0.1\textwidth}
    p{0.1\textwidth}
    p{0.1\textwidth}
    p{0.1\textwidth} %
    p{0.1\textwidth} %
    }
    \multicolumn{1}{p{0.001\textwidth}}{\textbf{}}   
    &
    \multicolumn{1}{p{0.105\textwidth}}{} 
    &
    \multicolumn{1}{p{0.105\textwidth}}{}   
    &
    \multicolumn{1}{p{0.105\textwidth}}{}  &
    \multicolumn{1}{p{0.105\textwidth}}{}
    &
    \multicolumn{1}{p{0.105\textwidth}}{}  
    &
    \multicolumn{1}{p{0.105\textwidth}}{} &
    \multicolumn{1}{p{0.105\textwidth}}{} 
    \\

    \rotatebox{90}{\centering \fontsize{6}{4} \textbf{Input}}
    &
    \includegraphics[width=1cm, height=1cm]{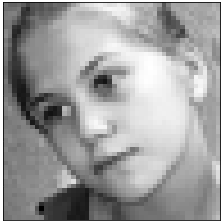}
    & 
    \rotatebox{90}{\centering \textbf{}}
    & 
    \rotatebox{90}{\centering \textbf{}}
    & 
    \rotatebox{90}{\centering \textbf{}}
    &
    \rotatebox{90}{\centering \textbf{}}
    &  
    \rotatebox{90}{\centering \textbf{}}
    &
    \rotatebox{90}{\centering \textbf{}}
    \\

    \multicolumn{1}{p{0.001\linewidth}}{}   
    &
    \multicolumn{1}{p{0.105\linewidth}}{\fontsize{6}{4} \selectfont \centering \textbf{$E_\mu$ IG}} 
    &
    \multicolumn{1}{p{0.105\linewidth}}{\fontsize{6}{4} \selectfont \centering \textbf{$E_\sigma$ IG}}   
    &
    \multicolumn{1}{p{0.105\linewidth}}{\fontsize{6}{4} \selectfont\centering \textbf{CV IG}}  &
    \multicolumn{1}{p{0.105\linewidth}}{\fontsize{6}{4} \selectfont \centering \textbf{$E_\mu$ GBP}}
    &
    \multicolumn{1}{p{0.105\linewidth}}{\fontsize{6}{4} \selectfont \centering \textbf{$E_\sigma$ GBP}}  
    &
    \multicolumn{1}{p{0.105\linewidth}}{\fontsize{6}{4} \selectfont \centering \textbf{CV GBP}}
    \\
\end{tabular}

    \captionsetup{justification=centering}
    \caption{Example explanations of a neutral person from FER+.}
    \label{fer_experiment_example_4}
    \end{varwidth}
    \end{figure}

Additional examples of explanation uncertainty (similar to Figures \ref{cifar_experiment_example_1} and \ref{fer_variant_experiment_example_1}) have been provided in Figures \ref{cifar_experiment_example_2} - \ref{fer_experiment_example_4} for both the CIFAR10 and FER+ datasets. It should be noted that the heatmaps in the additional examples are \textit{not} normalized, unlike the figures in the main paper which have proper normalization.

\end{document}